\PassOptionsToPackage{dvipsnames}{xcolor}
\documentclass[lettersize,journal]{IEEEtran}
\usepackage{amsmath,amsfonts}
\usepackage{algorithmic}
\usepackage{algorithm}
\usepackage{array}
\usepackage[caption=false,font=normalsize,labelfont=sf,textfont=sf]{subfig}
\usepackage{textcomp}
\usepackage{stfloats}
\usepackage{url}
\usepackage{verbatim}
\usepackage{graphicx}
\usepackage{cite}
\usepackage{multirow}
\usepackage{booktabs}
\usepackage[dvipsnames]{xcolor}
\usepackage{amssymb}
\usepackage{pifont}
\usepackage{ulem}
\usepackage{orcidlink}

\newcommand{\cmark}{\ding{51}}%
\newcommand{\xmark}{\ding{55}}%

\definecolor{myred}{RGB}{214, 39, 40}
\definecolor{myblue}{RGB}{31, 119, 180}
\definecolor{myorange}{RGB}{255, 127, 14}
\definecolor{mygreen}{RGB}{44,180,44}

\begin{document}

\title{Learning Goal-based Movement via Motivational-based Models in Cognitive Mobile Robots}

\author{Letícia Berto \orcidlink{0000-0001-5599-192X}, Paula Costa \orcidlink{0000-0002-1534-5744}, Alexandre Simões \orcidlink{0000-0002-1457-6305}, Ricardo Gudwin \orcidlink{0000-0002-9666-3954} and Esther Colombini \orcidlink{0000-0003-0467-3133}
\thanks{E. Colombini and L. Berto are with the Laboratory of Robotics and Cognitive Systems, Institute of Computing, University of Campinas, Brazil}
\thanks{R. Gudwin, P. Costa are with the Dept. of Computer Engineering and Industrial Automation, School of Electrical and Computer Engineering, University of Campinas, Brazil}
\thanks{A. Simões is with Dept. of Control and Automation Engineering, Institute of Science and Technology, São Paulo State University, Sorocaba, Brazil}
\thanks{All authors are with Artificial Intelligence and Cognitive Architectures Hub (H.IAAC), University of Campinas, Brazil}}



\maketitle

\begin{abstract}
Humans have needs motivating their behavior according to intensity and context. However, we also create preferences associated with each action's perceived pleasure, which is susceptible to changes over time. This makes decision-making more complex, requiring learning to balance needs and preferences according to the context. To understand how this process works and enable the development of robots with a motivational-based learning model, we computationally model a motivation theory proposed by Hull. In this model, the agent (an abstraction of a mobile robot) is motivated to keep itself in a state of homeostasis. We added hedonic dimensions to see how preferences affect decision-making, and we employed reinforcement learning to train our motivated-based agents. We run three agents with energy decay rates representing different metabolisms in two different environments to see the impact on their strategy, movement, and behavior. The results show that the agent learned better strategies in the environment that enables choices more adequate according to its metabolism. The use of pleasure in the motivational mechanism significantly impacted behavior learning, mainly for slow metabolism agents. When survival is at risk, the agent ignores pleasure and equilibrium, hinting at how to behave in harsh scenarios. 
\end{abstract}


\section{Introduction}
Traveling around the world, meeting beauty standards, being wealthy and independent, using renewable energies on a large scale, developing aerospace transportation, understanding the universe, exploring Mars, and mortality are desires, dreams, and goals of human beings. To achieve them, the first step is to be motivated. Motivation describes the wants or needs that direct behavior toward a goal \cite{lumen}. 

Motivations are vital for living beings, and they are responsible for essential functions, like (i) guiding behavior to attend to the most critical need at the moment, (ii) designating the amount of energy to be used for executing the selected actions, and (iii) generating learning signals \cite{baldassarre2011intrinsic, panksepp2004affective}. However, although we know the importance of motivation in our lives, we still do not have a consolidated explanation about its functioning. There are days when we genuinely desire to perform a particular action. For some reason, this motivation does not seem to exist on other days. When we are happy, we are willing to take more risks \cite{gruber2011dark}. However, this willingness for risk-taking does not apply in a situation that we assess as problematic \cite{gollwitzer1996psychology}. Some of our goals will not be achieved only by individual effort, but we work for it even so. 

For humans, motivation to engage in activities involves the perception of mastering the activity \cite{white1959motivation}, sense of control over the environment \cite{de2013personal} and appropriate level of learning challenge \cite{csikszentmihalyi1990flow}. As we can see, several factors impact our motivation, such as emotions, internal states, context, priorities, competing thoughts, concerns, life history, day of the week, and even time of year. We have many cognitive functions working together to determine each behavior that we will perform, making the decision-making increasingly complex.

Motivation is divided into \textit{Intrinsic} and \textit{Extrinsic} \cite{baldassarre2011intrinsic, Barto2013}. \textit{Extrinsic motivation} (EM) is related to external rewards that directly impact homeostasis (equilibrium state). It guides behavior learning, increasing survival chances in a particular environment by regulating homeostatic needs. \textit{Intrinsic motivation} (IM) is related to internal rewards detected within the brain's inherent satisfactions, and it allows learning skills without the necessity of immediate direct impact on homeostasis. These skills are used to learn complex behaviors involving long chains of actions to regulate homeostatic needs. Although EM and IM work differently, they are connected, and psychologists found evidence indicating IM plays an essential function in creatures’ behavior that yields a reward for learning activities \cite{ryan2000intrinsic, berlyne1960conflict, deci2013intrinsic}. IM is related to Reinforcement learning and has been attracting many researchers of cognitive computing and robotics \cite{oudeyer2009intrinsic, mirolli2013functions, kaplan2007search, schembri2007evolving, schmidhuber2010formal}.

Besides all the variables already mentioned, there is the dichotomy of \textit{wanting x liking} \cite{berridge2009dissecting}, related to the motivational mechanism. This dichotomy makes decision-making and learning even more complex. The \textit{want} phenomenon occurs in the present. However, it relates to the future when the desire will be satisfied, establishing a goal. For example, I am hungry, so I want to eat, and after that, I will feel good again. So, \textit{want} is related to \textit{needs} over time. The \textit{liking} phenomenon is associated with the present, and it is a subjective evaluation, internally perception, of a sensation or interaction (considering these, we will use the terms \textit{liking} and \textit{pleasure} as synonyms in this paper). I ate chocolate, and I liked it. Hence, \textit{liking} is related to the object and varies as a function of the agent's state, and environment's perception \cite{cos2013hedonic}. Even though these two phenomenons work differently, they are related because we usually \textit{want} things we \textit{like} \cite{Gudwin2019}. According to \cite{berridge2008affective}, pleasure influences cognition, and cognition influences pleasure. 

Given that the \textit{liking} mechanism works as a subjective evaluation, we can use it as a component to develop preferences. In this way, preferences are emerging features of learning about the reward. As in any learning situation, several parameters impact this process, like the context in which it happens, the state of the internal variables, the individual's needs, past experiences, and external factors like social impact and influence. So, we consider that preferences can be an evolutionary mechanism. Once established, they can bias our decision-making, resulting in faster decisions. That is an advantage since the traditional process of perceiving the environment, deciding, acting in the world, and analyzing the result is costly. However, preferences can speed up this process via shortcuts.

Although preferences allow the agent to make faster decisions, they can be tricky. The scenario when the preference was defined can be different from the one existing when using it, so it is necessary to evaluate if the preference remains the same or if it is required to update (or possibly change) it. In some situations, making a decision considering only the preference reward return could not be beneficial in the long-term run, so pondering between the short-term and long-term outcomes for choosing based on preferences or not is essential. Therefore, although having preferences can benefit our systems, it is fundamental to address whether we should use them.

Satisfying our needs or experiencing pleasure adds another layer of complexity to the motivation and behavior relationship. It requires one to reason considering context and priorities. Nevertheless, people are very different, even when acting in the same environment. This fact leads us to question, "How does this balancing mechanism work? How do distinct people learn and make decisions in similar surroundings? How do they act when needs and pleasure compete?". 

As we can see, many factors impact our motivation. Furthermore, our motivation affects our decision-making and behavior. A crucial remark is that we live in a social world, so our behavior impacts others and is one of the factors guiding our social interactions. For example, when we are starving, we look for food. Still, this physiological need can also impact our emotions and mood, like being stressed and impatient, making us less willing for social interactions. On the other side, if we feel sad or lonely, we seek our friends to feel better. Besides, we consider pleasures when engaging in activities or looking for new groups to interact with. We usually tend to choose activities that make us feel good and happier. So our needs also impact our social living.

To help address the questions mentioned before and investigate motivation's impact on social beings, we propose a motivation-based computational model. Based on this model, we built autonomous intelligent robots that will behave driven by their motivational mechanism. Using the dichotomy \textit{wanting x liking}, we created two motivational mechanisms: \textbf{M1} and \textbf{M2}. \textbf{M1} approaches the \textit{wanting} mechanism that considers only the reinforcements from the \textit{drive} reduction (\textit{needs} satisfied) in the learning process following the ideas of Hull's Drive Reduction theory. \textbf{M2} approaches the \textit{wanting} mechanism and the \textit{liking} one, which learns by reinforcements from \textit{drive} reduction and pleasure. We aim to analyze the impact of the hedonic dimension (\textit{liking}) in learning and evaluate the agent's balance between needs and pleasure. In this way, our experiments are incrementally complex.

Usually, in a real-world system, there are several \textit{needs}, which increases the complexity of the decision-making as each one is considered a feature. Also, having multiple features makes comprehending their impact on the learned behavior more challenging. In a previous work \cite{Berto_IGT}, we investigated only the \textit{wanting} mechanism with multiple \textit{drives} using M1 and considering short and long-term decisions. Results showed that in the long-term decision, the agent balances multiple \textit{needs}, which is advantageous in the long run, while prioritizing one in the short-term, which is disadvantageous in the long run.

In this paper, we executed a new set of experiments considering both motivational models under a new perspective: examine how distinct agents, but with the exact motivational mechanisms, learn to act in the same environment according to their specific necessities, how the environment influences our decisions and how good is using the heuristic data (the pleasure in this case) when making decisions. To reduce the investigation's complexity and approach it in more depth, we developed an agent with only one \textit{drive}, equipped with M1 and M2. With the experience and results obtained from the previous work \cite{Berto_IGT}, our agent makes long-term decisions. We started our studies focused on understanding how these mechanisms underlie the behavior exhibited in individuals with biological systems. Focusing on approaching these aspects first leads us to create intelligent cognitive agents \cite{Krichmar2018}. Then we can evolve to more complex and broader scenarios in future works.

We ran a series of experiments considering different agents, environments, and motivation mechanisms. We simulated three types of agents with different energy consumption rates -- called metabolism -- to analyze how agents with distinct metabolic curves would act in the same environment. To better understand how the environment impacts learning and the strategies adopted by each agent, we simulated two environments.

We used Q-learning \cite{qlearn, qlearnBookDana, qlearnBarto} with a function approximator to allow for complex state and action representations as the Reinforcement Learning (RL) algorithm. The results show that the agent learned better strategies when running a \textit{want + liking} motivational mechanism in the most challenging environment.

\subsection{Contributions}
As contributions of this work, we emphasize the following:
\begin{itemize}
    \item The proposal of a motivational computational model based on the Drive Reduction Theory and hedonic dimensions;
    \item The construction of a decision-making mechanism to the motivational system based on RL;
    \item The use of drives explicitly to learn the decision-making and then implicitly to make a decision using the policy learned;
    \item An analysis of the \textit{wanting x liking} dichotomy with artificial creatures;
    \item An investigation of the impact of different environments in the agent's behavior strategy;
    \item An investigation of the impact of different agent's structure and profiles in agent's behavior strategy;
    \item An exploration of the impact of using fixed inherited preferences without updating them during the motivational-based learning in decision-making.
\end{itemize}

All the code from this work is public available at our repository\footnote{\url{www.github.com/larocs/motivatedAgent}}. 

\section{Related Works}\label{sec:rw}
Aiming to create autonomous and intelligent artificial creatures that can act according to their own needs, many researchers have implemented artificial agents that take into account homeostasis and pleasure in the decision making \cite{canamero1997modeling, canamero1997hormonal, canamero2003designing, cos2013hedonic, gadanho2003learning, keramati2011reinforcement, konidaris2006adaptive, salichs2011new, coutinho2005towards, goerke2006emobot, lemhaouriaffect, lewis2016hedonic, lones2017hormone, cao2017collaborative, guerrero2021robot, ziemke2009role, gadanho2001emotion, kitano1995model, breazeal2004designing, salichs, velsquez1997modeling, vouloutsi2013modulating}. 

In \cite{konidaris2006adaptive}, the agent learns to satisfy multiple drives with priorities using the Sarsa(0) algorithm. In \cite{salichs2011new, salichs}, the authors propose a decision-making system based on drives, motivations, and emotions in which the strengths of the internal drive and external stimulus interact. In this way, a motivated behavior is triggered by a strong stimulus if the drive is low and a mild stimulation if the drive is high. This approach can explain why we can choose actions unrelated to our current needs (for example, eating because we like the food and not because we are hungry).

Cao et al. \cite{cao2017collaborative} follow the homeostatic drive concept to design a social behavior controller for robots used to perform tasks with users in HRI experiments. They defined five needs (physical and social) correlated to five drives, and the decision-making system combines the hierarchical approach and the POSH reactive plans. The moment to change between satisfying one drive to another was defined a priori and always followed the same sequence.

Vouloutsi et al. \cite{vouloutsi2013modulating}, and Breazeal  \cite{breazeal1998motivational, breazeal1998regulating, breazeal2004designing} used homeostatic control that classifies the level of each drive into under homeostasis, homeostasis and over homeostasis. \cite{vouloutsi2013modulating} defined multiple drives to be monitored by the homeostatic control. On top of homeostasis, they applied an allostatic control to maintain the system balanced by selecting behavior and assigning priorities to satisfy the needs. The allostatic control switches from a reactive (reflexive) to an adaptive (deals with unpredictability) level, so the agent is not motivated by direct drive satisfaction but aims to match requirements to make an action that leads to a goal available.

Breazeal \cite{breazeal1998motivational, breazeal1998regulating, breazeal2004designing} conducted research using social and stimulation drives for regulating human-robot interaction. In their experiments, they defined multiple drives, each defining specific behaviors. The drives are also a component of the emotional system in a way that homeostatic drives potentiate positive emotions. In contrast, negative emotions are potentiated when a drive is not in homeostasis. The drives were used in action selection and emotional expression, resulting in the robot interacting with the person according to its internal needs and allowing it to express these needs to others.

Using a different approach, Canamero and others \cite{canamero1997modeling, canamero1997hormonal, canamero2003designing, Lones2014, lones2017hormone} propose hormone driven autonomous agents. In their experiments, the agent is required to maintain different homeostatic variables, but it cannot directly sense the level of its homeostatic variables. Instead, the architecture uses artificial hormones secreted to determine the agent's needs in response to these deficits. Among several aspects, they explored motivations that drive behavior selection; emotions modulated by changes in drives; hormones affecting the robot's perceptive, attentional and motivational mechanisms and the intensity of the selected behavior; and multiple agents, each attempting to maintain their own needs in a competitive and non-competitive environment.

The work proposed by Belkaid et al. \cite{belkaid2015} presents a model to allow the robot to build a representation of its peripersonal space based on subjective and motivated perception. They modeled the feeding drive (appetitive) and the safety drive (aversive), which induce an extension or a retraction of the reachable space in the corresponding stimuli direction. Their results showed that the emotional modulation of the peripersonal space makes the robot interact in a way that expresses aspects of their internal states, being more aggressive or fearful in their scenario.

Considering the significant evidence that most living beings internally modulate reward value as a function of their context to expand their range of adaptivity, Cos et al. \cite{cos2013hedonic} explored how the reward is internally processed (resulting in a hedonic value) as a function of the agent's motivational state influences the physiological stability and behavioral adaptivity. In their experiments, the hedonic value is computed using the agent's internal state and the intensity of the physiological effect resulting from a consummatory interaction. Similarly, Gadanho \cite{gadanho2003learning} uses homeostatic variables as part of the goal and adaptive systems, and the state of these variables determines the agent's well-being (used as a reinforcement).

Most of these works use drives to modulate the agent's emotions, observing their behavioral reaction given an emotional state. They primarily use the \textit{drives}' value explicitly in the decision-making process to select an action, defining a motivational behavior. Differently from them, we propose using the drives’ value explicitly in the decision-making learning phase through reinforcement learning just for learning, defining a motivational-based learning model. Then, our action selection using the policy learned is purely reactive (using the \textit{drive}'s value implicitly) after learning the best action that reduces the \textit{drives}.

To the best of our knowledge, just the work proposed by Lewis et al. \cite{lewis2016hedonic} investigated the use of pleasure in homeostasis and decision-making of a motivated autonomous robot that must survive in its environment is similar to ours. They used needs associated with hunger and thirst that could be satisfied by consuming the respective elements in the environment. They used different scenarios, increasing the complexity of finding the resources needed: (I) resources are plentiful, equally distributed, and easily accessible, (II) placed an obstacle in the center of the arena and placed resources so that one area contains only food resources and the other only drink resources, (III) placed an obstacle in the center of the arena, and one resource is more available than other (plentiful food, scarce drink). Concerning pleasure, they defined three models: (I) Pleasure modeled by a hormone released as a function of the satisfaction of homeostatic needs, (II) Different fixed values of the hormone, and (III) Additional hormone (a constant amount) released linked to the execution of consummatory behaviors of \textit{eating} or \textit{drinking}, this additional release is unrelated to the satisfaction of needs. Their results indicate that pleasure has value for homeostatic control in terms of improved viability and increased flexibility in adaptive behavior.

Our work differs from \cite{lewis2016hedonic} in the following aspects: (I) our decision-making is based on reinforcement learning, (II) we simulated three different agent's profiles concerning increase/decrease of the physiological needs, but using the same cognitive architecture, (III) despite the environments with an obstacle, the agent can easily perceive all the resources available 'easily' in \cite{lewis2016hedonic}, in our research the agent has a limited vision, which requires the exploration of the environment to discover the spots that provides energy, (IV) the agent does not know a priori how much energy and pleasure each spot provides, requiring exploration and consummation to find out, (V) environments that provide the same/different resource's values to the needs, (VI) fixed and unique pleasure value associated to each spot, (VII) there is no specific behavior to satisfy the drive, (VIII) our agent is sensitive to the drives and not hormones.

\section{Hull's Drive Reduction Theory}\label{sec:hull}
Although there are several theories of motivation \cite{heckhausen2008motivation}, in this work, we adopted Hull's proposal. Hull's \textbf{Drive Reduction Theory} \cite{hull, hull1952behavior} is based on the concept of homeostasis \cite{cannon1939wisdom}, which is the idea that our body works to maintain a state of balance, like regulating its temperature to ensure that we do not become too hot or too cold. Therefore, motivation would be the basis to drive our body to this state of balance, either physical or psychological.

For Hull, a \textit{drive} indicates the state of arousal or tension caused by biological or physiological needs \cite{hullInfos}. The measure of the \textit{drive} is directly proportional to the intensity of the behavior that would result from it. For example, if you feel a little thirst, you can satisfy yourself with the thought of drinking water later. However, if you feel very thirsty, you cannot stop thinking about it. When present, the \textit{drive} energizes the behavior but does not direct it. Orientation comes from learning, which is a consequence of reinforcement. So, if a behavior results in \textit{drive} reduction, it is reinforced. Hence, which response (behavior) can reduce the \textit{drive} in that specific situation is learned by a feedback signal.

\begin{figure}[h]%
\centering
\includegraphics[width=0.35\textwidth]{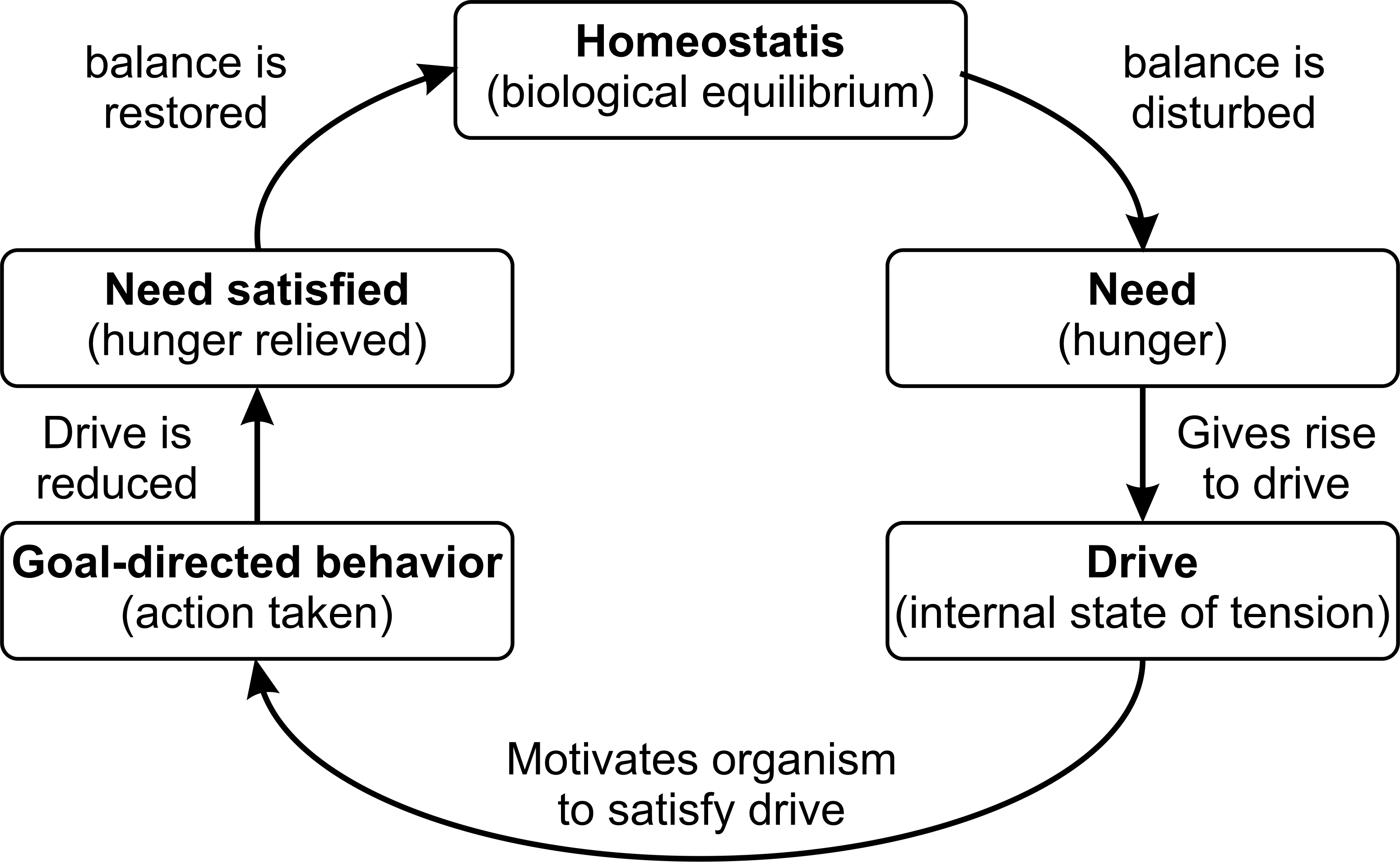}
\caption{Hull's theory. Adpated from \cite{hullCycle}.}\label{fig:hullCycleTheory}
\end{figure}

In summary, Hull's theory (summarized in Figure \ref{fig:hullCycleTheory}) states that a \textit{drive} emerges from organic needs, providing energy for behavior. Hence, a \textit{drive} reduction generates reinforcement and produces learning. However, this theory has some limitations. For example, people with anorexia did not eat (and did not want to) despite the body's strong biological need for food. Another point is that external sources could also provide energy for the behaviors. For example, a person who is not thirsty may feel an intense desire for a drink after tasting, seeing, or smelling their favorite beverage. Furthermore, learning also occurs without necessarily having to be involved with the experience of reducing the \textit{drive} associated with the situation. In addition, learning takes place even by the incitement of the \textit{drive} and not only through its reduction.

Hull’s theory has been the basis of many motivational theories that attempt to address its flaws, like the \textit{Hierarchy of needs} from Maslow’s \cite{maslow, maslow1943theory}.

\subsection{Hull’s Theory adapted to the robotics domain}
Following Hull's theory, we hypothesize that a robot can learn how to behave by directly attending to balancing \textit{needs} and pleasure. To do so, we propose employing \textit{drives} as part of the robot state in a Reinforcement Learning model. Our decision-making process uses the current level of arousal in each \textit{drive} along with sensory data to decide how to behave. Hence, reducing one or more \textit{drives} in the direction of homeostasis results in reinforcement signals that drives the learning process. Furthermore, this reinforcement signal derives from a function that can vary among needs or preferences according to the robot's experience. Hence, the robot's context will also influence behavior once immediate rewards unrelated to the drives can also be received.

\begin{figure}[h]%
\centering
\includegraphics[width=0.3\textwidth]{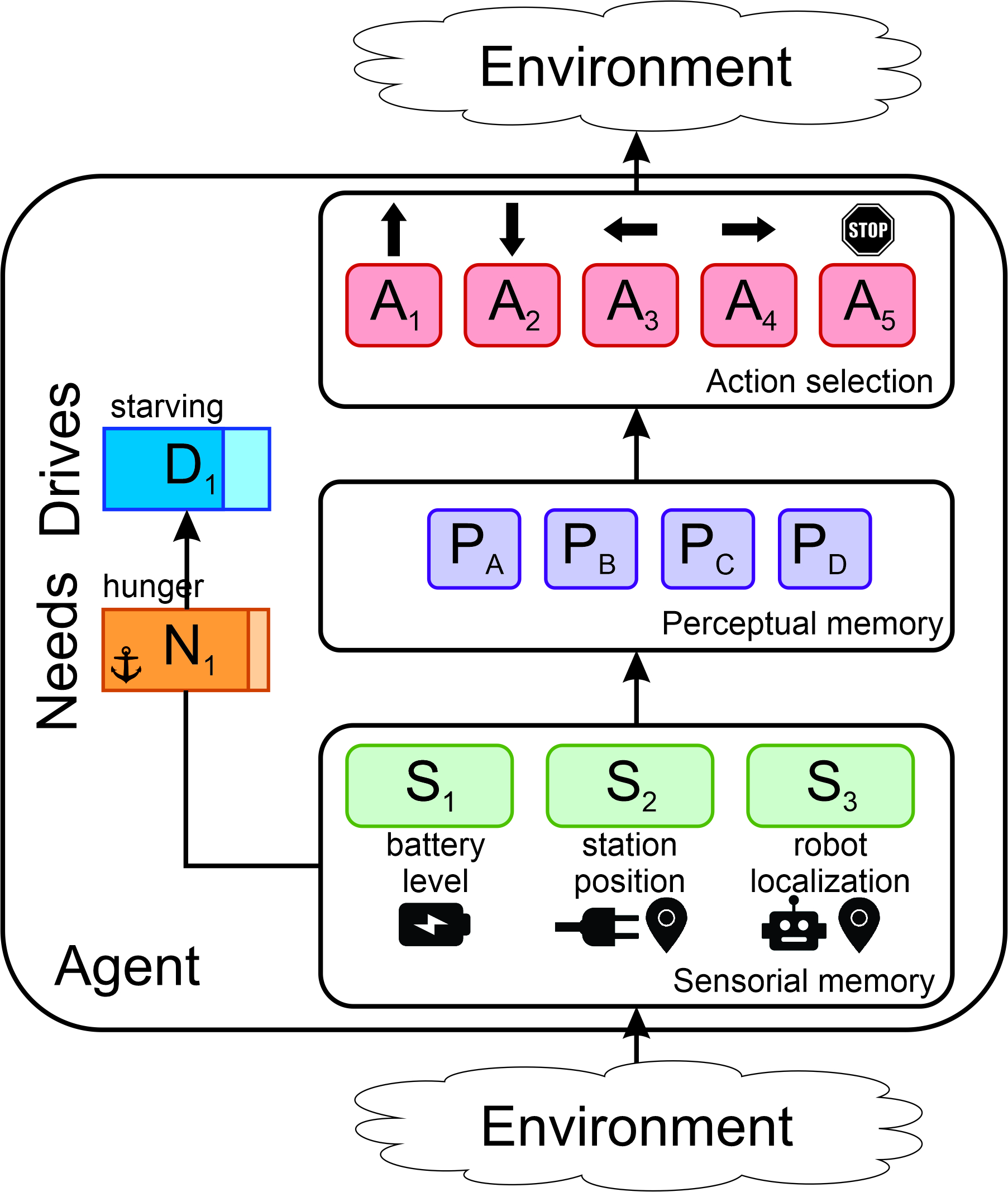}
\caption{Hull's Drive Reduction Theory was adapted to the robotics domain. In this example, an agent is equipped with 3 sensors ($S_1$,$S_2$ and $S_3$). These sensors are used to create a higher-level interpretation of the environment by representing different objects/places. Our agent has one need $N_1$ with its corresponding homeostasis level. The sensors are also used to assess how much this $N_1$ is unsatisfied, represented by its corresponding drive $D_1$. Actions that drive our agent in the environment can be selected by a decision-making mechanism to, for instance, reduce the level of unbalance of the agent.} \label{fig:hullCycle}
\end{figure}

To illustrate an example of Hull's Drive Reduction Theory adapted to the robotics domain, we show in Figure \ref{fig:hullCycle} the agent's setting used in this work. The agent is equipped with battery, localization, and station detector sensors. The battery sensor is related to the \textit{energy need}, which must be satisfied according to its level of homeostasis. If unbalance is detected, i.e., the battery level is not the same as the desired energy need, the corresponding \textit{drive (Survival)} is activated with a level corresponding to this unbalance. Hence, the agent can decide to behave by selecting actions that decrease this \textit{drive}.

\section{Motivational-based learning model}\label{sec:DRT_RL}

We implemented Hull's theory using RL from a computational perspective considering the dichotomy of \textit{wanting} x \textit{liking} related to the motivational mechanism. To connect these theories to our implementation, we established two motivation-based models: \textit{Model 01}, which considers only the \textit{wanting} mechanism, and \textit{Model 02}, which considers the \textit{wanting} and the \textit{liking} mechanism.

\begin{figure}[h]%
\centering
\includegraphics[width=0.5\textwidth]{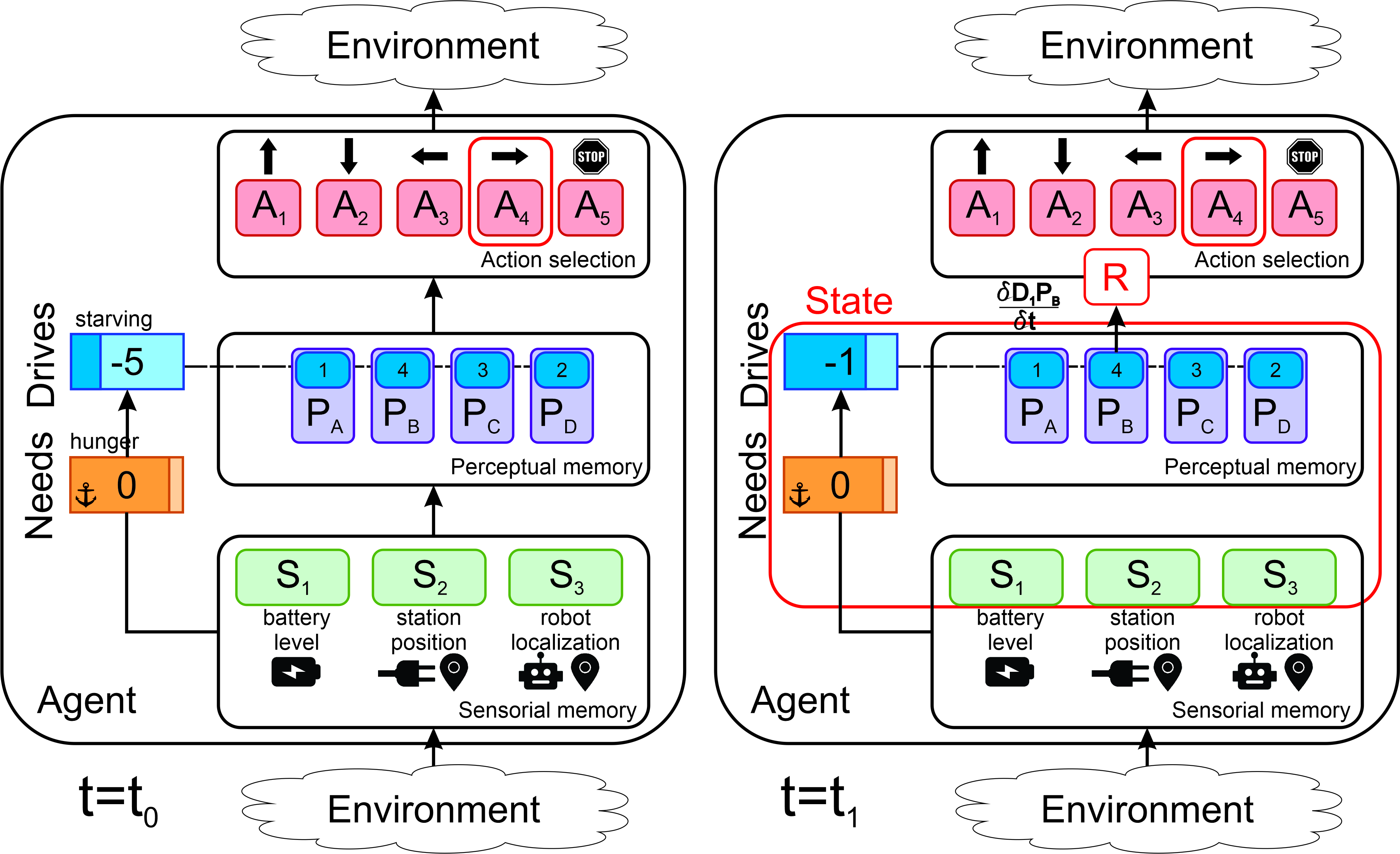}
\caption{\textit{Wanting} mechanism. Behavior learning occurs from the expected reward resulting from \textit{drive} reduction.}\label{fig:BD_M1}
\end{figure}

\textbf{Model 01 (M1):} In this model, we considered only the \textit{wanting} mechanism. Hence, its functioning follows Hull's drive reduction theory (reduce the \textit{drive}) illustrated in Figure \ref{fig:hullCycle}, adapted to generate a reinforcement instead of driving an action. In Figure \ref{fig:BD_M1}, we present the \textit{M1} proposed motivational model. In this model, the agent has $n$ sensors ($S_1-S_n$) used to build $j$ percepts ($P_1-P_j$). Percepts refer to the recognition and interpretation of sensory stimuli, and they are created while the agent is interacting in the world. This percept can be a new one (for example, the first time the agent sees an object) or can be updated (the object is modified). In our case, the percepts correspond to the recharge stations. Then $P_A$, $P_B$, $P_C$ and $P_D$ corresponds to perceptions associated to stations A, B, C and D, respectively. Each percept is associated with a value. This value indicate how much this percept modifies the arousal level in the related \textit{drive} ($D_1$). In our experiments, considering the environment where all the stations provide the same recharge value, the percepts values are $P_A$ = $P_B$ = $P_C$ = $P_D$ = 3. Considering the environment with different recharge values, the percepts values are $P_A$ = 1, $P_B$ = 4, $P_C$ = 3 and $P_D$ = 2. We use the these values in our example. Hence, when performing an action ($A_i$) that alters a \textit{drive}, the agent calculates the expected reward ($R$) based on $\delta D_{i_{P_j}} / \delta t$. Following an RL approach, the agent can learn to pick actions that maximize its \textit{drive} reduction gain. Hence, the resulting action indicates the more efficient behavior to attend to the agent's current \textit{needs} in the direction of homeostasis. 

\begin{figure}[h]%
\centering
\includegraphics[width=0.5\textwidth]{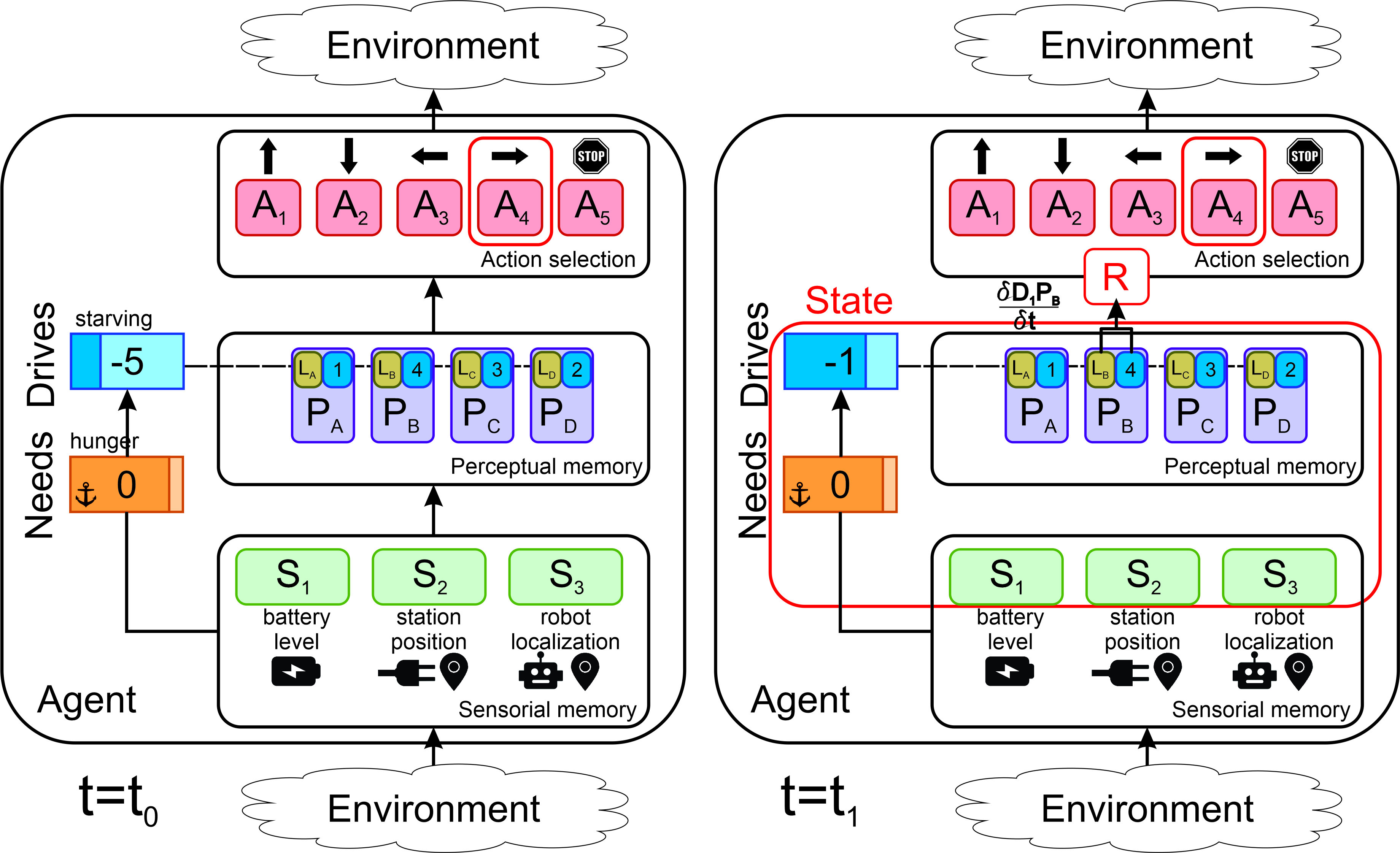}
\caption{\textit{Wanting + Liking} mechanisms. Behavior learning occurs from the expected reward resulting from \textit{drive} reduction and hedonic dimension ($L_j$).}\label{fig:BD_M2}
\end{figure}

\textbf{Model 02 (M2):} In our second motivational model, we considered both \textit{wanting} and \textit{liking} mechanisms. \textit{Wanting} works the same as \textit{M1} but now the reward $R$ also depends on the \textit{Liking} component, as we can see in Figure \ref{fig:BD_M2}. \textit{Liking} corresponds to the hedonic dimension composed of hedonic values, which are subjective evaluations of sensations associated with the percept (object/event) and unrelated to need satisfaction. The hedonic dimension might have an important role in establishing the agent's preferences \cite{young1961motivation,frijda1986emotions,cos2013hedonic}. Hence, they represent a single value ($L_j$) associated with each percept ($P_j$). In our experiments, these ($L_j$) values are $L_A$ = 3, $L_B$ = 2, $L_C$ = 4, and $L_D$ = 1. Learning these values can occur in two ways: (I) fixed value learned from the evolutionary process, (II) over time according to the experiences. In the scope of our research, we used the first option: fixed values associated with each object available in the environment. 

In summary, in \textit{M1}, the motivation comes exclusively from \textit{drive} reduction expected reward, while in \textit{M2}, it comes from \textit{drive} reduction plus the pleasure (hedonic value) expected rewards. 

\section{Experimental Setup }\label{sec:setup}
This section presents the environment configuration, the agent's mind and the cognitive cycle used by our robot in the following experiments.

\subsection{Environment}
In this work, we employ simulated simple agents to ensure the focus on the analysis of the proposed model, avoiding possible complexities added by real robots. In previous work \cite{Berto_IGT} we evaluated the \textit{wanting} mechanism using a high fidelity simulator. Although the simulator provides more fidelity to the real world, it brings some complexities, like the lack of environmental control. Considering that we are proposing two new motivational-based learning models, we need to have control of the environment to effectively analyze the impacts of the models on the agent and guarantee that the scenarios and conditions are the same for all of them. To this end, we designed a simulated environment that meet our needs.

\begin{figure}[h]%
\centering
\includegraphics[width=0.3\textwidth]{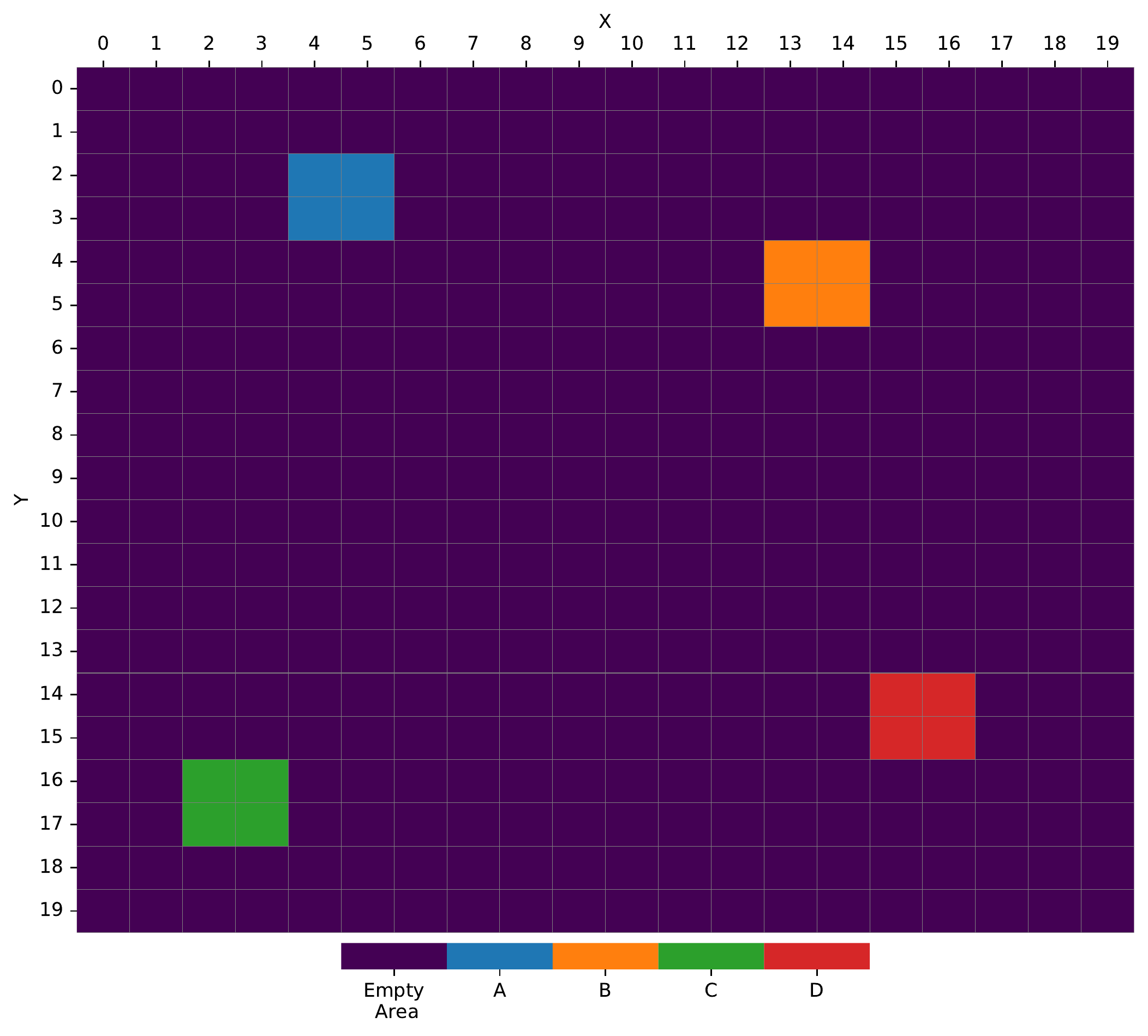}
\caption{20x20 environment used in the experiments.}\label{fig:env}
\end{figure}

All the experiments were executed using a simulated agent moving in a 20x20 deterministic environment (illustrated in Figure \ref{fig:env}). There are four fixed recharge areas ($A$, $B$, $C$, $D$) in this environment, and each of them is associated with a specific recharge value. These areas are used to simulate $j$ percepts  ($P_A$, $P_B$, $P_C$, $P_D$) that can be sensed by the agent. In all experiments, a single need ($N_1$) and its corresponding drive ($D_1$) will be employed. Hence, in the experiments that employ the \textit{M1} model, each percept has a single value associated to it corresponding to the $\delta D_{1_{P_j}}/ \delta t$ that this particular percept causes in ($D_1$). We also simulate environments with percepts with different drive reduction rates (different recharge values, shown in Table \ref{tab:rechargeValues}). For  experiments with the \textit{M2} model, there is a pleasure value ($L_j$) associated with the percept. So, it is possible to analyze the agent's needs and preferences in different situations. 

\addtolength{\tabcolsep}{+1pt}  
\begin{table}[htb]
\begin{center}
\caption{Values associated with each recharge area.}\label{tab:rechargeValues}%
\begin{tabular}{@{}ccccc@{}}
 
\toprule
 &&\multicolumn{2}{c}{Want $\delta D_{1_{P_j}}/\delta t$}  & Like $L_j$ \\ 
\cmidrule{3-4} \cmidrule{5-5}
& Percept $P_j$ & Same  & Different  & Station \\
& & Recharge & Recharge & Pleasure\\

\midrule
\textcolor{myblue}{$\blacksquare$} &A & 3 & 1 & 3 \\
\textcolor{myorange}{$\blacksquare$}&B & 3 & 4 & 2 \\
\textcolor{mygreen}{$\blacksquare$}&C & 3 & 3 & 4 \\
\textcolor{myred}{$\blacksquare$}&D & 3 & 2 & 1 \\
\bottomrule 
\end{tabular}
\end{center}
\end{table}
\addtolength{\tabcolsep}{-1pt} 

\subsection{Agent's Mind}
\begin{figure*}[h]
\centering
    \includegraphics[width=0.9\textwidth]{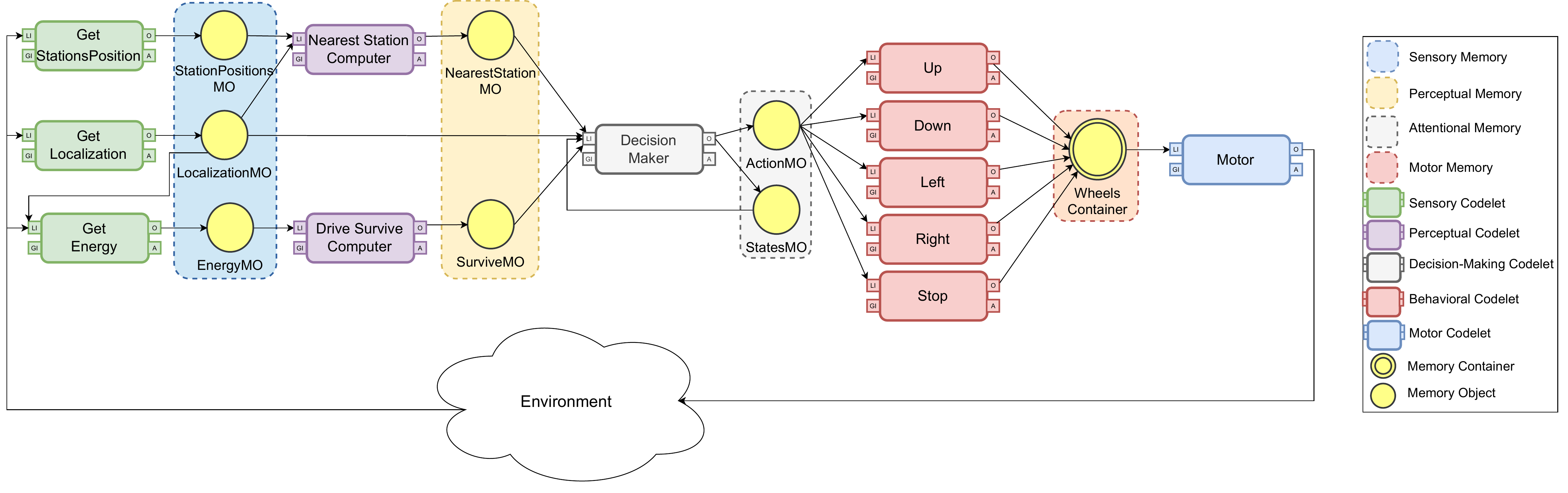}
    \caption{Robot's cognitive cycle.}
    \label{fig:robotCycle}
\end{figure*}

To implement our proposed computational models,  we developed a cognitive agent that follows the cognitive cycle illustrated in Figure \ref{fig:robotCycle} and detailed next. 

\subsubsection{Sensing}
\begin{figure}
    \centering
    \includegraphics[width=0.3\textwidth]{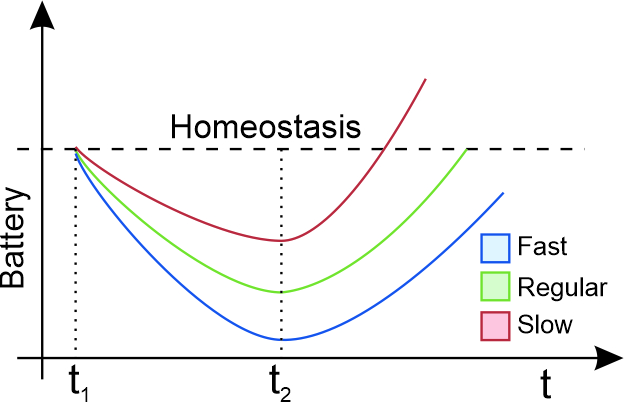}
    \caption{Battery decay for fast, regular, and slow metabolism agents. From $t_1$ to $t_2$, as no recharge is carried out, the agents escape homeostasis with different decay rates. In $t_2$, they begin recharging. The effect of charging is added to their metabolism rate. Hence, they increase the battery level while spending it to survive. }
    \label{fig:metHomeos}
\end{figure}

Our agent is equipped with three sensors: a simulated battery sensor ($S_b$), a localization sensor ($S_l$), and a station detector ($S_s$). The localization and station detector sensors are used to compute where the agent is and the agent's distance to the closest recharge station. They are used as part of the state for the decision-making mechanism. However, only the battery or energy sensor $S_b \in [0-50]$ is used to define the agent's unbalance associated to its single \textit{need} $N1$: \textbf{energy}.  Hence, the battery level is $max(50, energy + recharge value)$. The robot energy level decays per simulation step  according to the agent metabolism curve. We define \textbf{three different metabolism types} to simulate agents with distinct profiles: \textbf{slow}, \textbf{regular}, and \textbf{fast}. In this way, each has a different battery decay rate, impacting the energy consumption mode. In \textit{slow metabolism}, the agent consumes $0.1$ energy per time step, while in \textit{regular metabolism}, the consumption is $1$. Finally, for the \textit{fast metabolism} agent, the consumption is $3$ per time step. Figure \ref{fig:metHomeos} depicts the difference among different metabolisms.

\subsubsection{Perception}
According to Hull's theory, the \textbf{energy} \textit{need} gives rise to a \textbf{survival} \textit{drive}. In our experiments, we compute the \textit{survival drive} ($D_1$) by the difference between the current energy value (given by $S_b$) and the homeostasis level associated with this \textit{need}. We defined $homeostasis_{N_1} = 30$, but this value can be changed if necessary. It represents the agent's equilibrium state.

Given the sensor's range, our agent has the following possible values for the survival \textit{drive} $D_1$:

\begin{equation}
D_1 = 
    \begin{cases}
        [-30, ..., -0.1], S_b < homeostasis_{N_1};\\
        0, S_b = homeostasis_{N_1;}\\
        [0.1, ..., 20], S_b > homeostasis_{N_1}.
    \end{cases}
    \label{eq:valuesSurvivalDrive}
\end{equation}

\subsubsection{Decision-making}
The decision-making is composed of two phases: (I)  learning, through interaction with the world, the best behavior based on the drives, and (II) using the knowledge and experience acquired in the previous phase to satisfy the exact needs in a similar environment. The difference is that in phase (I), the policy is updated, while in phase (II) not. In phase (II), the agent can deal with situations not experienced during the learning phase, but with the policy learned, it can find the best solution, generalizing the strategy.

The agent is connected to the environment through perception and action. At each timestep, it perceives the state ($s \in S$ ) of the environment, where $S$ is the set of possible states, and selects an action $a \in A(s)$, where $A(s)$ is the set of possible actions to state $s$ to execute. The action executed changes the environment's state, and the agent sees the value associated with this transition as a reward. Then, it calculates the difference between the expected reward and the reward received, using it to improve prediction accuracy for future rewards and consequently improve future predictions. We can define this process as learning.

In the learning phase, we used Q-learning as the reinforcement learning algorithm. The update rule is given by:
\begin{equation}
\begin{aligned}
    Q(s, a) =& Q(s, a) + \\ & \alpha   (R_{t+1} + \gamma \cdot max_{a'}Q(s', a') - Q(s, a)),
\end{aligned}
\end{equation}

\noindent where, $\alpha$ is the \textit{learning rate}, and $\gamma$ is the temporal discount factor. In these experiments: $\alpha=0.0001$ and $\gamma=0.9$.

Due to the continuous nature of our states, our agent employs the Q-learning algorithm with a function approximator to learn the best behavior policy that maximizes the agent's needs and pleasure. In this case, we represent state and action by a feature vector and the TD target ($R_{t+1} + \gamma \cdot max_{a'}\hat{q}(S_{t+1}, A_{t+1}, w)$) which leverages the max of the current function approximation value. 

RL algorithms have better convergence guarantees if the environment is mathematically modeled as a Markov Decision Process (MDP). Hence, we describe next the MDP formulation for our problem. 

\begin{itemize}
    \item \textbf{Actions} $\mathcal{A}$: Stop, Up, Down, Left, Right. Each action changes the agent's position by $1$, within the environment's range. We employed the $\epsilon$-greedy policy ($\epsilon=1.0$, with linear decay up to $0.01$) as a choice of actions.
    \item \textbf{States}: As we are working with a function approximator, we use a linear combination of features to approximate the agent's state at each time step.
    For this, we have developed the following set of features $\mathcal{F}$ to compose our state:
    \begin{itemize}
        \item \textbf{$D_1$}: \textit{Survival drive} absolute value.
        \item \textit{MinDist}: Euclidean distance from the robot to the nearest recharge area.
        \item \textit{Up, Down, Left, Right}: Binary features relative to the position of the nearest recharge area to the agent. Suppose the closest recharge area is located above or below the agent's current $Y$ position, to the left or right of the agent's current $X$, respectively.
        \item \textit{SeeA, SeeB, SeeC, SeeD}: Binary features that indicate if the agent perceives a recharge area within a maximum distance, simulating the sensor range. In our case, this maximum value is $6$, and the features relate to recharging areas A, B, C, and D, respectively. 
        \item \textit{Y$_{0}$, ..., Y$_{19}$}: Binary features corresponding to the agent's current $Y$ position. 
        \item \textit{X$_{0}$, ..., X$_{19}$}: Similar to $Y$ features, but relative to the agent's current $X$ position.
    \end{itemize}
    
    The final feature set size is $50$. The weights ($W$) that associate this feature set $F$ to the action set $A$ are randomly initialized by a uniform distribution $\mathcal{U}(0.001, 0.009)$. During the training phase, the weights are updated at each time step with gradient descent using:
    
    \begin{equation}
        \begin{aligned}
            \Delta w = & \alpha [(R_{t+1} + \gamma max_{a'}\hat{q}(S_{t+1}, a', w) - \\ & \hat{q}(S_t, A_t, w))\triangledown_w\hat{q}(S_t, A_t, w)]
        \end{aligned}
    \label{eq:features}
    \end{equation}

    \item \textbf{Rewards}:
    Following the motivational models proposed in section \ref{sec:DRT_RL}, we model two reward functions to study the agent's behavior accordingly. The first considers only the motivation generated by the \textit{drive} reduction (Eq. \ref{eq:rewardDrive}), and another considers the \textit{drive} reduction and the pleasure associated with each percept/recharge area (Eq. \ref{eq:rewardDrivePleasure}).
    
        \begin{equation}
        R_1 =
        \begin{cases}
            D_1, D_1 < 0;\\
            1, int(D_1) = 0;\\
            -(D_1 * 0.5), D_1 > 0.
        \end{cases}
        \label{eq:rewardDrive}
    \end{equation}

    \begin{equation}
        R_2 = R_1 + L_j,
        \label{eq:rewardDrivePleasure}
    \end{equation}
    
\noindent where $L_j$ is the pleasure associated with the percept the agent is interacting, if any. In Eq. \ref{eq:rewardDrive}, we define the goal of keeping the agent's energy level as close as possible to homeostasis by rewarding the agent when the drive $D_1$ is $0$. Hence, the agent is in equilibrium. In other cases, we punish the agent's choice proportionally to the distance from homeostasis (hence, the absolute value of $D_1$). It is important to remark that being below homeostasis is hazardous for the robot as its life will end if the battery level reaches $0$. Thus, the further below homeostasis, the greater the risk and the more severe the punishment. On the other hand, being above homeostasis generates a feeling of discomfort (thinking about human beings, it would be the equivalent of eating more than necessary). In this way, the further above homeostasis, the greater the discomfort. Nevertheless, as it is associated only with well-being and not life risk, we apply a $0.5$ gain (it is still bad for the robot, but not as much as being below homeostasis).
    
For our second reward function, described by Eq. \ref{eq:rewardDrivePleasure}, we kept the motivation generated by the \textit{drive} but also inserted a pleasure associated with each percept/recharge area. Reward $R_2$ models a goal of balancing needs and pleasure. 

\end{itemize}

\subsection{Experiments and settings}

In this work, we executed two sets of experiments: \textbf{M1} and \textbf{M2}. For each set, we vary the agent's metabolism curve and the percepts influence in the \textit{drive}. However, we perform training and test phases with the same configuration for all experiments. Table \ref{tab:experiments} depicts each experiment and environment configuration. 

\begin{table}[htb]
\caption{Experiments configuration. For each experiment we list: the motivational model mechanism, the experiment id, the reward function, the agent metabolic curve type, if the recharge areas provide the same modification in the drive, if the liking component is present.}\label{tab:experiments}%
\begin{center}
\begin{tabular}{@{}ccccccc@{}}
\toprule
Model & Expe. no. & Reward & Metabolism & $\delta D_{i_{P_j}}/\delta t$  &  $L_j$ \\

\midrule
\multirow{6}{*}{M1}&EXP01& $R_1$&Slow & \textcolor{mygreen}{=} & \textcolor{myred}{\xmark}  \\
&EXP02& $R_1$&Regular & \textcolor{mygreen}{=} & \textcolor{myred}{\xmark}  \\
&EXP03& $R_1$&Fast &  \textcolor{mygreen}{=} & \textcolor{myred}{\xmark}  \\
&EXP04& $R_1$&Slow & \textcolor{mygreen}{$\neq$ } & \textcolor{myred}{\xmark}  \\
&EXP05& $R_1$&Regular & \textcolor{mygreen}{$\neq$} & \textcolor{myred}{\xmark}  \\
&EXP06& $R_1$&Fast &  \textcolor{mygreen}{$\neq$} & \textcolor{myred}{\xmark}  \\
\bottomrule
\multirow{6}{*}{M2}&EXP07& $R_2$&Slow & \textcolor{mygreen}{ =} & \textcolor{mygreen}{\cmark}  \\
&EXP08& $R_2$&Regular & \textcolor{mygreen}{=} & \textcolor{mygreen}{\cmark}  \\
&EXP09& $R_2$&Fast &  \textcolor{mygreen}{=} & \textcolor{mygreen}{\cmark}  \\
&EXP10& $R_2$&Slow & \textcolor{mygreen}{$\neq$ } & \textcolor{mygreen}{\cmark}  \\
&EXP11& $R_2$&Regular & \textcolor{mygreen}{$\neq$} & \textcolor{mygreen}{\cmark}  \\
&EXP12& $R_2$&Fast &  \textcolor{mygreen}{$\neq$} & \textcolor{mygreen}{\cmark}  \\

\bottomrule
\end{tabular}
\end{center}
\end{table}

For each experiment, we trained the agent for $25,000$ episodes with a maximum of $5,000$ steps each or until the energy runs out. For each episode, we randomly selected the initial energy value ($\mathcal{U}(0,50)$) and the robot position $(X,Y)=(\mathcal{U}(0,19),\mathcal{U}(0,19))$ at the beginning of each episode. 

In the test phase, we load the weights learned and set $\epsilon = 0$. Hence, only the greedy action is selected. Also, the robot can act in the environment while its energy $S_b > 0$ or for a maximum of $8,000$ steps.  

\subsubsection{Data Visualization}
Although we ran $25,000$ episodes during learning, we plot the results using the average $100$ episodes. Our graphics illustrate the mean and standard deviation for reward and the number of actions performed, as it can be seen in the first row of Figure \ref{fig:EXP01_EXP02_EXP03}. To better understand the agent's behavior throughout training, we also create a heatmap that contains the number of visits to each environment position, as illustrated in the second row of Figure \ref{fig:EXP01_EXP02_EXP03}.

Finally, we ran $50$ episodes using the learned policy in the test phase. We started the robot in homeostasis ($S_b=30$, $D_1=0$) and in $(X,Y)$ positions in the environment where it is possible to reach a station considering the agent metabolic curve. In this phase, all initializations are the same for all experiments. To illustrate the results obtained, we plot $D_1$ average obtained for each test, as shown in the last row of Figure \ref{fig:EXP01_EXP02_EXP03}.

\section{M1 experiment set - \textit{Wanting} mechanism} \label{sec:expsDrive}
In this experiment set we employ the first motivational model \textbf{M1}, modeled via the reward function $R_1$ (Eq. \ref{eq:rewardDrive}), in which the agent's goal is to maintain homeostasis. For this set, we run experiments to assess how intrinsic characteristics of an agent, like its metabolic curve, can influence behavior selection when different agents are submitted to the same environment. We also run experiments to assess the influence of the environment in agents with similar characteristics.

\subsection{Same Recharge\small{(EXP01 - EXP03)}}
In the experiments described next, all recharge areas provide the same recharge value (as defined in Tables \ref{tab:rechargeValues} and \ref{tab:experiments}).

\begin{figure*}
    \centering
    \includegraphics[width=0.3\textwidth]{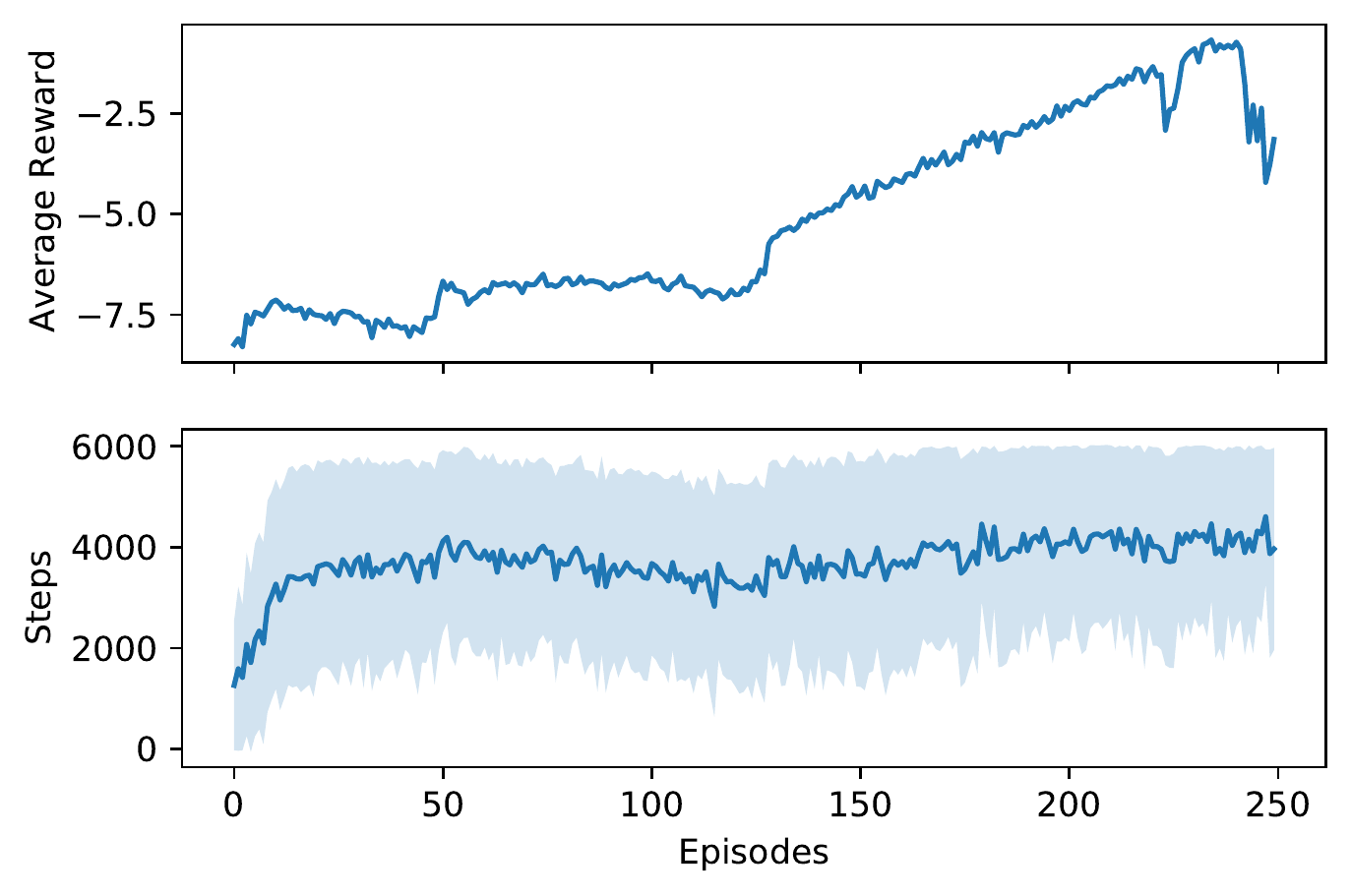}
    \includegraphics[width=0.3\textwidth]{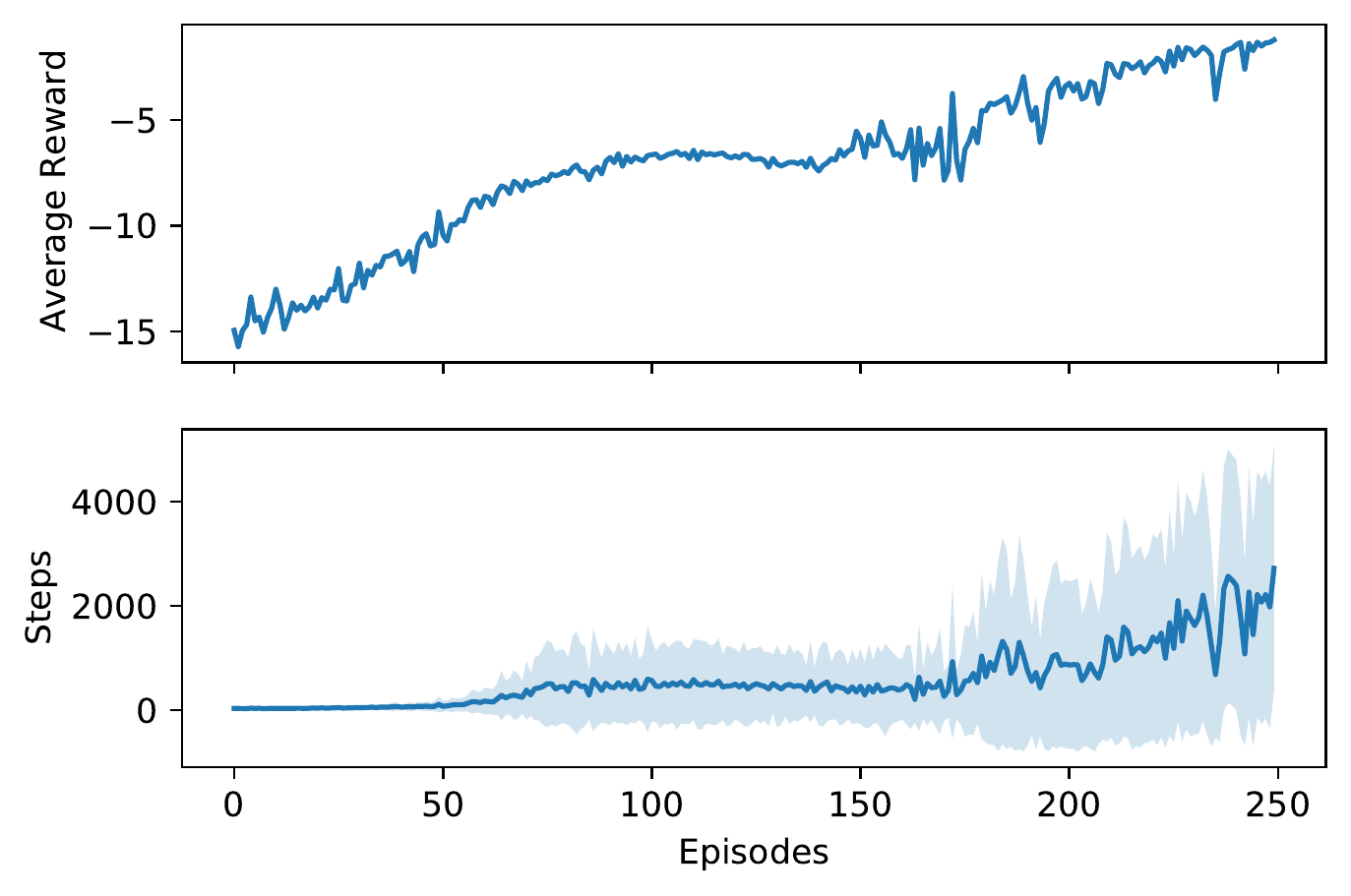}
    \includegraphics[width=0.3\textwidth]{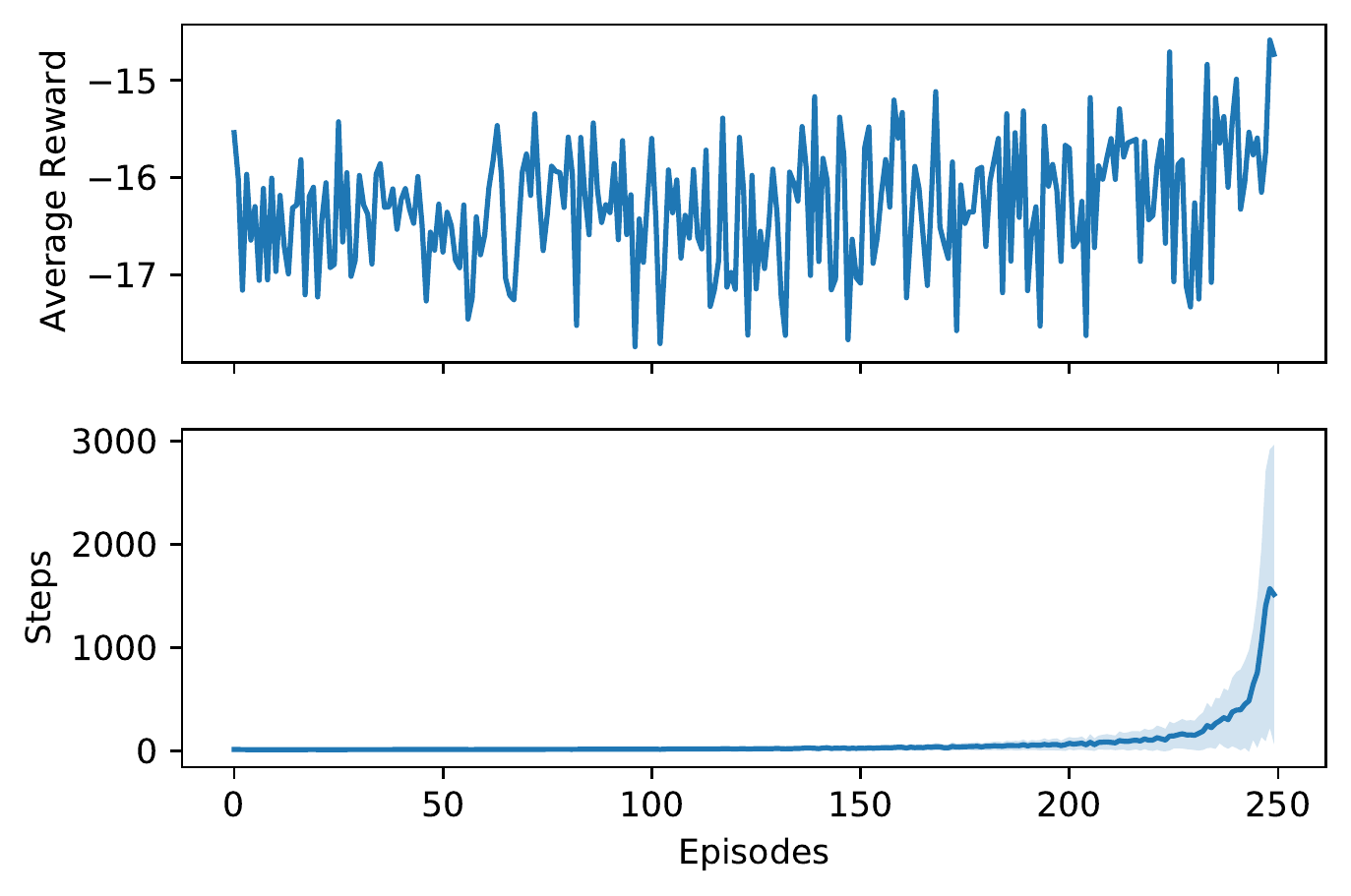}
    \includegraphics[width=0.3\textwidth]{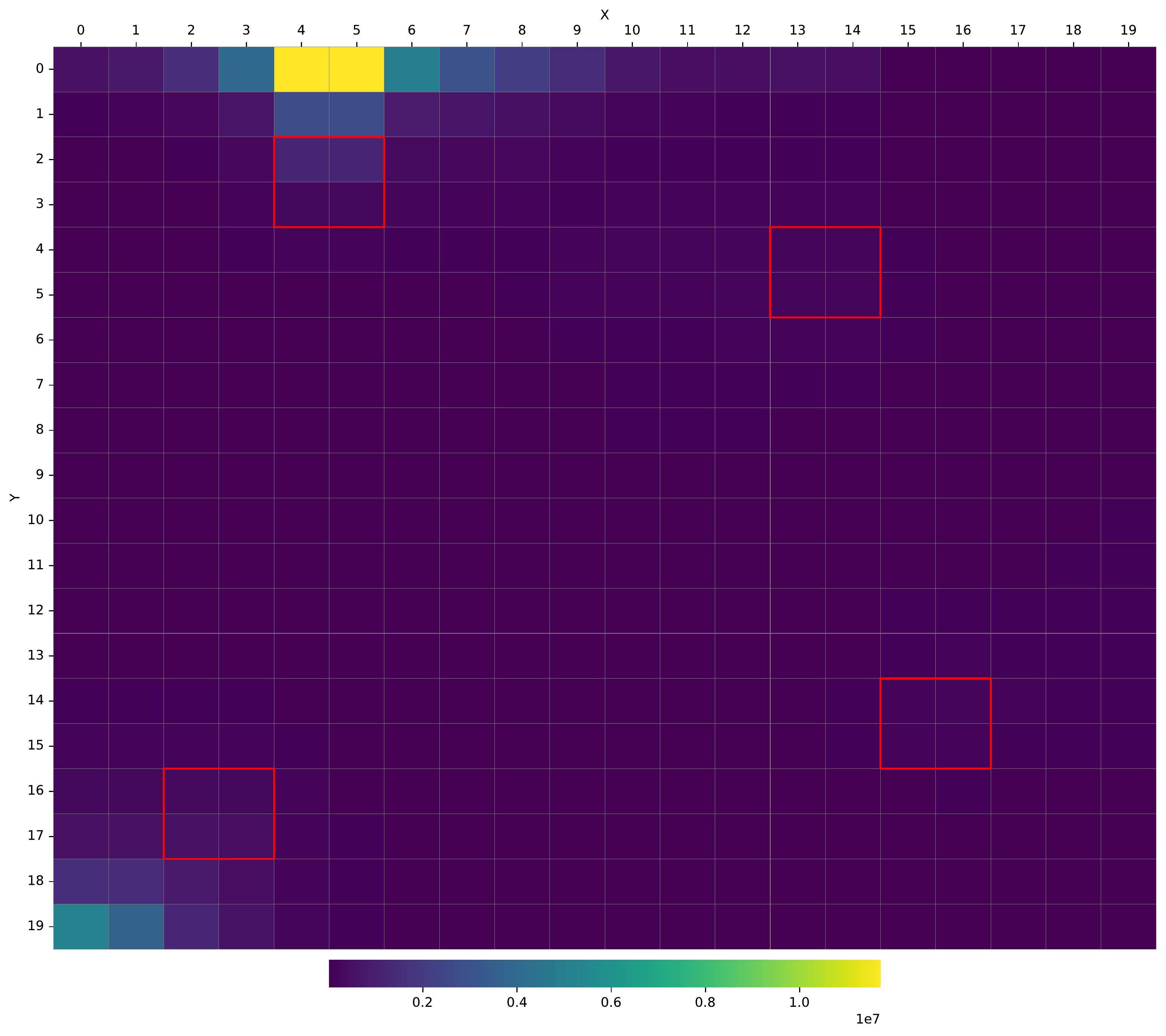}
    \includegraphics[width=0.3\textwidth]{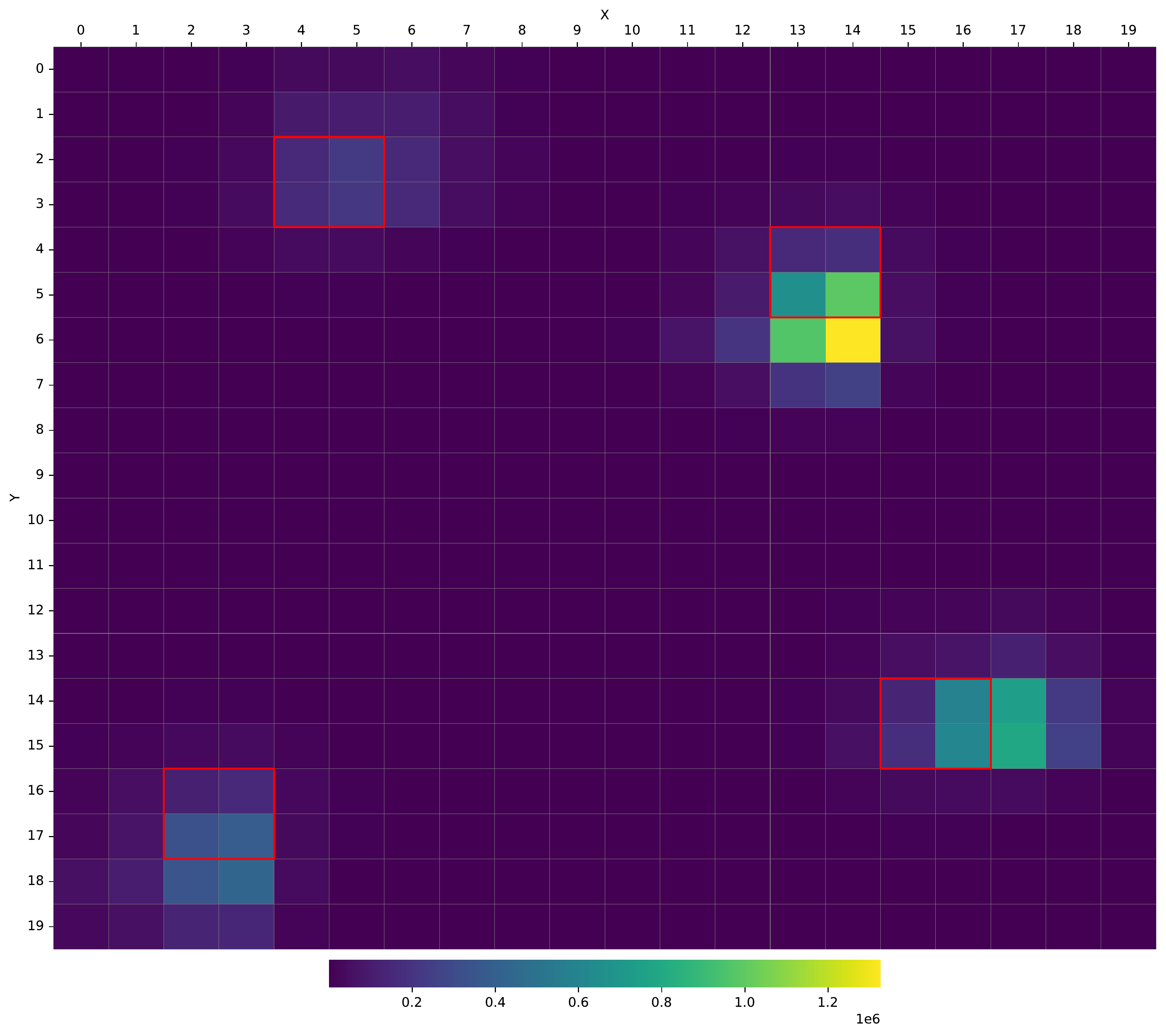}
    \includegraphics[width=0.3\textwidth]{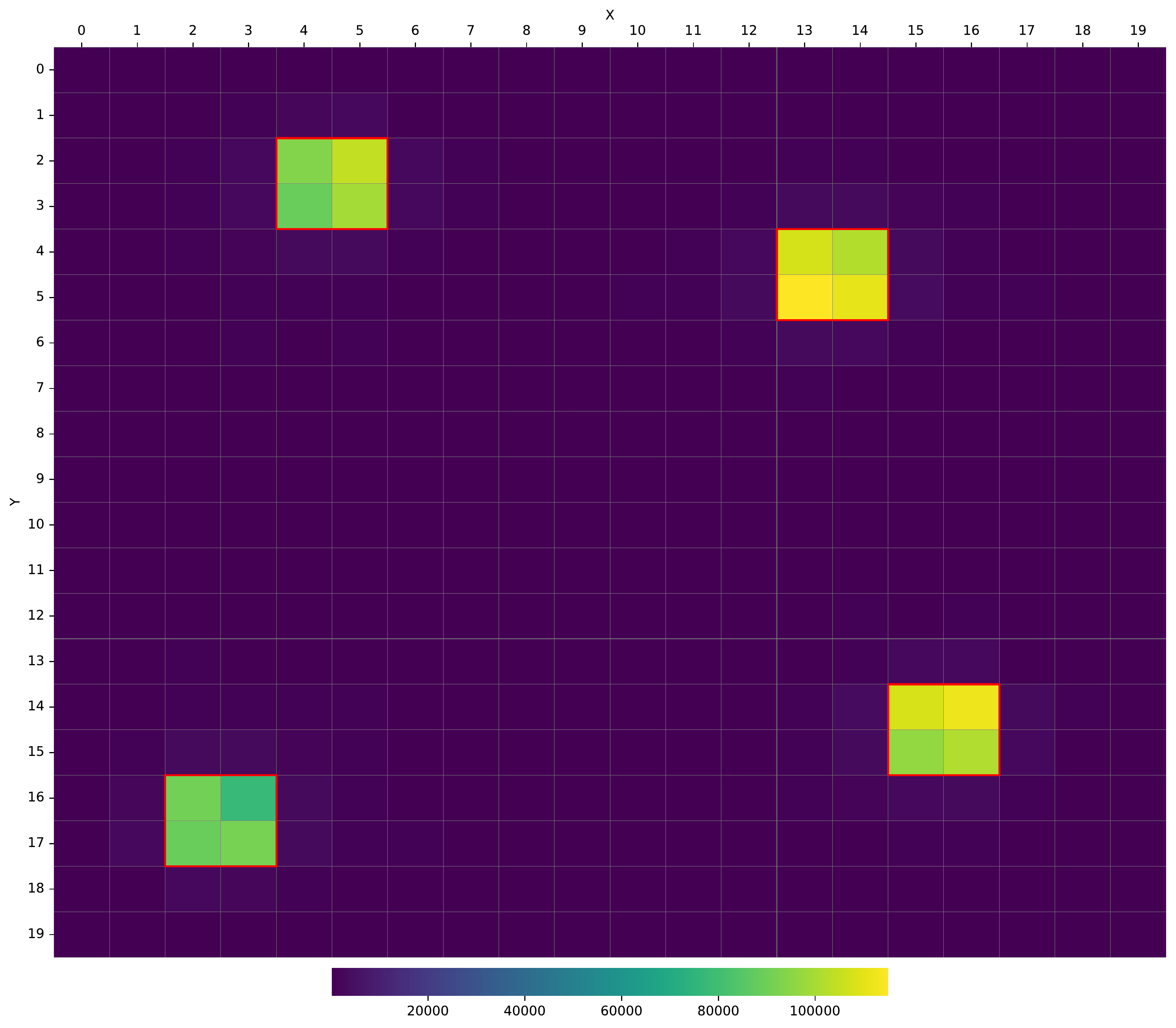}
    \includegraphics[width=0.3\textwidth]{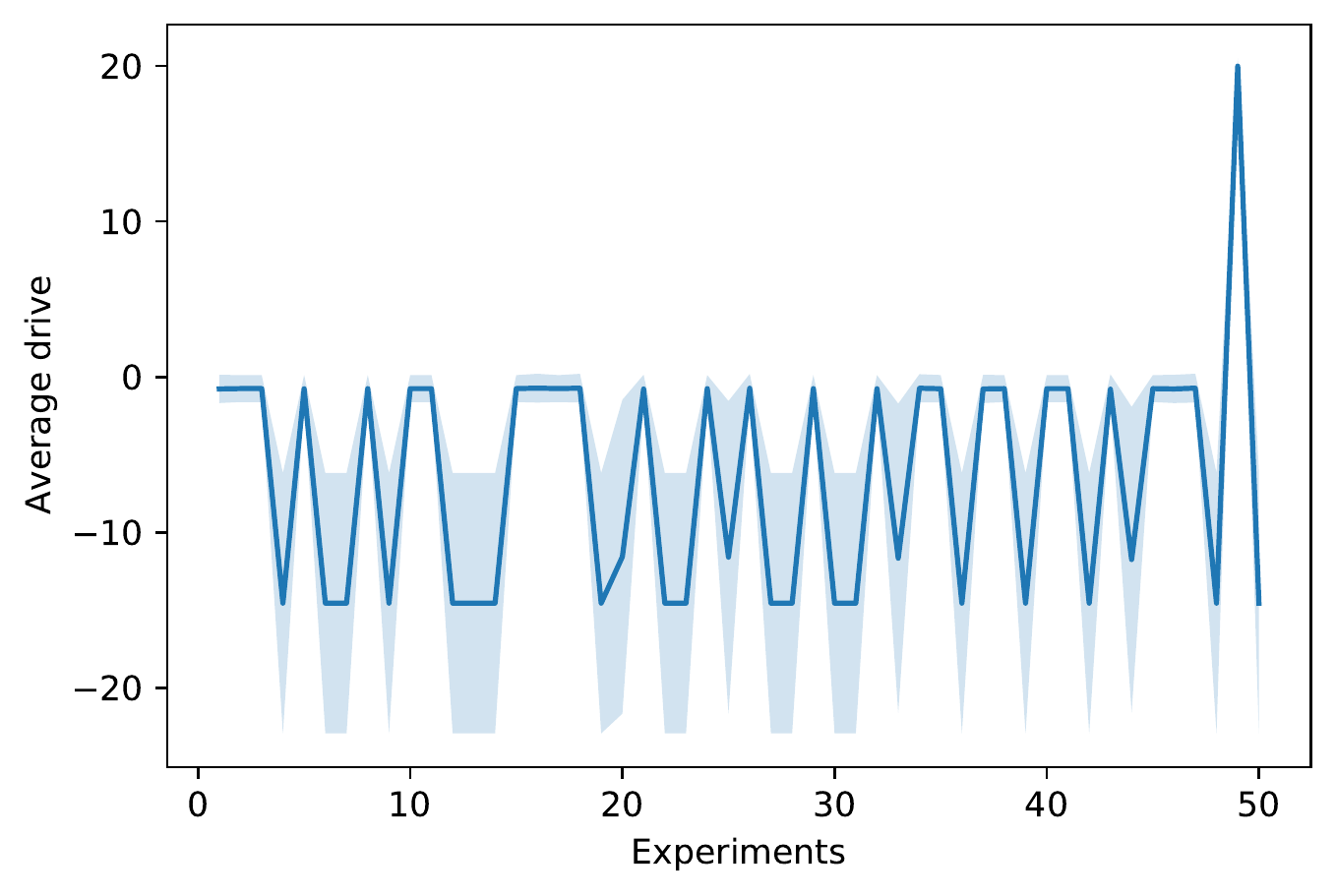}
    \includegraphics[width=0.3\textwidth]{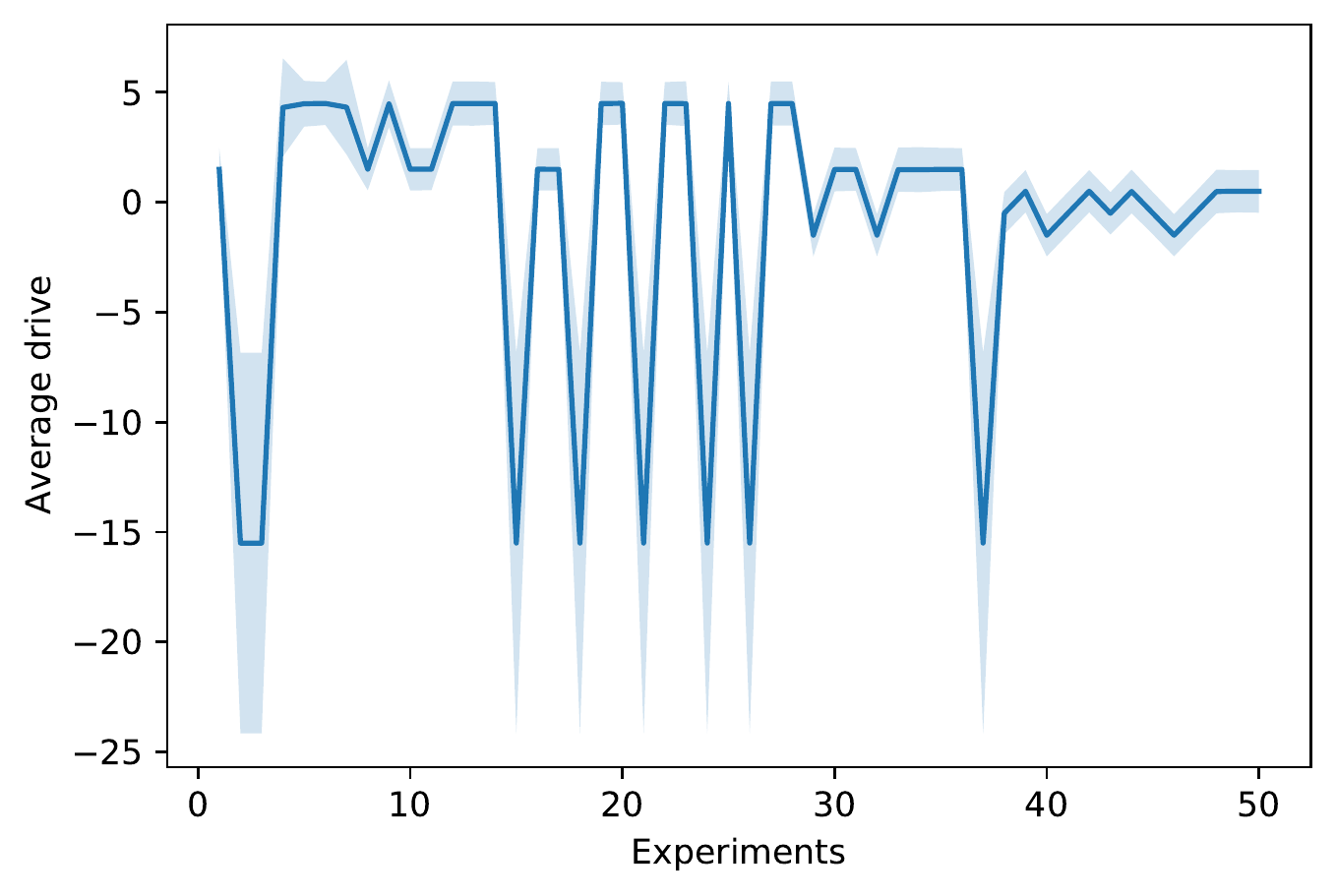}
    \includegraphics[width=0.3\textwidth]{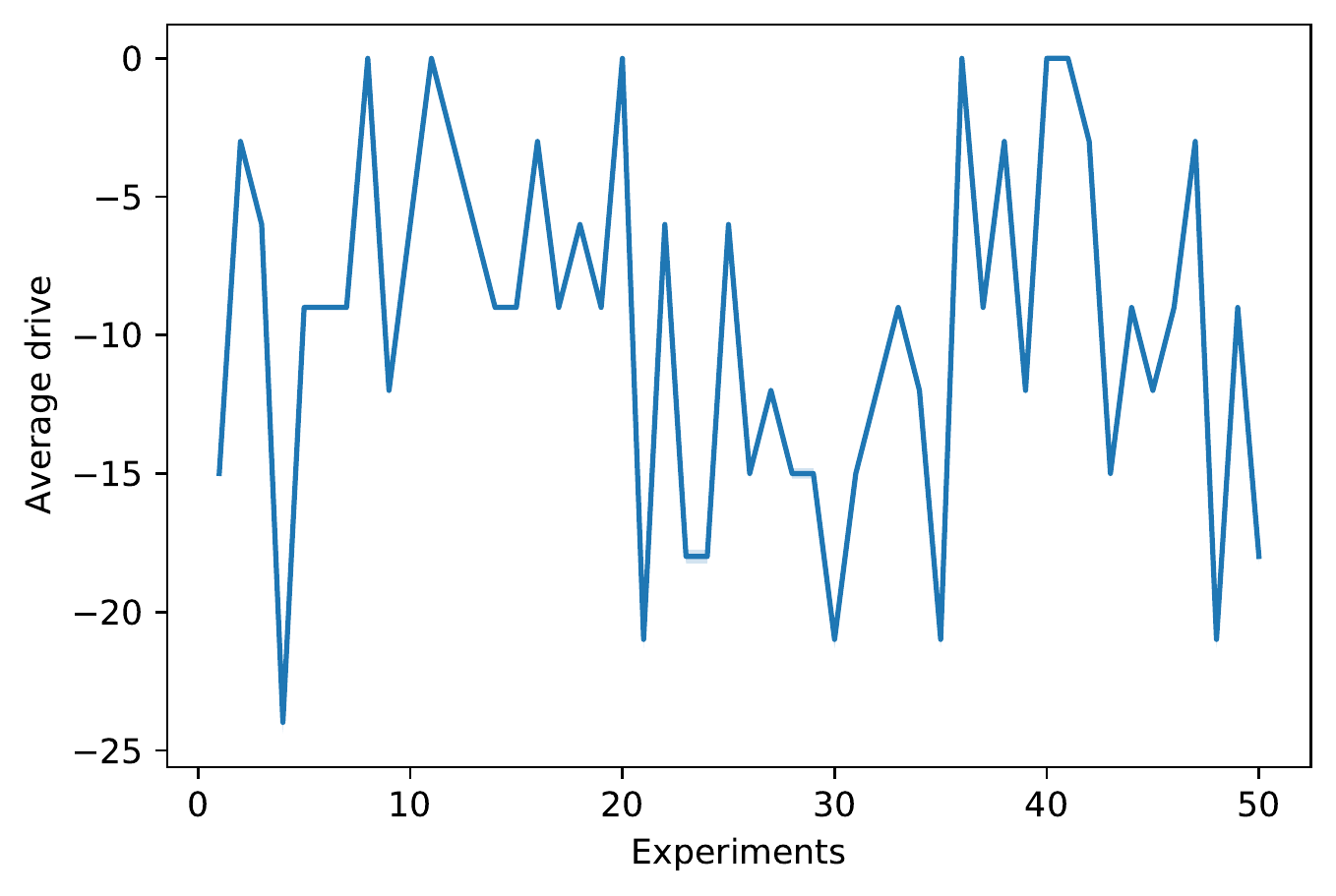}
    \caption{From left to right: EXP01 (slow metabolism), EXP02 (regular metabolism) and EXP03 (fast metabolism). From top to bottom: 1) average reward (top) and number of actions (bottom) per episode; 2) visits per environment position during training; 3) Average Drive Level in 50 tests when testing policies learned.}
    \label{fig:EXP01_EXP02_EXP03}
\end{figure*}

Analyzing first the average reward during the learning phase (illustrated in Figure \ref{fig:EXP01_EXP02_EXP03} first row), we can see that it converges for all agents, indicating they learned that staying in homeostasis (or close to it) is the expected goal. In EXP01, as the number of steps before and after episode $13,000$ is similar, this reward gain indicates that the agent learned to approach homeostasis from episode $13,000$ onwards. In EXP03, the agent faces the most challenging learning scenario because its energy consumption per step is very high. It is impossible to reach the charging station on time, depending on the agent's location, which makes the rewards very unstable. The number of actions remains low during practically the entire training. Reward and lifetime only improve at the end of the training phase due to the $\epsilon$-greedy policy employed. 

Regarding the strategies adopted during the training phase, each agent developed a specific one that fits best to their metabolism, which we can see in the second row of Figure \ref{fig:EXP01_EXP02_EXP03}. In EXP01, although the metabolism of this agent is slow, allowing it to explore the environment, it stays close to the stations to maximize the reward using the strategy of charging the battery as fast as possible when there is an imbalance in homeostasis. In EXP02, the agent remains closer to the stations or inside them. In EXP03, the agent stays inside the power stations as long as possible. After all, given its high battery consumption, it is risky to walk around the environment. Therefore, the agent remains closer or inside stations as metabolism consumption increases.

To validate the policy learned, we have the average drive obtained during the test phase illustrated in the third row of Figure  \ref{fig:EXP01_EXP02_EXP03}. In EXP01, the agent remains in homeostasis or below it (but not far) for most experiments. On the other hand, the regular metabolism agent (EXP02), that needs more energy than the agent in EXP01, maintains itself above (but close to) homeostasis, given that it is hazardous to stay with negative drives with this metabolism. Finally, EXP03 shows that it is harder to stay in homeostasis with a fast metabolism, and the typical average drive is negative in an environment where the maximum recharge value provided is the same for consumption.

Thus, in this scenario, where all stations provide the same recharge value, the slower the metabolism, the easier it is to control the drive and maximize the agent lifetime.

\subsection{Different Recharge \small{(EXP04 - EXP06)}}

In the experiments described next, all recharge areas provide different recharge values (as defined in Tables \ref{tab:rechargeValues} and \ref{tab:experiments}).

\begin{figure*}
    \centering
    \includegraphics[width=0.3\textwidth]{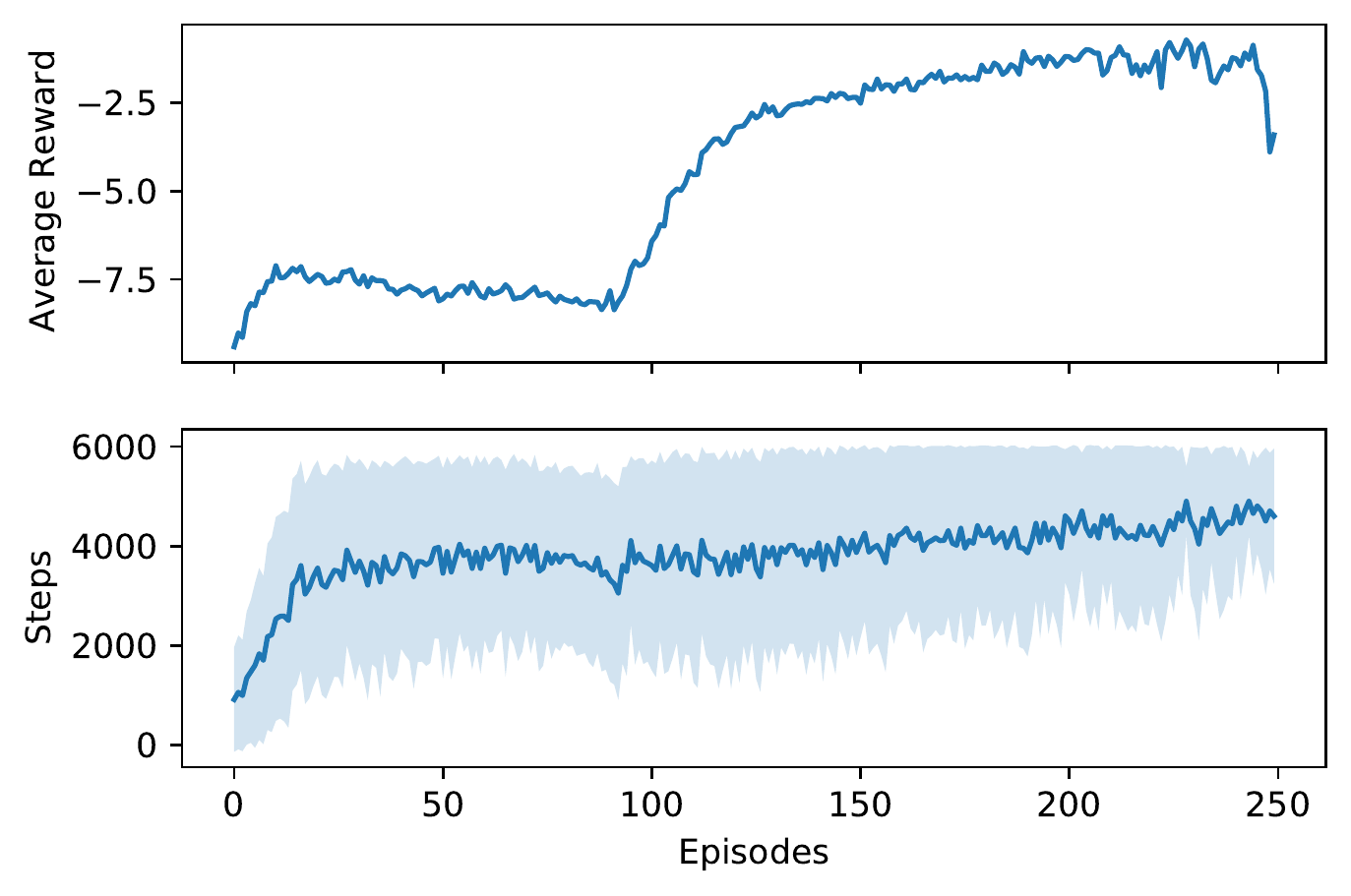}
    \includegraphics[width=0.3\textwidth]{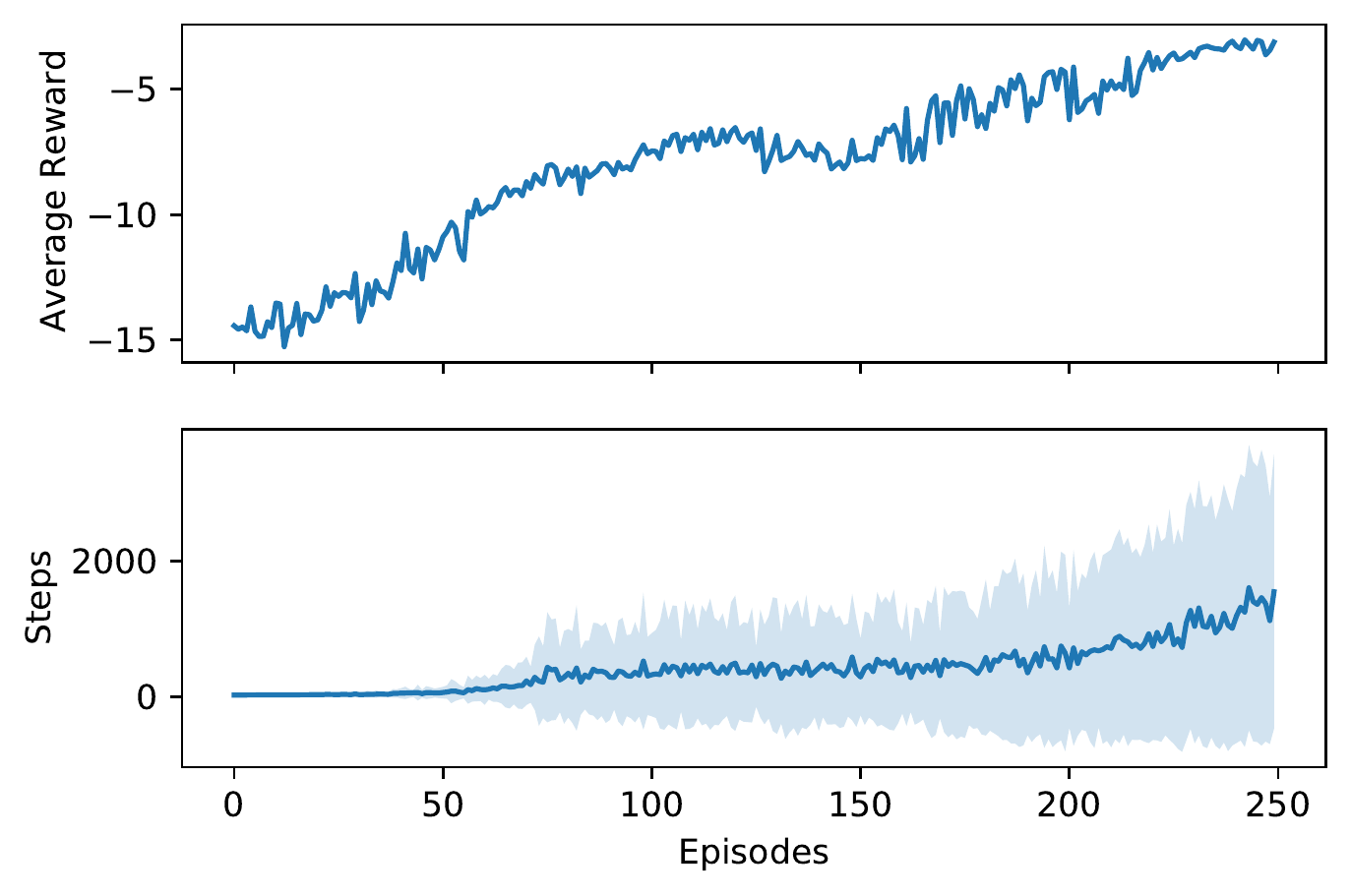}
    \includegraphics[width=0.3\textwidth]{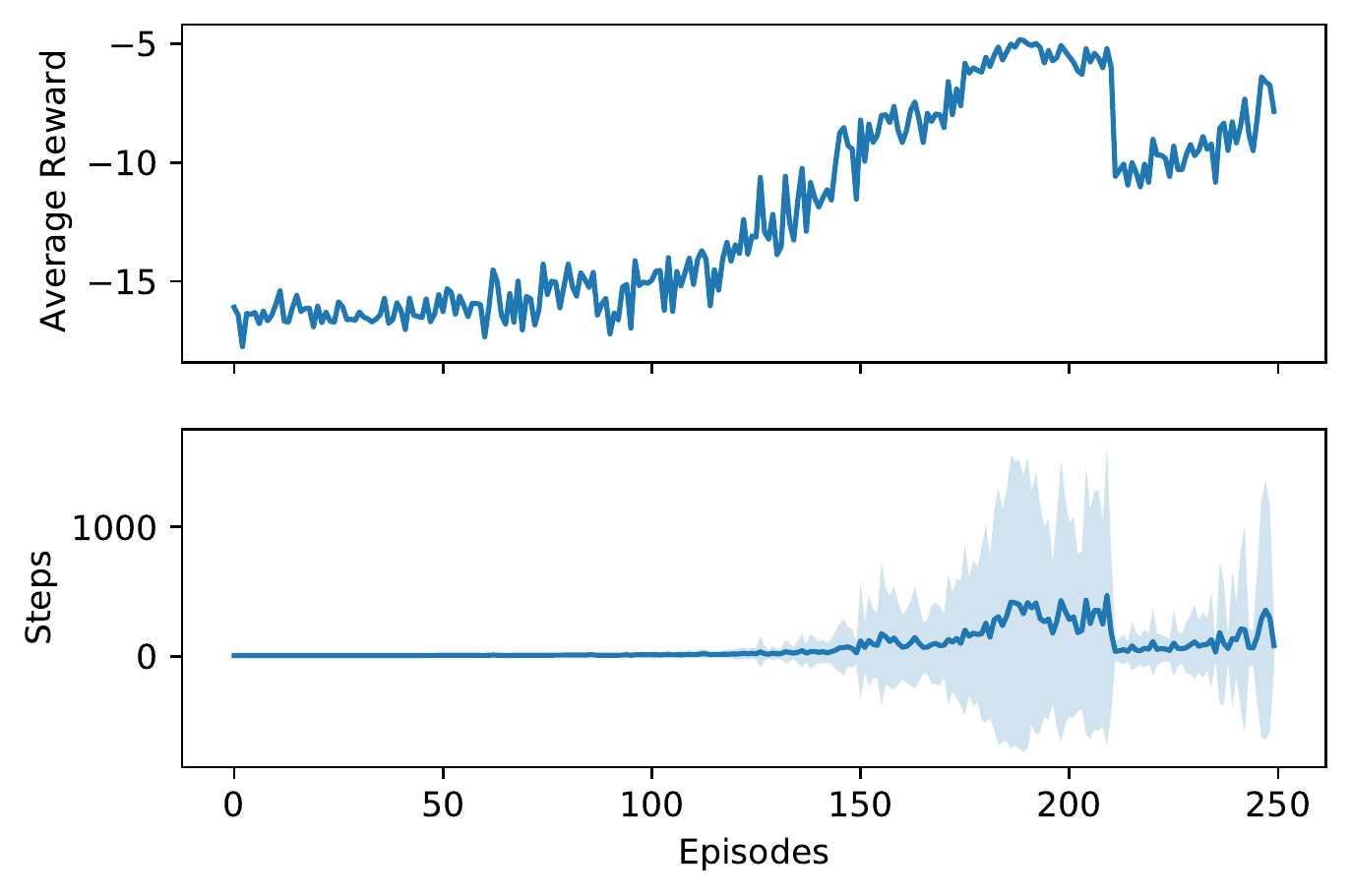}
    \includegraphics[width=0.3\textwidth]{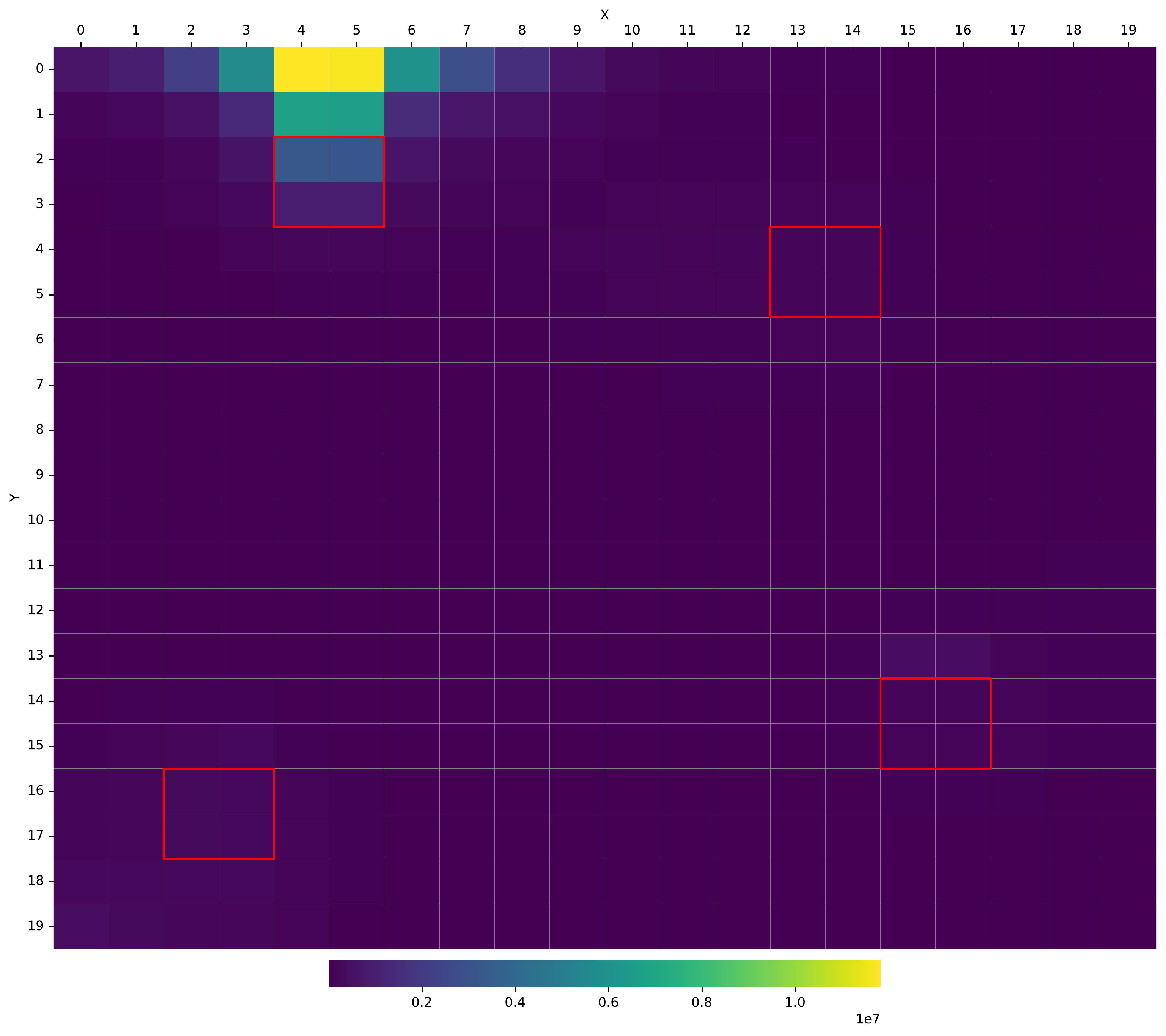}
    \includegraphics[width=0.3\textwidth]{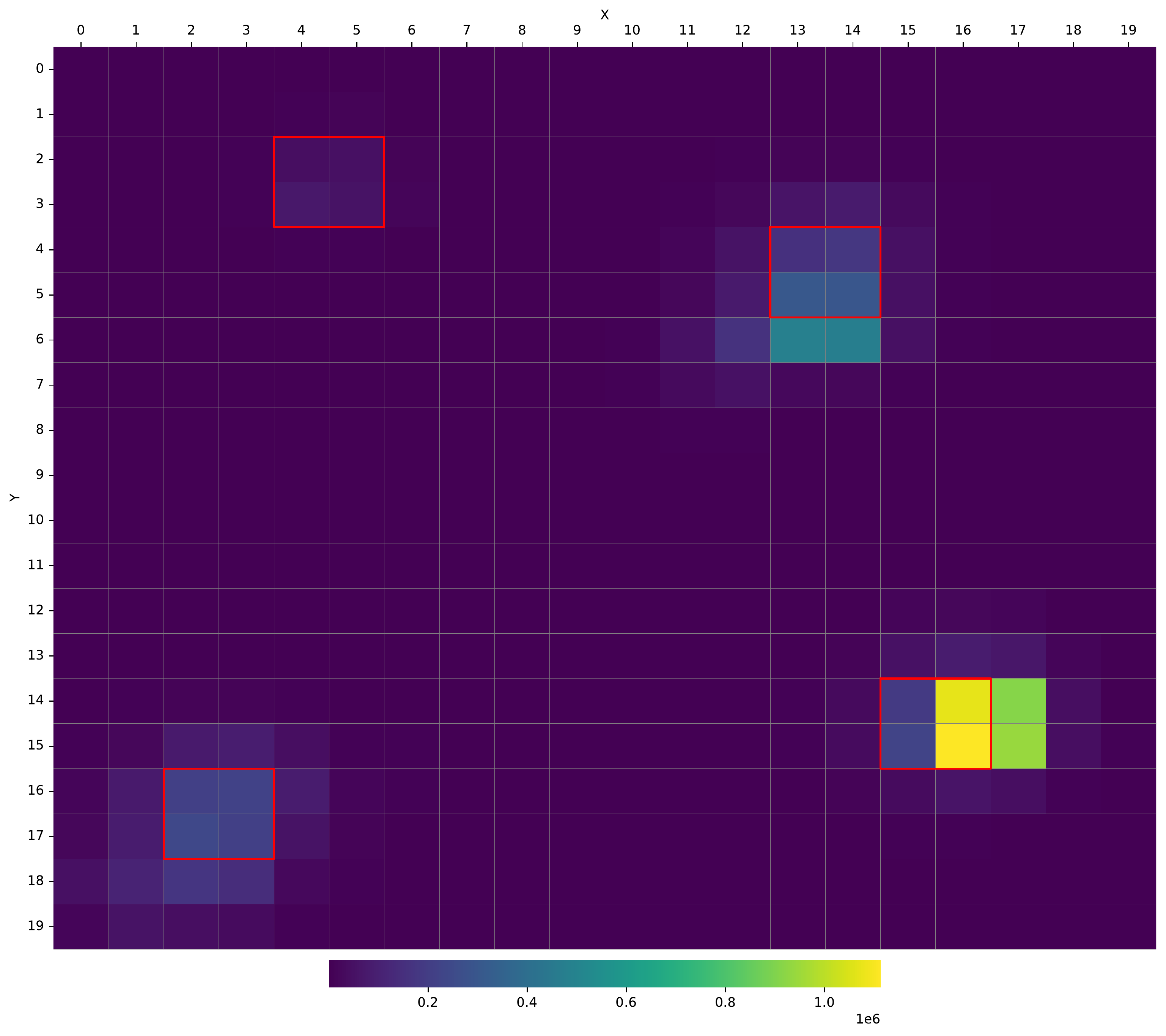}
    \includegraphics[width=0.3\textwidth]{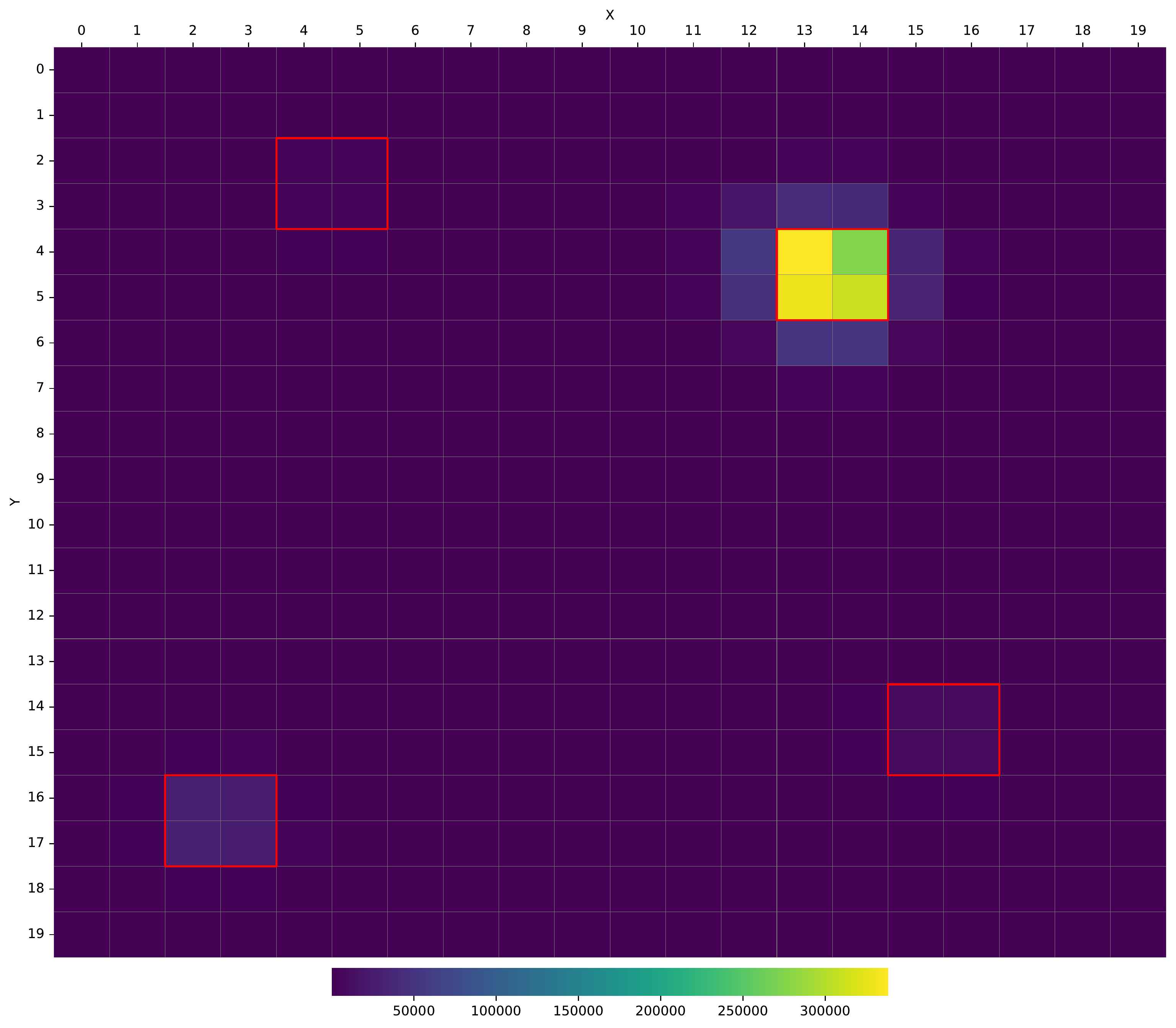}
    \includegraphics[width=0.3\textwidth]{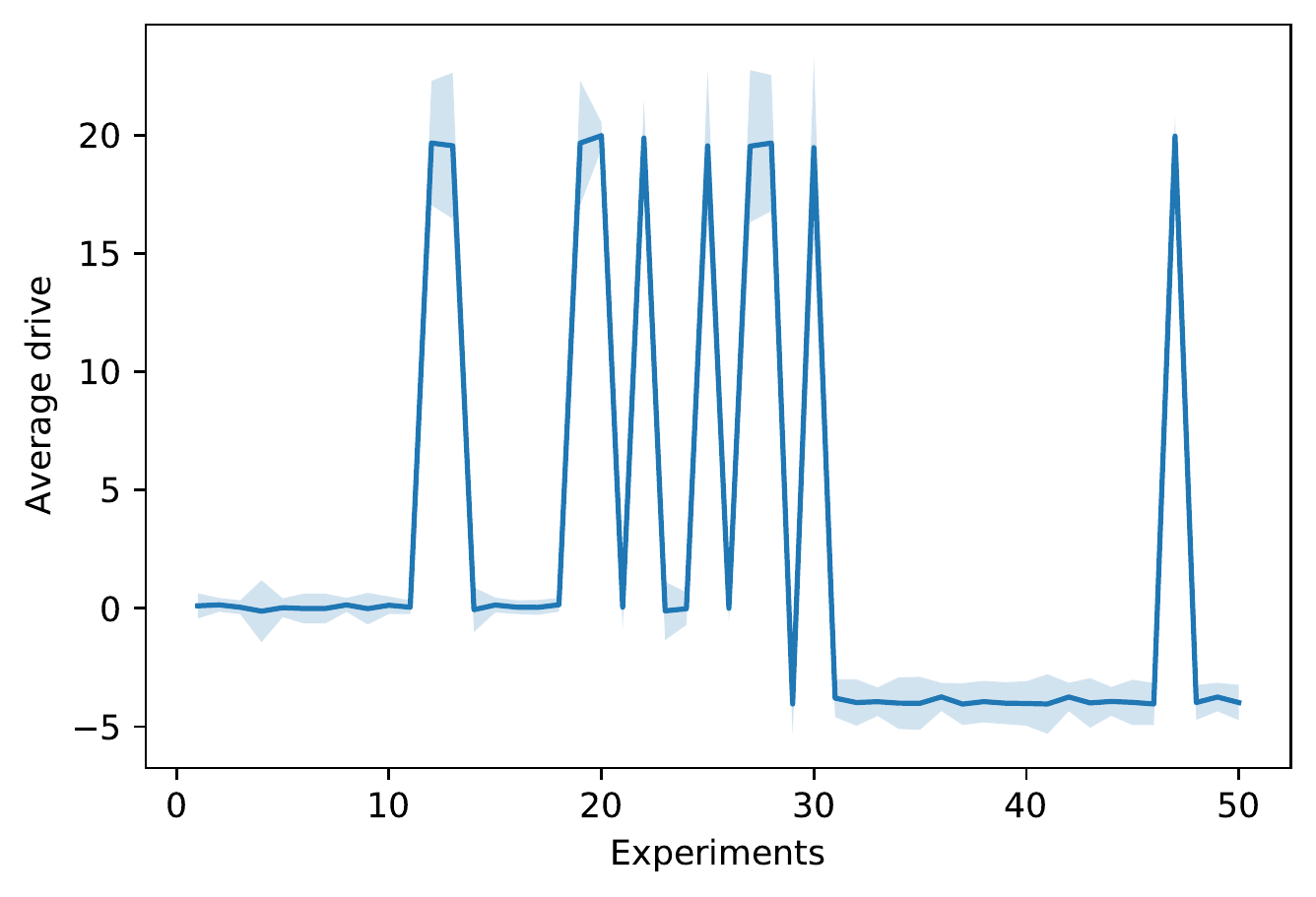}
    \includegraphics[width=0.3\textwidth]{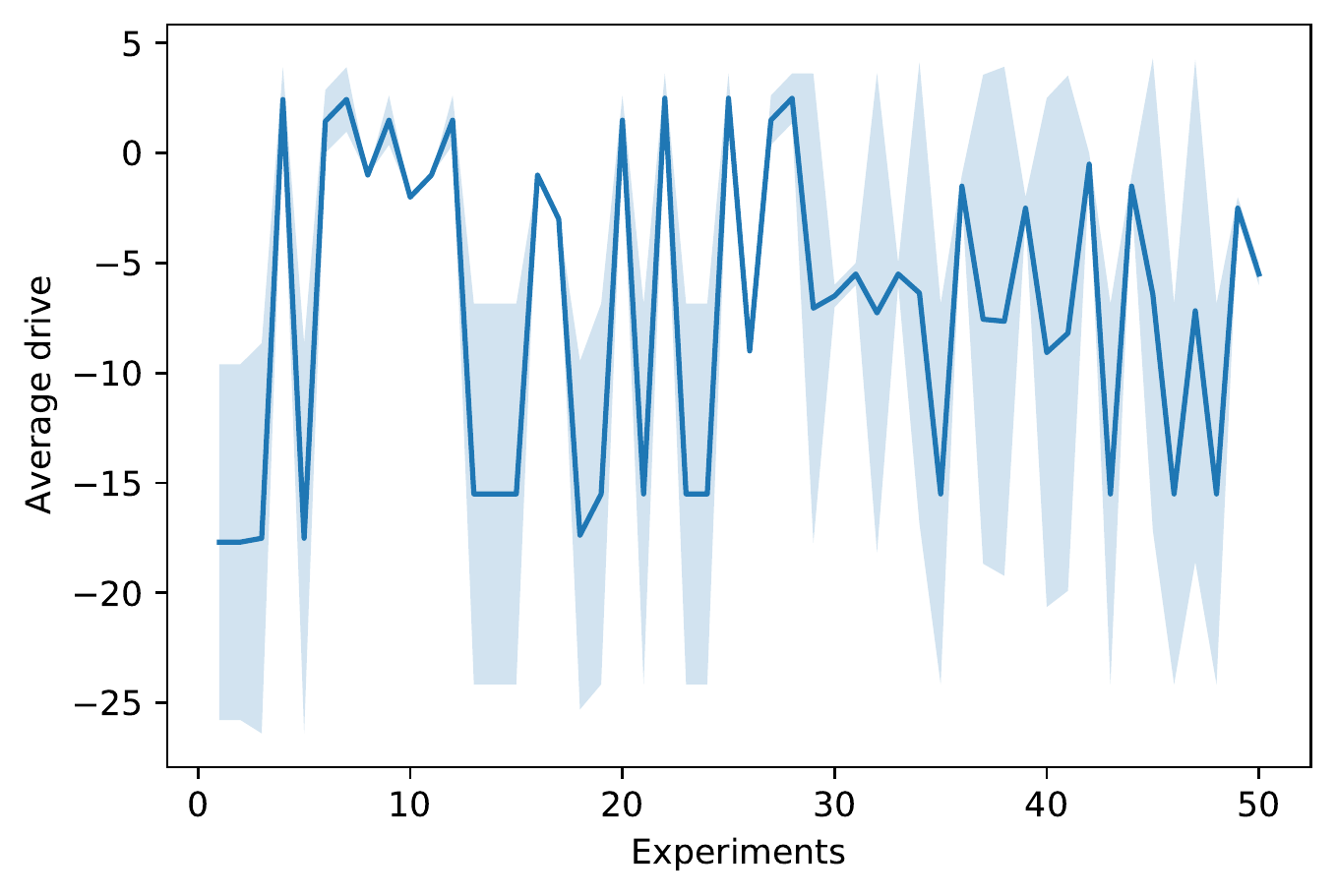}
    \includegraphics[width=0.3\textwidth]{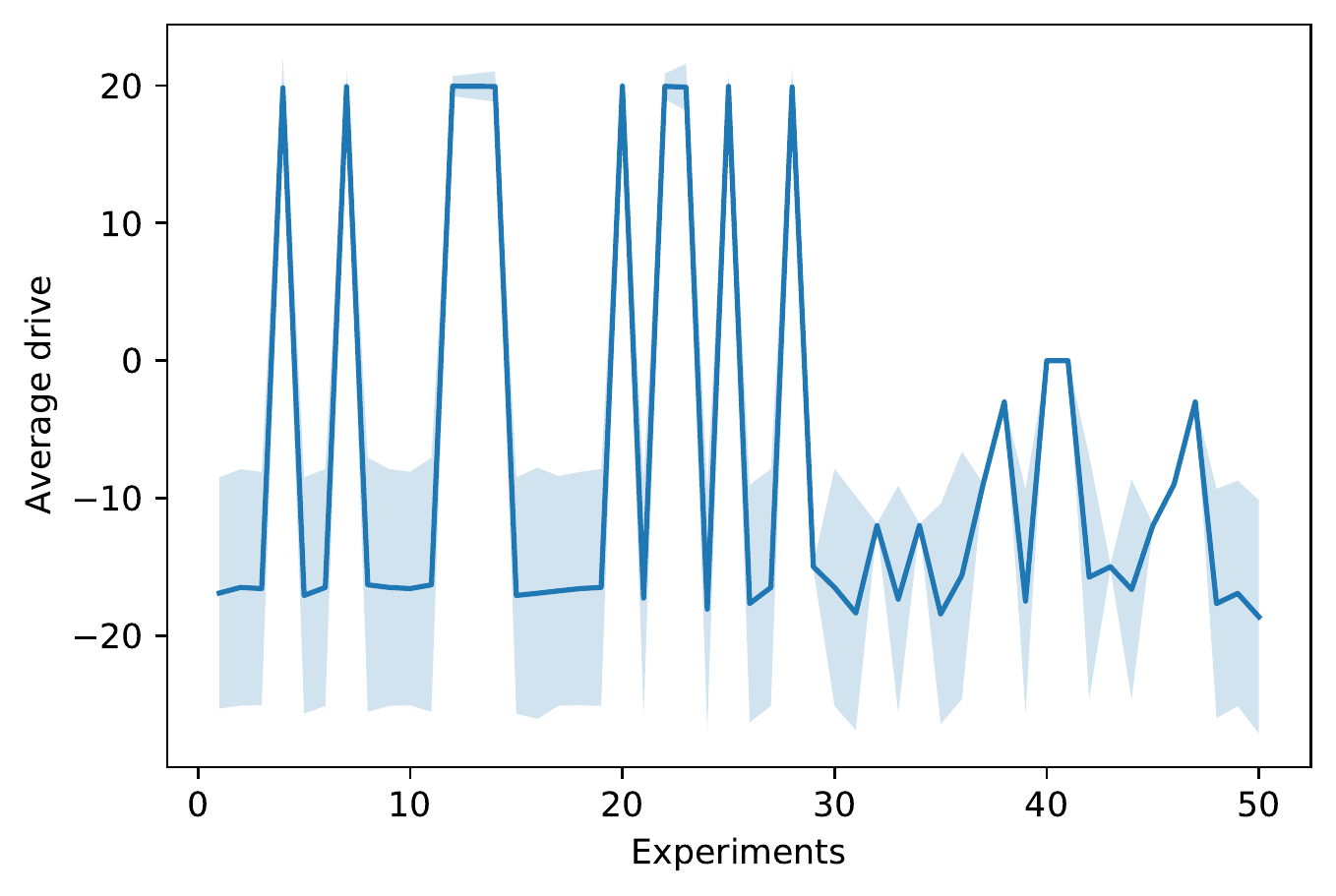}
    \caption{From left to right: EXP04 (slow metabolism), EXP05 (regular metabolism) and EXP06 (fast metabolism). From top to bottom: 1) average reward (top) and number of actions (bottom) per episode; 2) visits per environment position during training; 3) Average Drive Level in 50 tests when testing policies learned.}
    \label{fig:EXP04_EXP05_EXP06}
\end{figure*}

Considering the average reward illustrated in Figure \ref{fig:EXP04_EXP05_EXP06} first row, all agents learned to balance the drive by trying to stay in (or close enough) homeostasis during the learning phase. Comparing with the previous experiments, we can see that the rewards in EXP04 and EXP05 are similar to their equivalent metabolism in EXP01 and EXP02, respectively. However, in this scenario, the slow metabolism learns faster: in EXP01, the reward increases around the episode $13,000$, while in EXP04, it starts at around $10,000$. Regarding the fast metabolism in EXP06, the reward decreases slightly towards the end. Still, this decay may be due to low battery starting and far from stations (or even close to a station unfavorable for its \textit{needs}). But in EXP06, the rewards are generally higher and more stable than the equivalent EXP03. 

These improvements observed between equivalents agents considering two different environments occur due to the environment's settings that now provide better options to satisfy the need of each agent. There is a station with a low value that fits best for slow metabolism, one with medium values for regular metabolism, and one with a higher value that is better for fast metabolism.

Concerning the behavior in the environment, the agents learned about all the stations, choosing to stay closer to the ones that fit better their needs, as shown in the second row of Figure \ref{fig:EXP04_EXP05_EXP06}. In EXP04, the agent spent more time near station $A$, which provides the lowest recharge value ($\delta D_1{_{P_A}}/\delta t = 1$). As the agent's metabolism is slow, this station is good enough, and the agent does not risk getting over homeostasis quickly. In EXP05, the agent visits the stations with the highest recharges, staying longer at station $D$ ($\delta D_{1_{P_D}}/\delta t = 2$), which matches best their regular metabolism. For the fast metabolism (EXP06), the best strategy is to stay inside the station $B$ that provides the highest recharge value ($\delta D_{1_{P_B}}/\delta t = 4$), which is what the agent learned. 

When testing the policy learned (last row of Figure \ref{fig:EXP04_EXP05_EXP06}), we can note that the robot behavior is strictly related to the agent's initial position in the environment. In EXP04, the agent reaches homeostasis when initiated close to stations with low recharge values. However, it gets the maximum or negative drive when it is close to stations that offer higher values than the necessary for this metabolism. So the strategy learned varies between complete recharging and waiting for the natural consumption or waiting to be negative and then recharge. The same occurs in EXP05 and EXP06, but due to the metabolism consumption is more challenging to stay in homeostasis depending on the initialization position and drives urge.

To summarize the findings obtained in this first set of experiments using only the M1 mechanism (EXP01 - EXP06), we can note that each individual learned the best strategy to balance their drives. Also, we validated that the environment's setting impacts the motivational-learning process, given that when the environment offers options that fit better with our needs, the agent can learn faster and behave more efficiently.

\section{M2 experiment set - \textit{Wanting} + \textit{Liking} mechanisms} \label{sec:expsDrivePleasure}

For this set, we executed the same experiments as the previous one where we assess how intrinsic characteristics of an agent, like its metabolic curve, can influence behavior selection and the influence of the environment in agents with similar characteristics. However, we add a pleasure value to each station. This new configuration implies a more complex scenario that corresponds to the \textit{want x like} dichotomy.

In this experiment set we employ the second motivational model \textbf{M2}, modeled via the reward function $R_2$ (Eq. \ref{eq:rewardDrivePleasure}), in which the agent's goal is to maintain homeostasis while getting pleasure. 

\subsection{Same Recharge \small{(EXP07 - EXP09)}}
This set of experiments is equivalent to EXP01 - EXP03, considering the agent's metabolism and the recharge values (all recharge areas provide the same value). However, now each station provides a pleasure value (as defined in Table \ref{tab:rechargeValues}). Then, we based our analysis on comparing these results with the EXP01 - EXP03 ones.

\begin{figure*}
    \centering
    \includegraphics[width=0.3\textwidth]{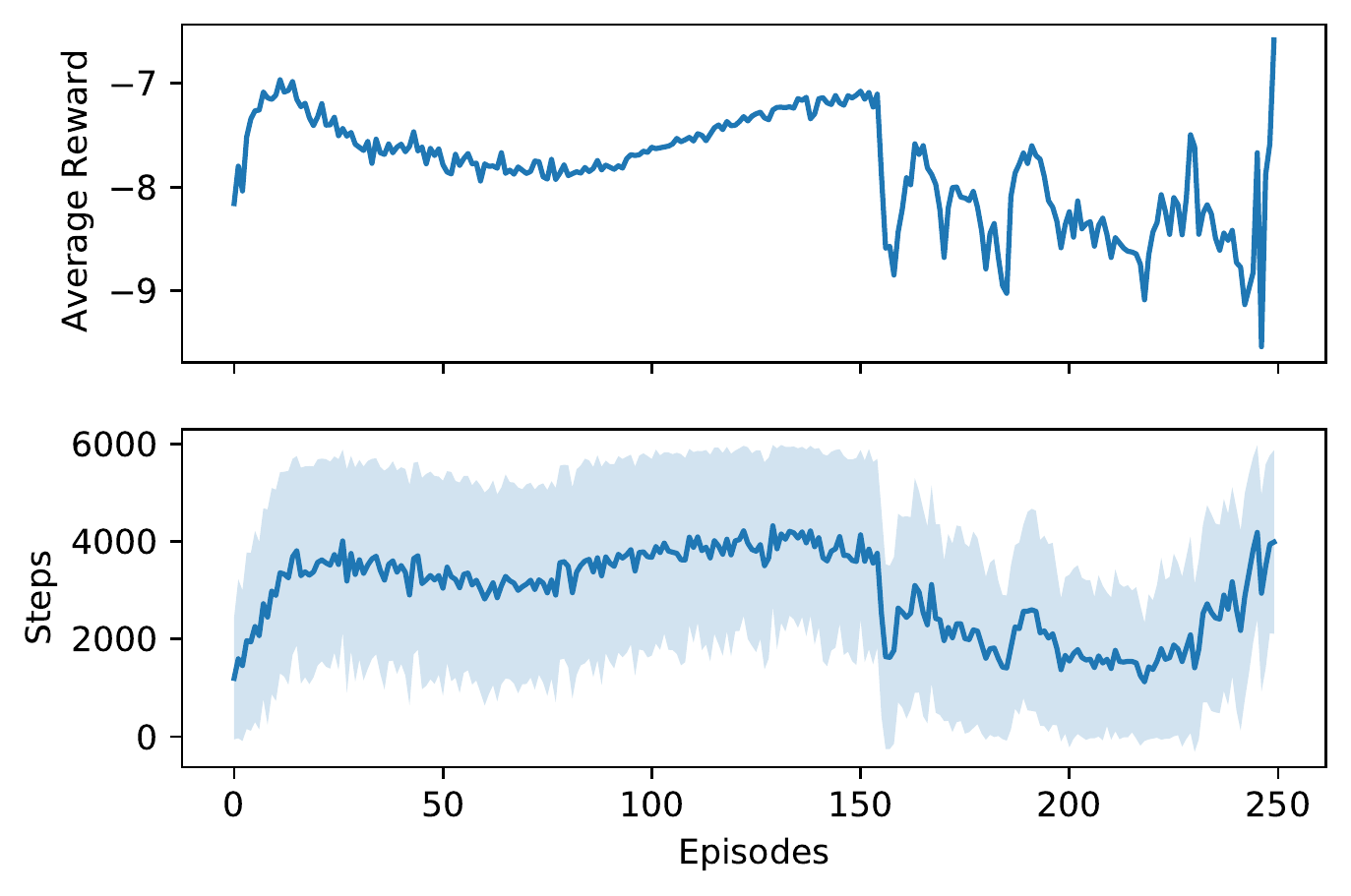}
    \includegraphics[width=0.3\textwidth]{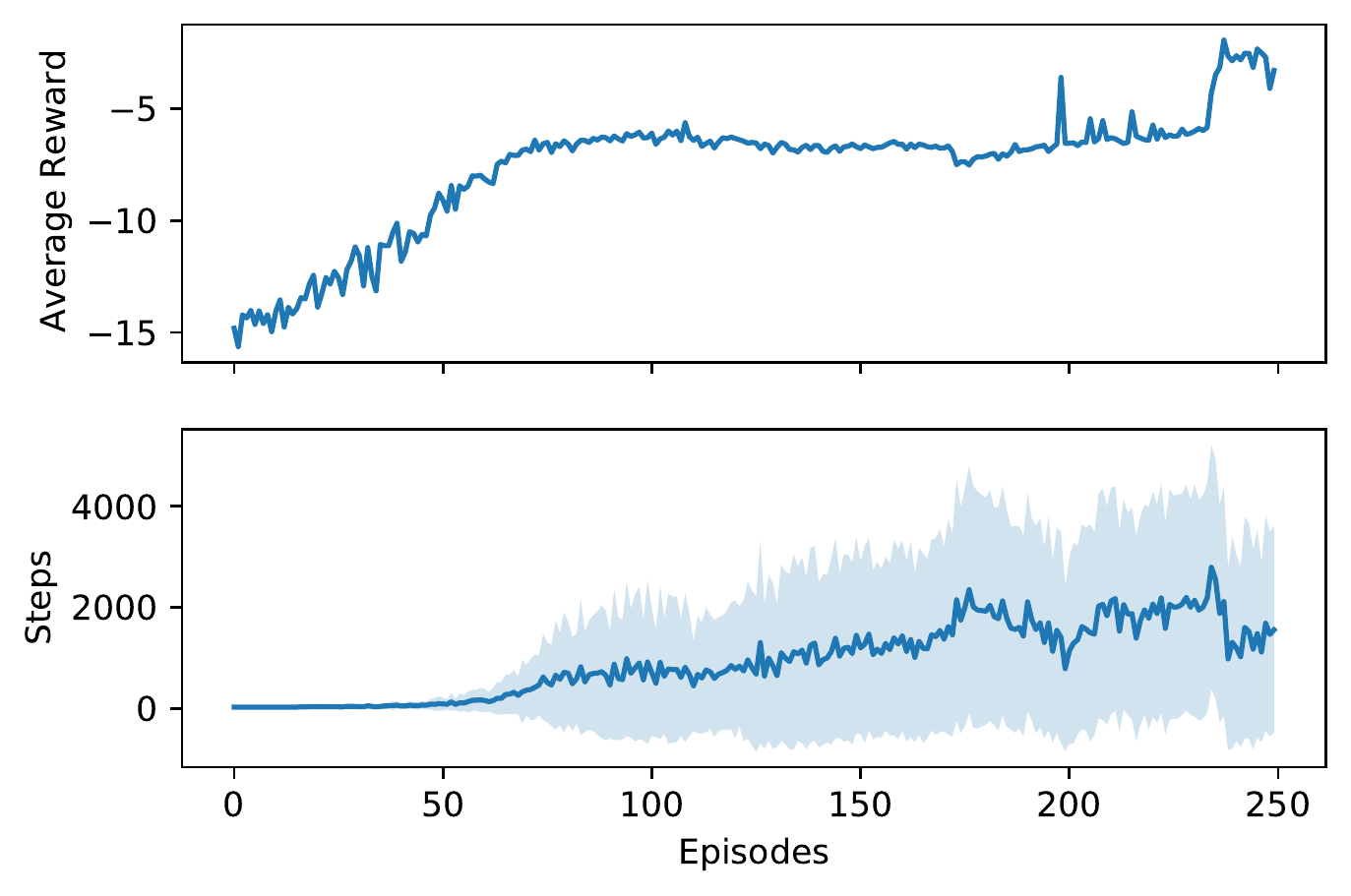}
    \includegraphics[width=0.3\textwidth]{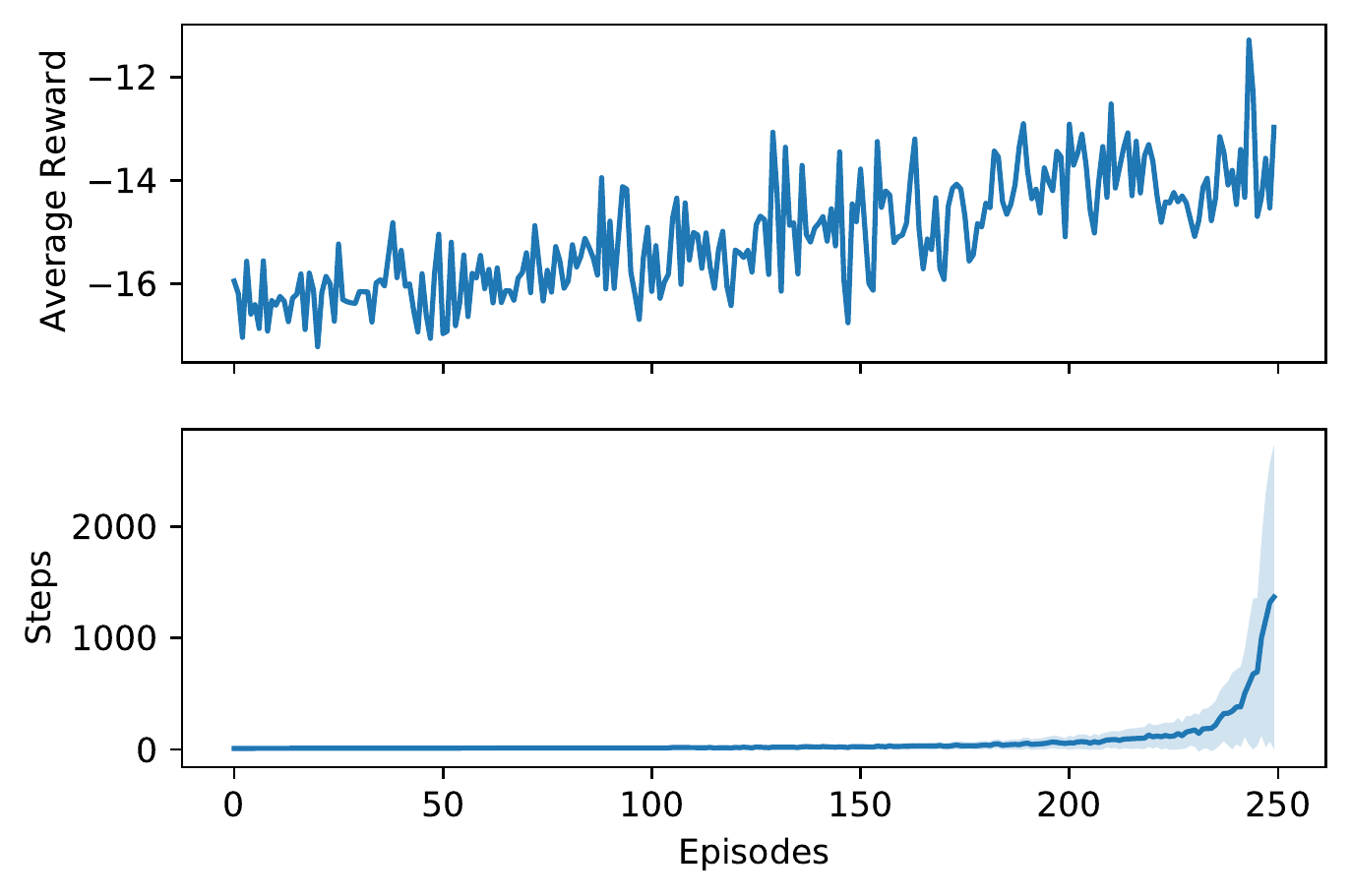}
    \includegraphics[width=0.3\textwidth]{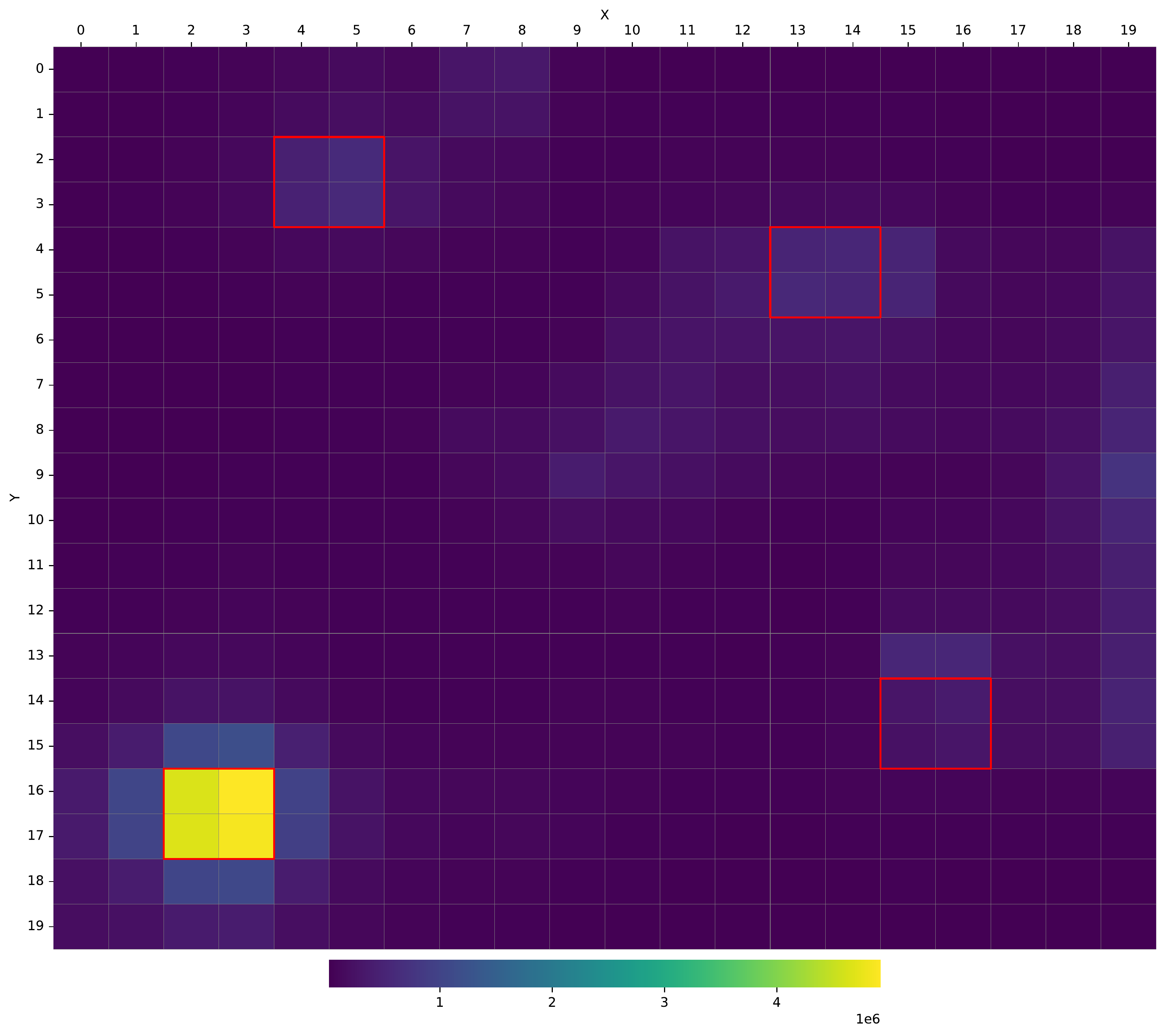}
    \includegraphics[width=0.3\textwidth]{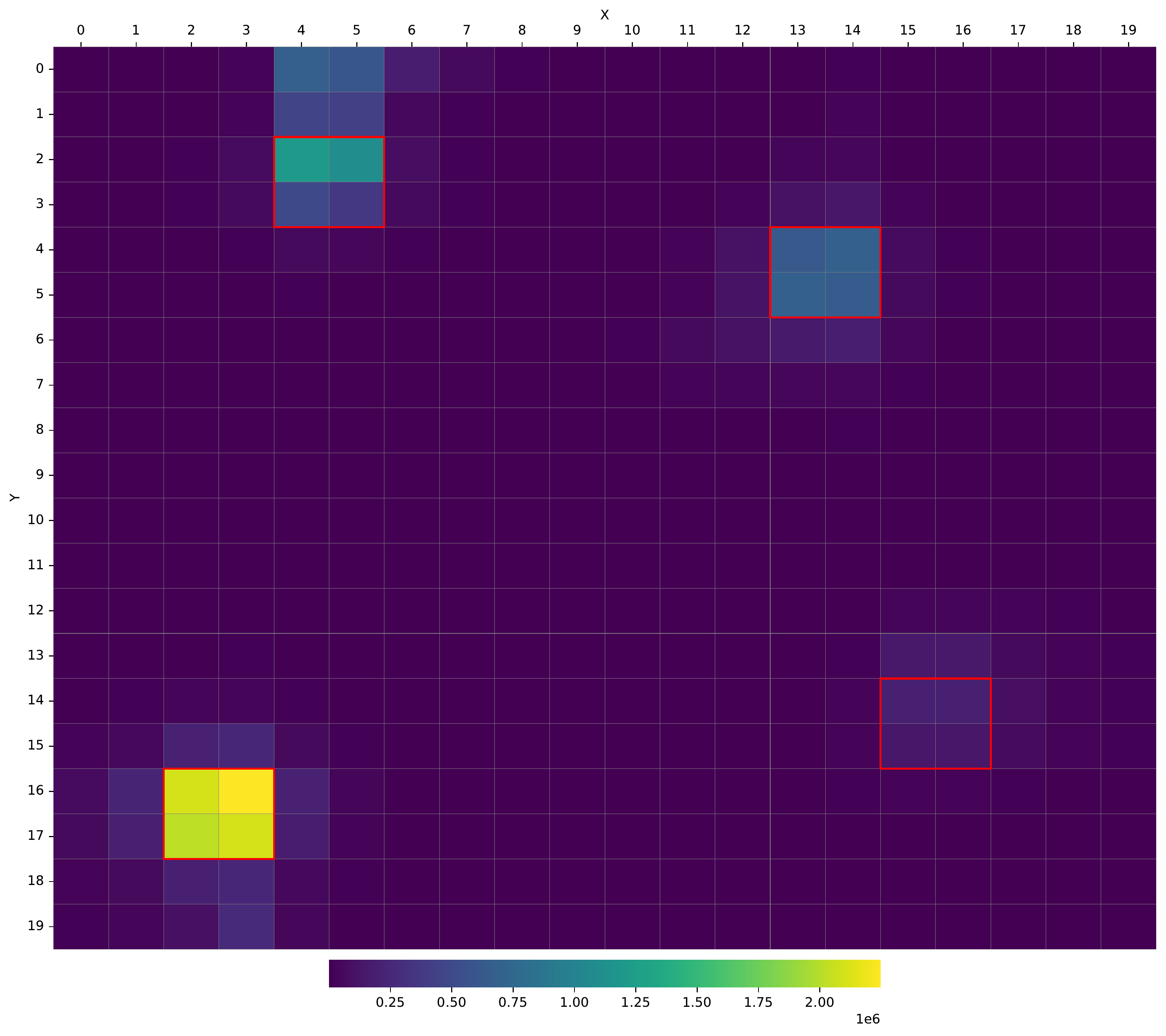}
    \includegraphics[width=0.3\textwidth]{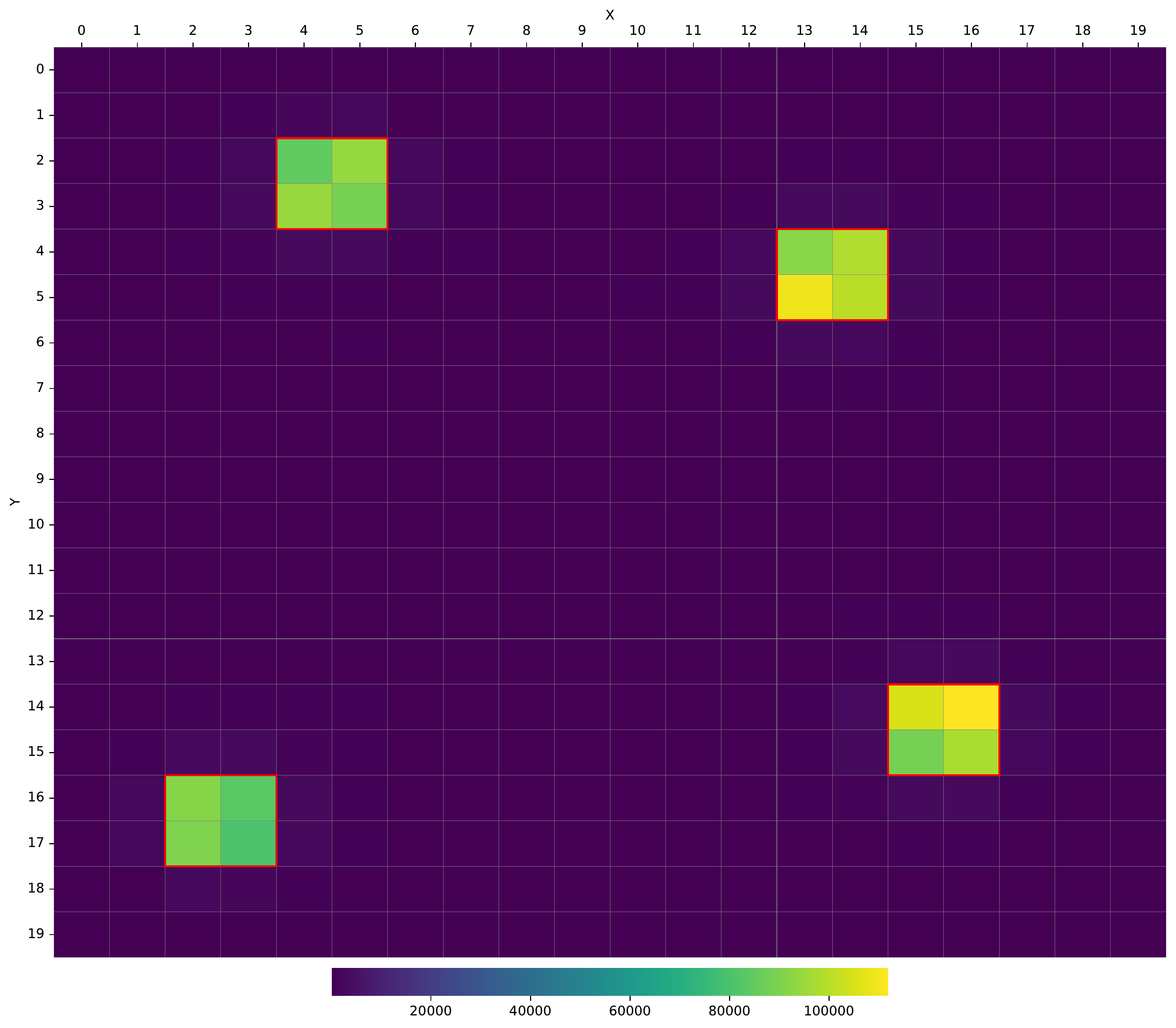}
    \includegraphics[width=0.3\textwidth]{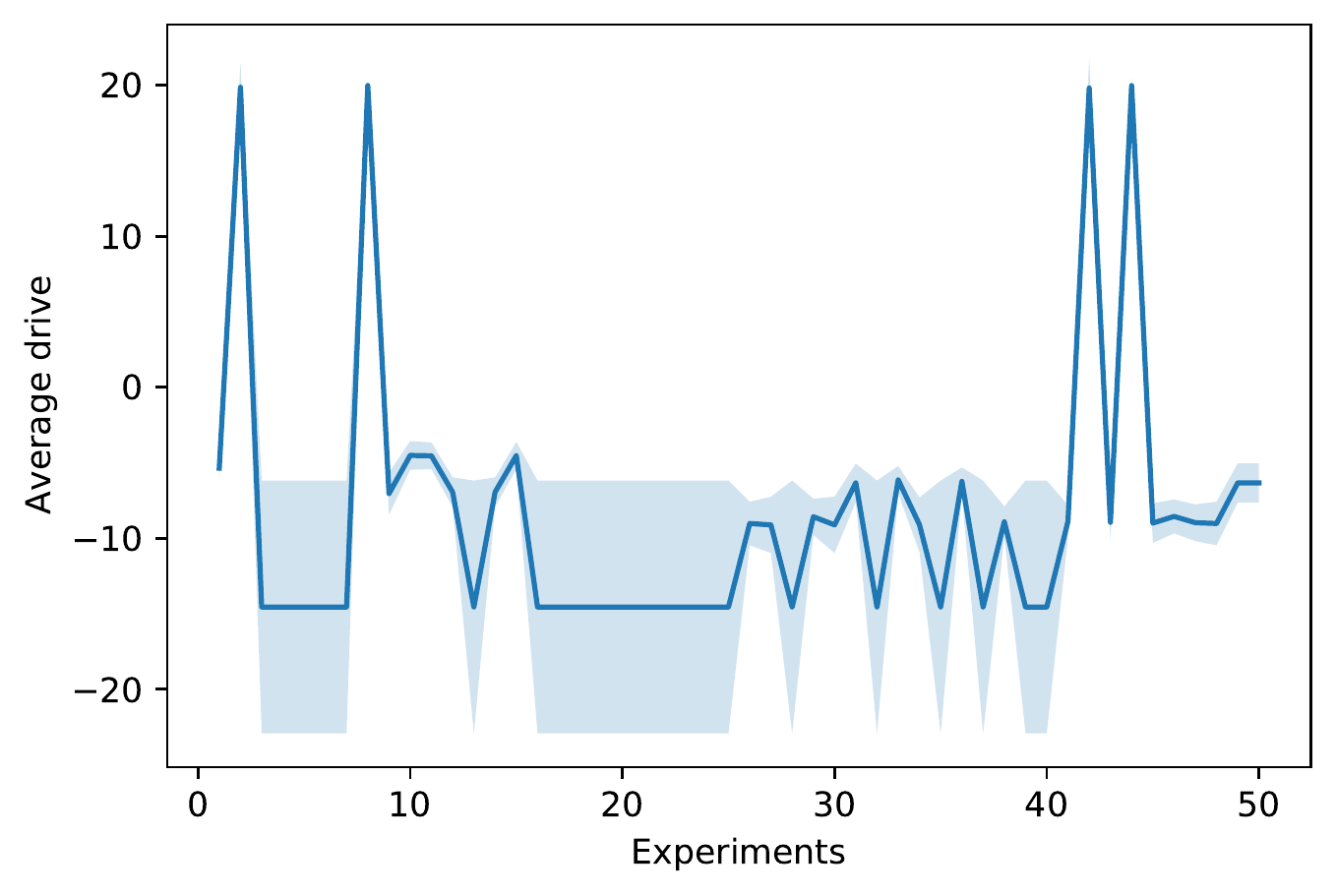}
    \includegraphics[width=0.3\textwidth]{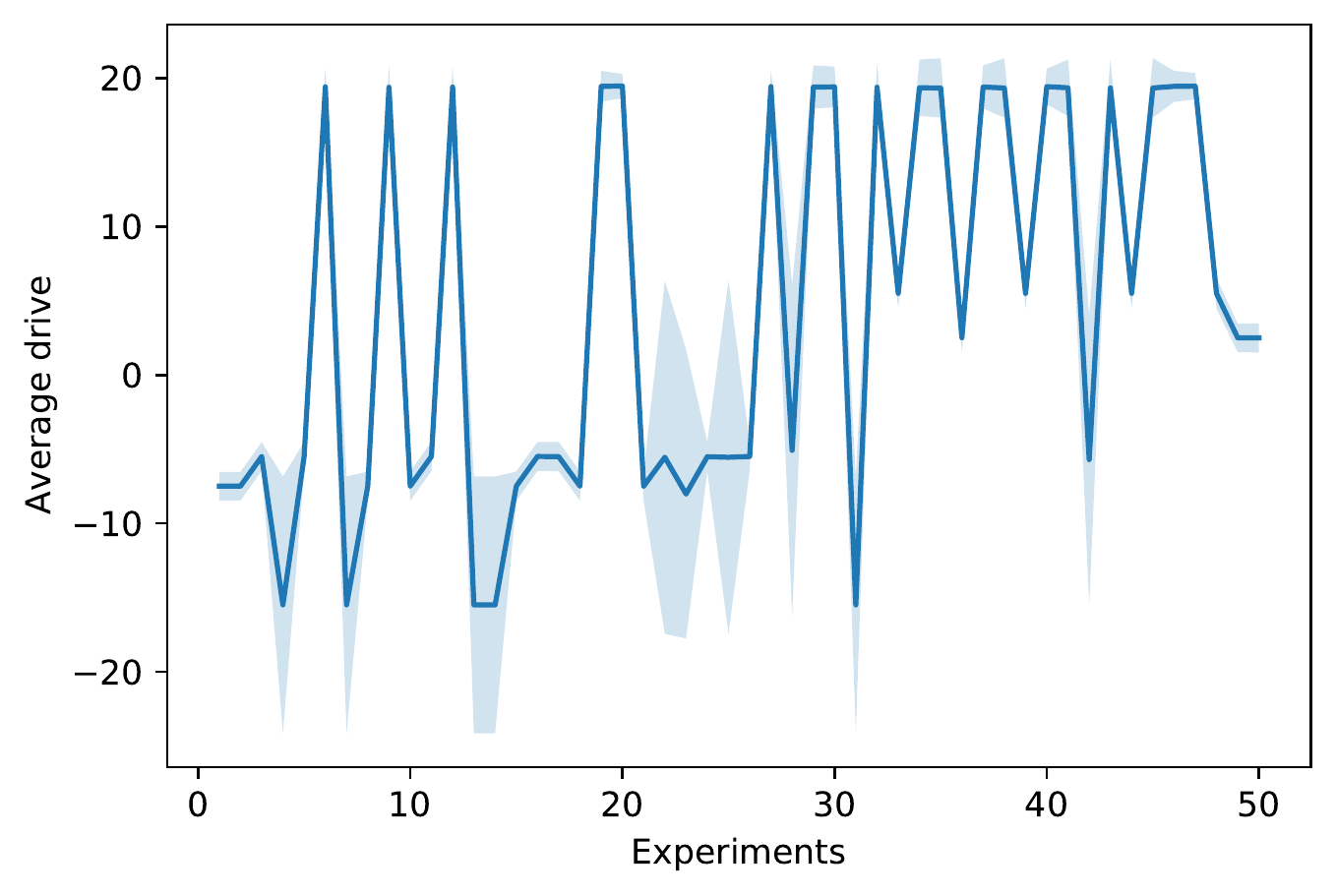}
    \includegraphics[width=0.3\textwidth]{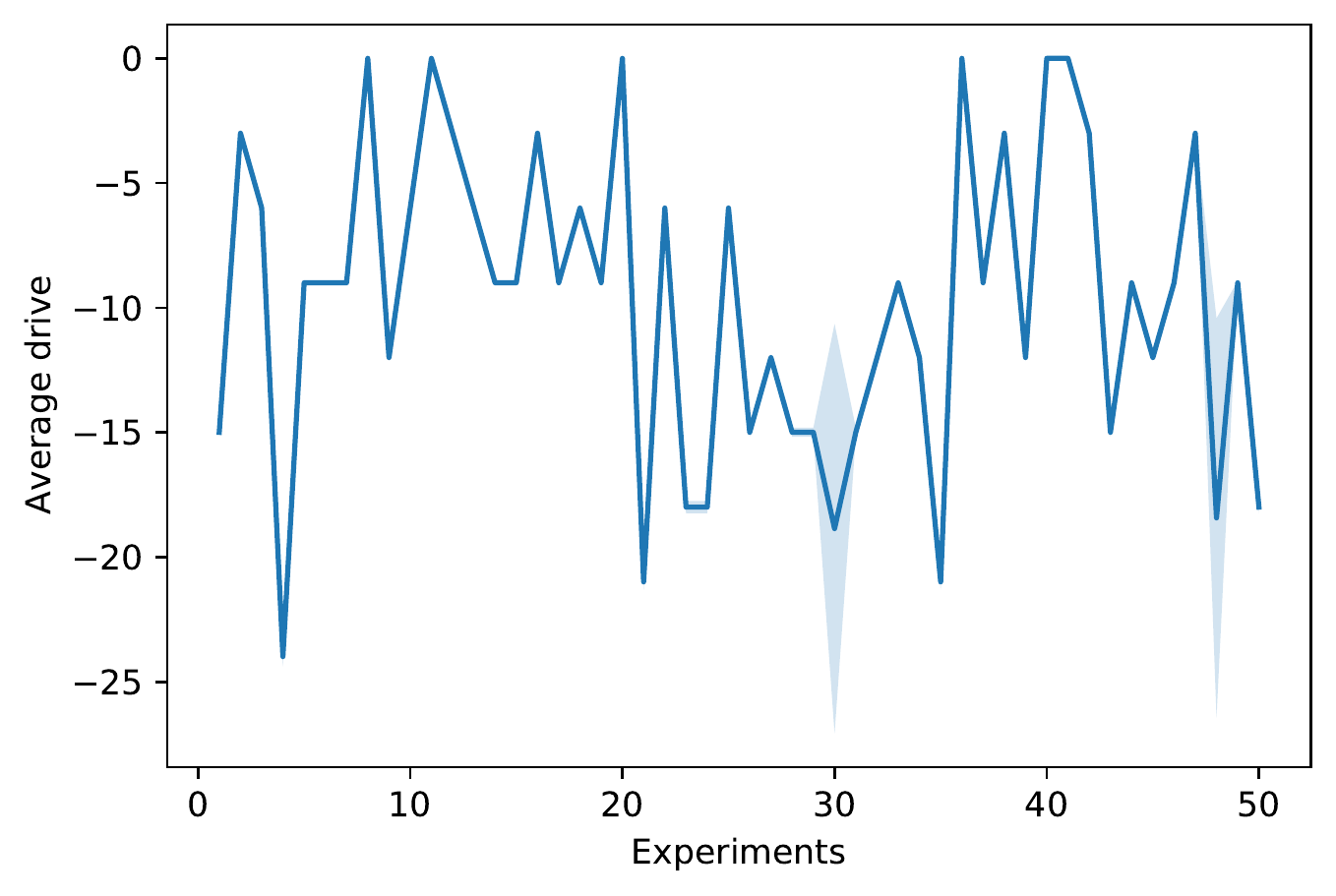}
    \caption{From left to right: EXP07 (slow metabolism), EXP08 (regular metabolism) and EXP09 (fast metabolism). From top to bottom: 1) average reward (top) and number of actions (bottom) per episode; 2) visits per environment position during training; 3) Average Drive Level in 50 tests when testing policies learned.}
    \label{fig:EXP07_EXP08_EXP09}
\end{figure*}

Considering first the performance in the training phase, the reward from EXP08 and EXP09 improves over time, but in EXP07, it gets worse, as seen in the first row of Figure \ref{fig:EXP07_EXP08_EXP09}. Approaching the slow metabolism first, the equivalent experiment without pleasure (EXP01) shows that the robot could satisfy the drive in that case. Notwithstanding, in EXP07, the agent prefers getting pleasure instead of benefiting the drive, which makes it get worse rewards. Regarding regular metabolism, EXP08 and EXP02 have similar behavior in the training phase, but in EXP08, the agent shows a plateau in reward between $5,000$ and almost $24,000$, changing only the number of actions. It performs better than any other agent in this set (for the same recharge scenarios). Finally, the performance of the fast metabolism (EXP09) is better and more stable than in EXP03.

Still in the training phase, but evaluating now the movement behavior in the environment illustrated in Figure \ref{fig:EXP07_EXP08_EXP09}, second-row. In EXP07, although the agent does not need to recharge a lot due to its metabolism, it chooses to stay inside station $C$ most of the time. This station has the highest pleasure, proving the agent prefers getting pleasure instead of balancing the energy need (different behavior compared with EXP01, when the agent stayed outside but close to the station to recharge as soon as an imbalance is detected). In EXP08, the agent spends more time at stations $C$, $A$, and $B$, respectively, which are the ones with higher pleasure values in the same order. So this agent also prioritizes pleasure. Regarding EXP09, due to the fast energy consumption of this agent, getting energy is much more critical than having pleasure. So for this agent, the closest station is the best, independent of the pleasure value. As we can see, the behavior with (EXP09)/without(EXP01) pleasure is almost the same.

We can check the average drive obtained during the test phase (Figure \ref{fig:EXP07_EXP08_EXP09}, third row) to validate our analysis. As expected, the results of EXP07 and EXP08, which suffer from adding pleasure as we see before, are very different from their equivalent ones without pleasure (EXP01 and EXP02). In EXP01 (Figure \ref{fig:EXP01_EXP02_EXP03}, first column, last row), the agent remains in homeostasis in most experiments. However, in EXP07, the agent did not stay at homeostasis level but below it in most experiments, which is the expected behavior that gives the greater rewards (gain in direction to homeostasis + gain the pleasure). In EXP08, the agent prefers the pleasure gain instead of the homeostasis one, so it maintains a high drive value in many experiments. On the other side, in EXP09, the agent's average drive is almost the same as EXP03 (Figure \ref{fig:EXP01_EXP02_EXP03}, third column, last row), and this occurs because the agent's energy consumption is too high, which makes it prioritize its needs and not pleasure. Hence, pleasure in this kind of metabolism does not change the agent's behavior.

\subsection{Different Recharge \small{(EXP10 - EXP12)}}
This set of experiments is equivalent to EXP04 - EXP06, considering the recharge values (all recharge areas provide different values). The difference is that each station now provides a pleasure value (as defined in Table \ref{tab:rechargeValues}). Then, we based our analysis on comparing these results with the EXP04 - EXP06 ones when evaluating the motivational mechanism impact and with EXP07 - EXP09 when comparing the environment's impact.

Given that this set of experiments is the most complex one, we present the analysis in a different order from the previous ones. Instead of discussing the training results of all metabolisms first and then the testing phase, here we discuss the training phase followed by the test phase of each metabolism.

\begin{figure*}
    \centering
    \includegraphics[width=0.3\textwidth]{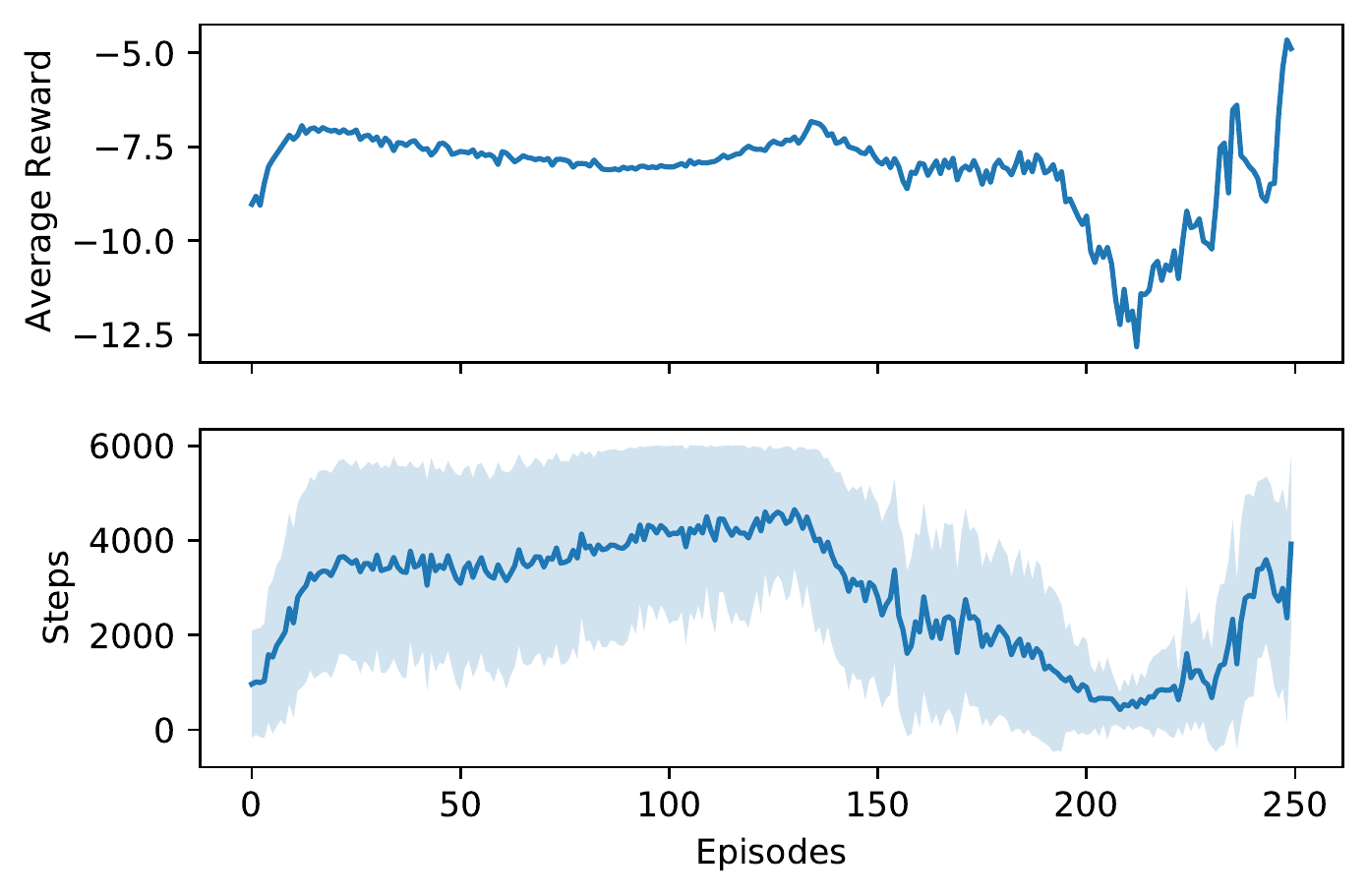}
    \includegraphics[width=0.3\textwidth]{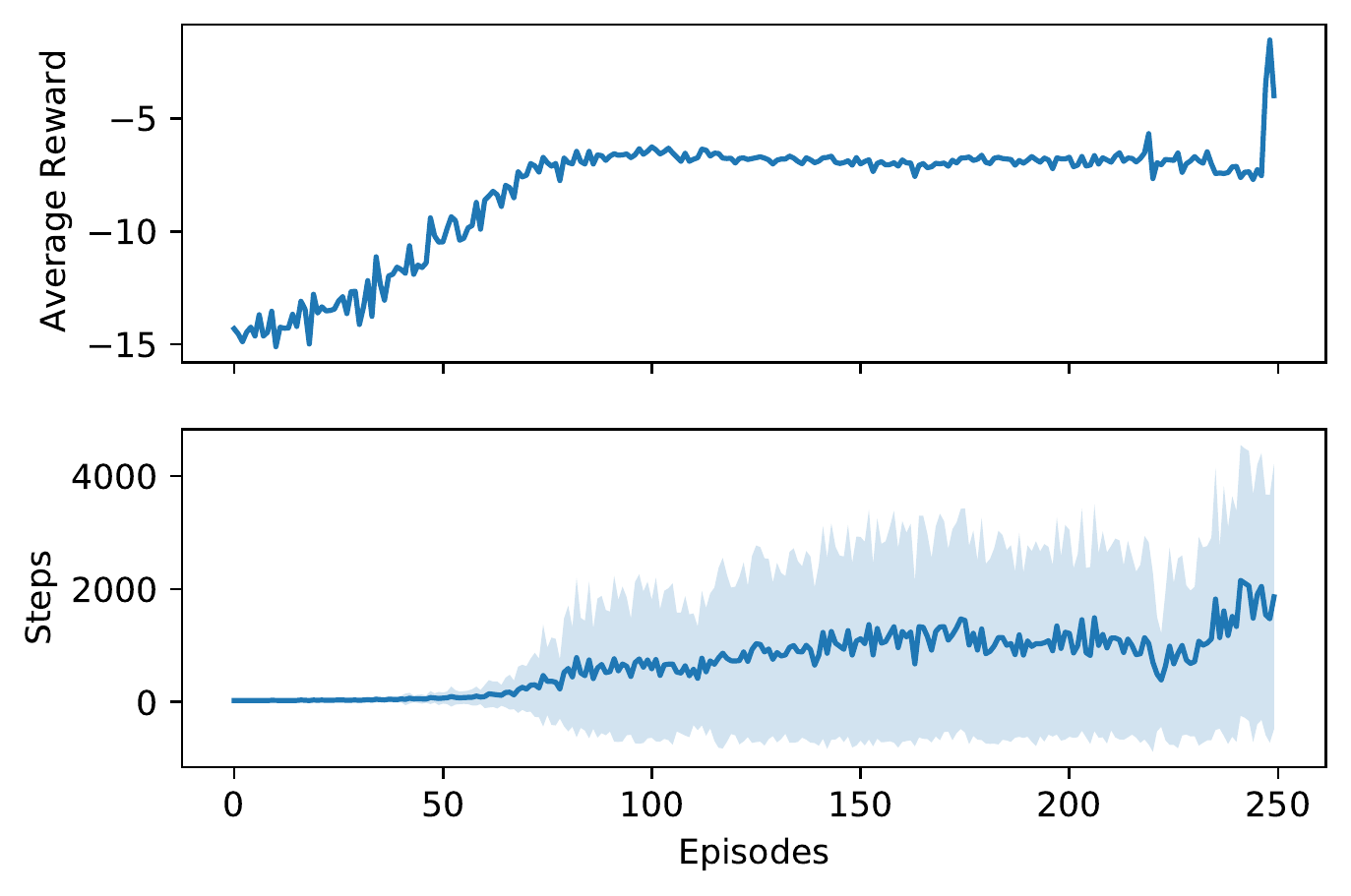}
    \includegraphics[width=0.3\textwidth]{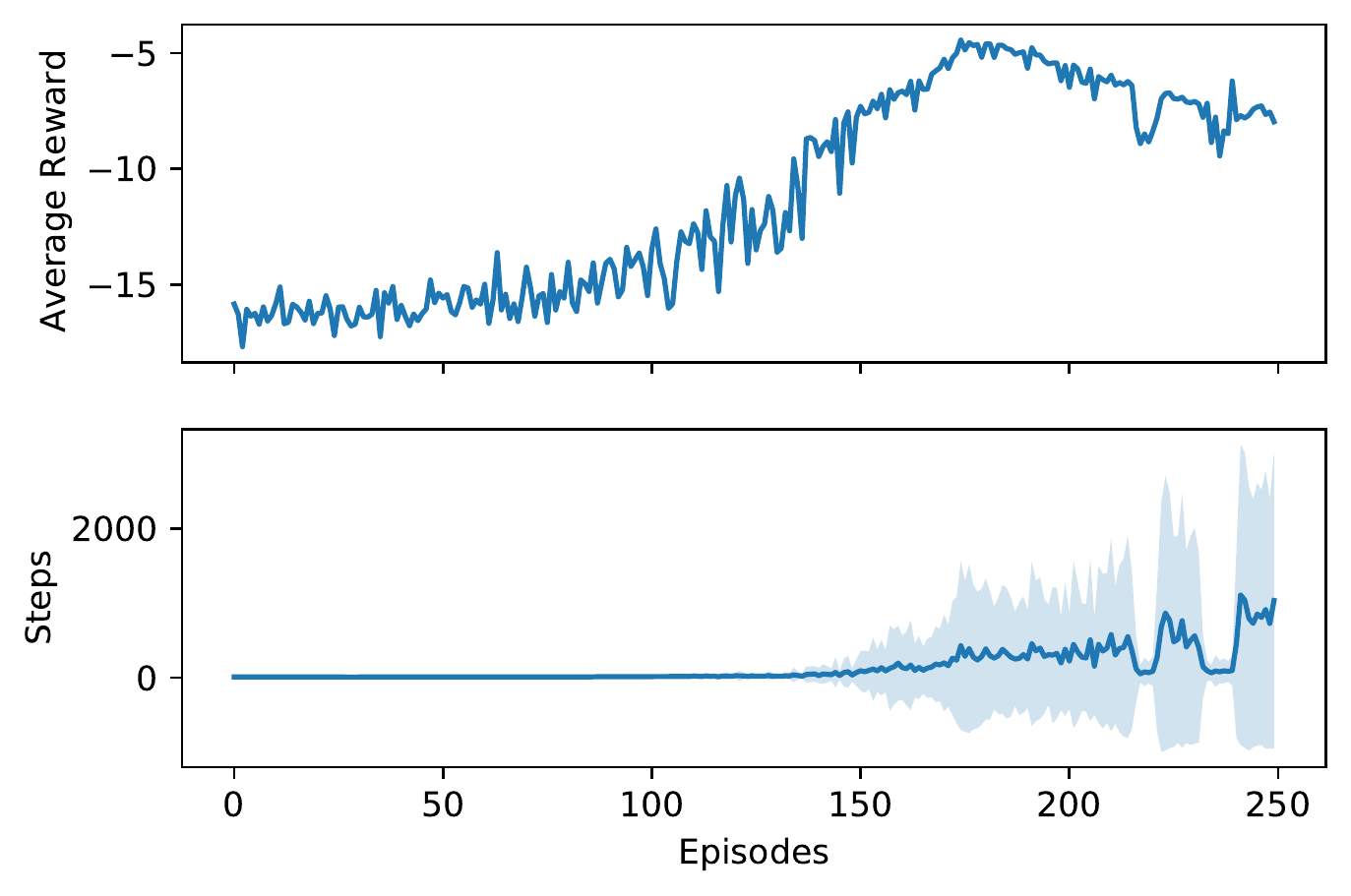}
    \includegraphics[width=0.3\textwidth]{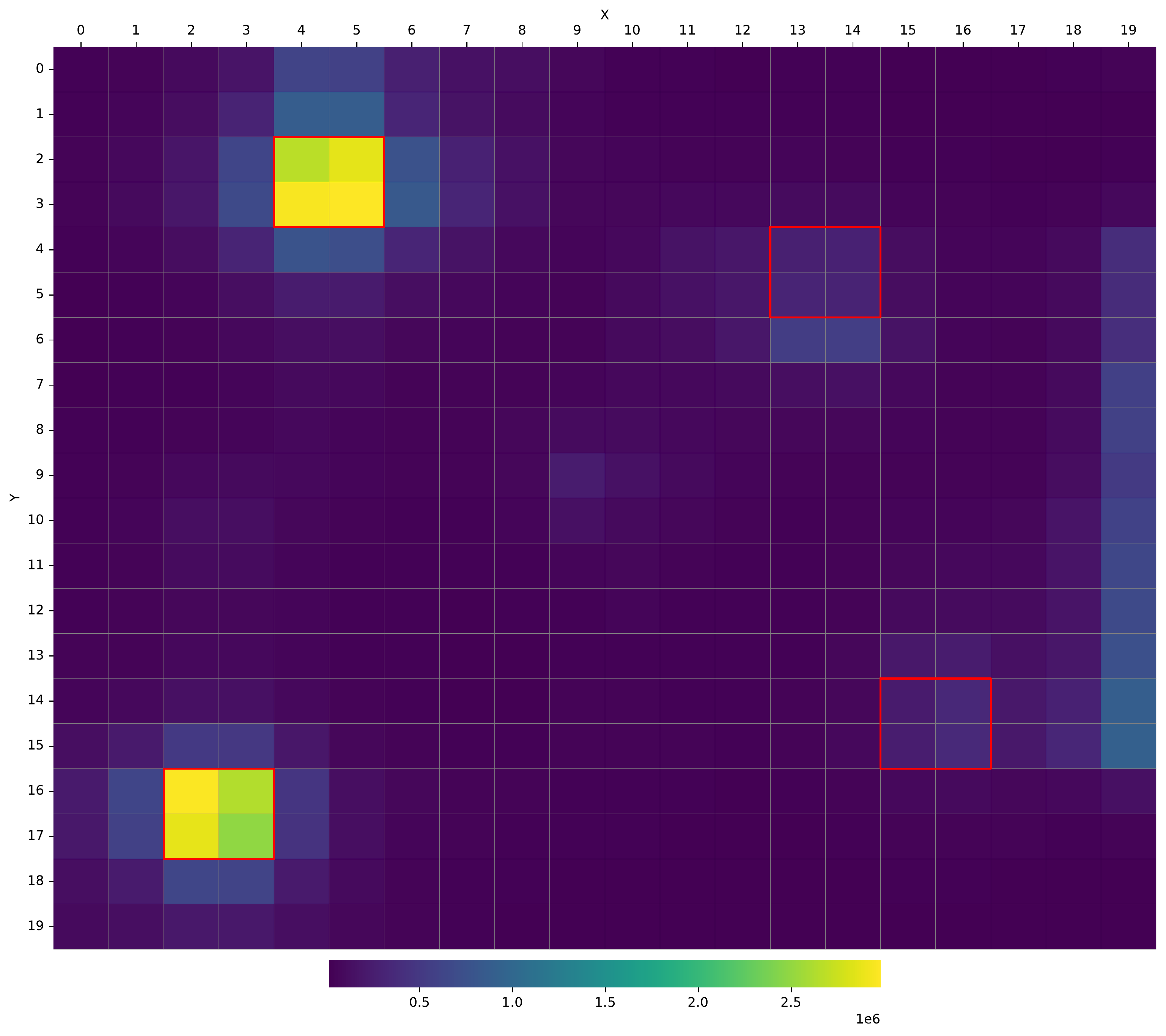}
    \includegraphics[width=0.3\textwidth]{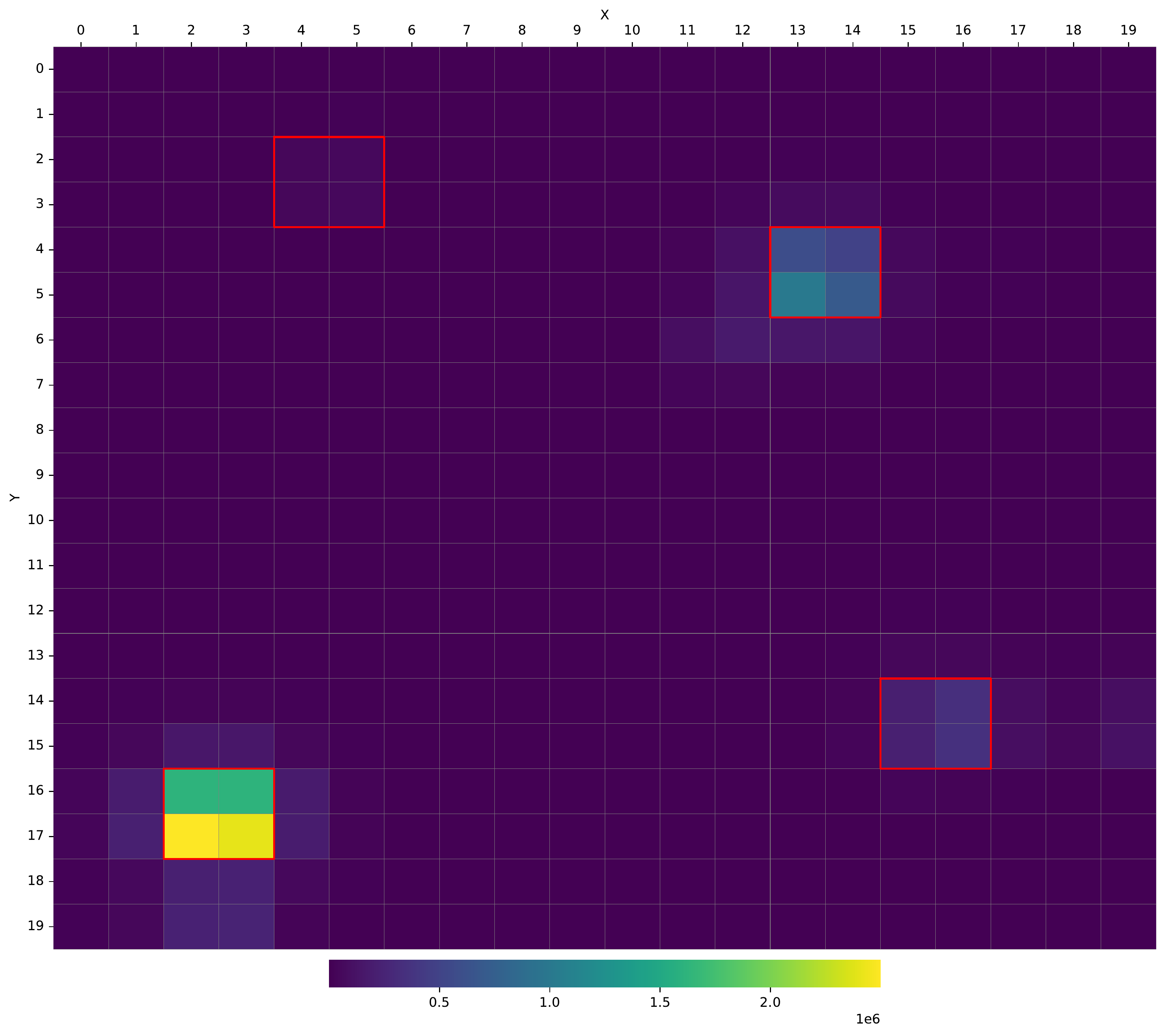}
    \includegraphics[width=0.3\textwidth]{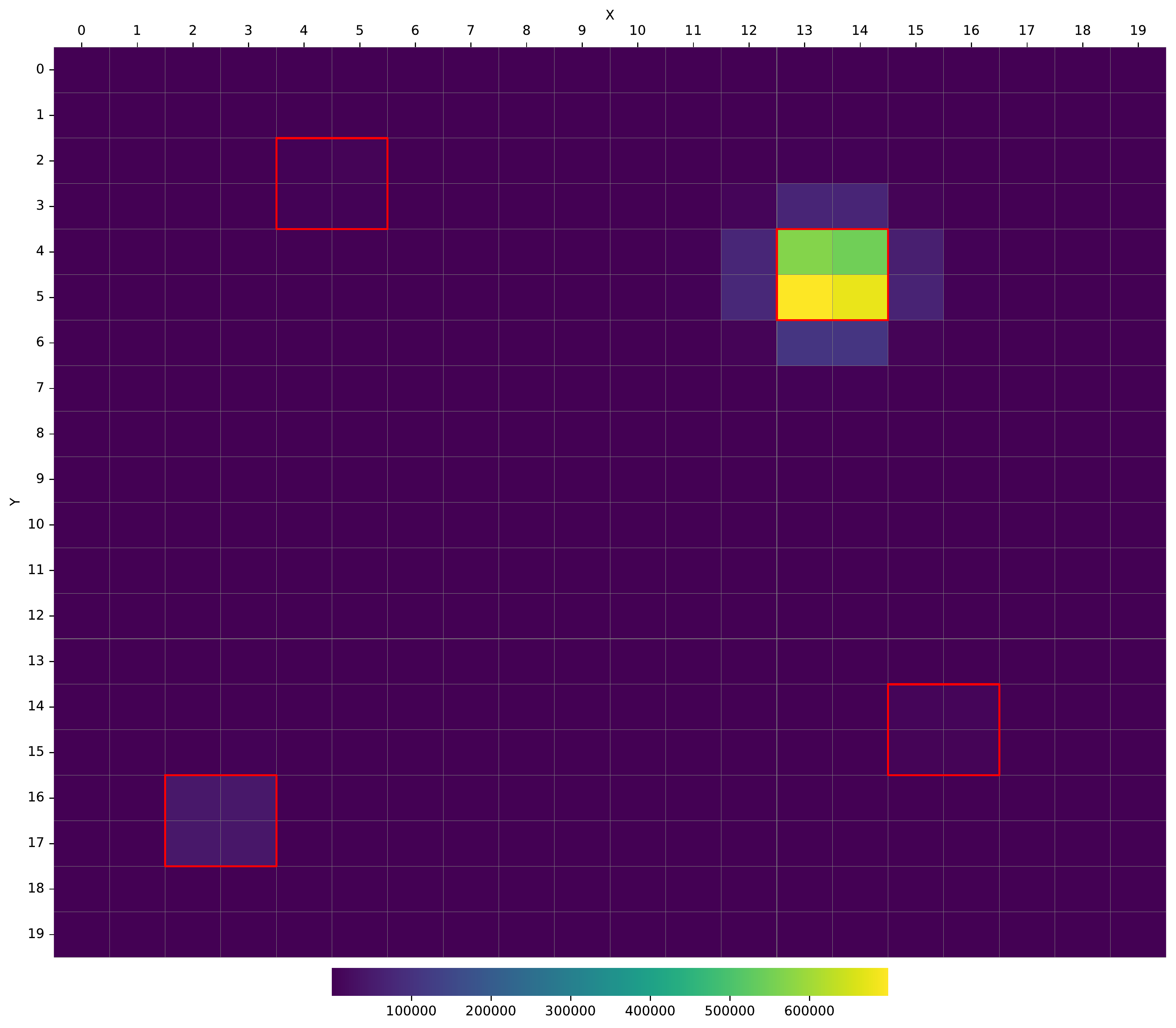}
    \includegraphics[width=0.3\textwidth]{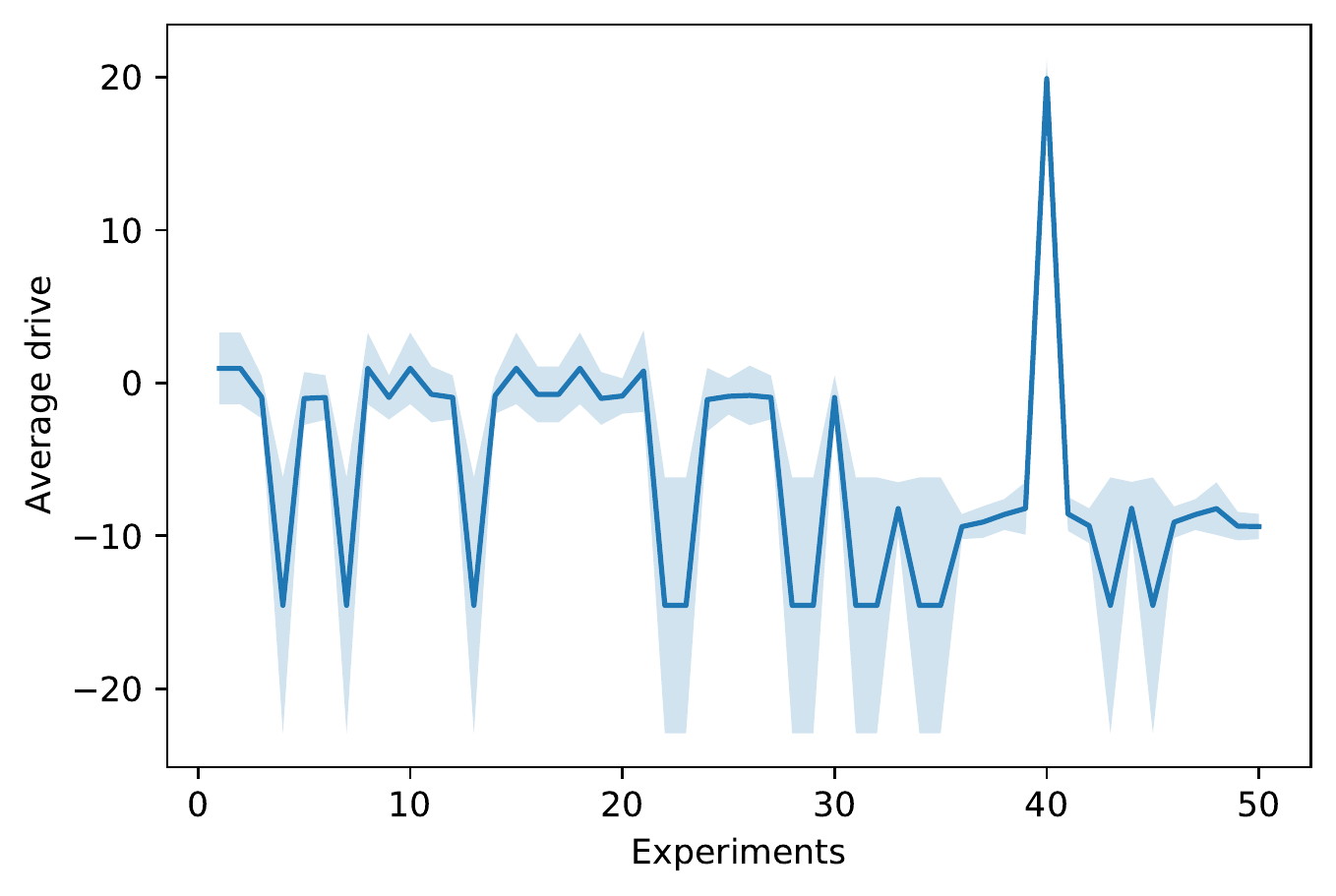}
    \includegraphics[width=0.3\textwidth]{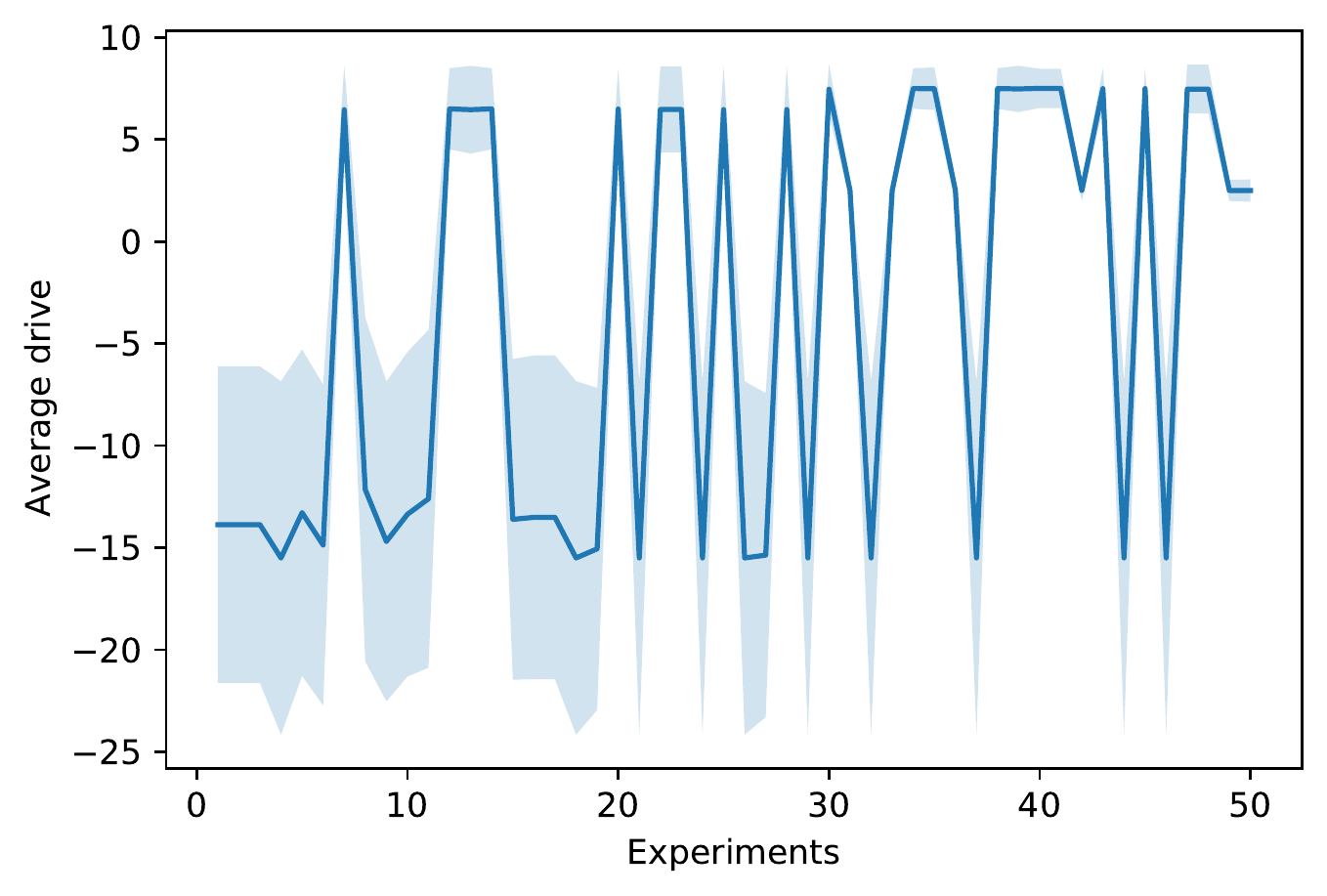}
    \includegraphics[width=0.3\textwidth]{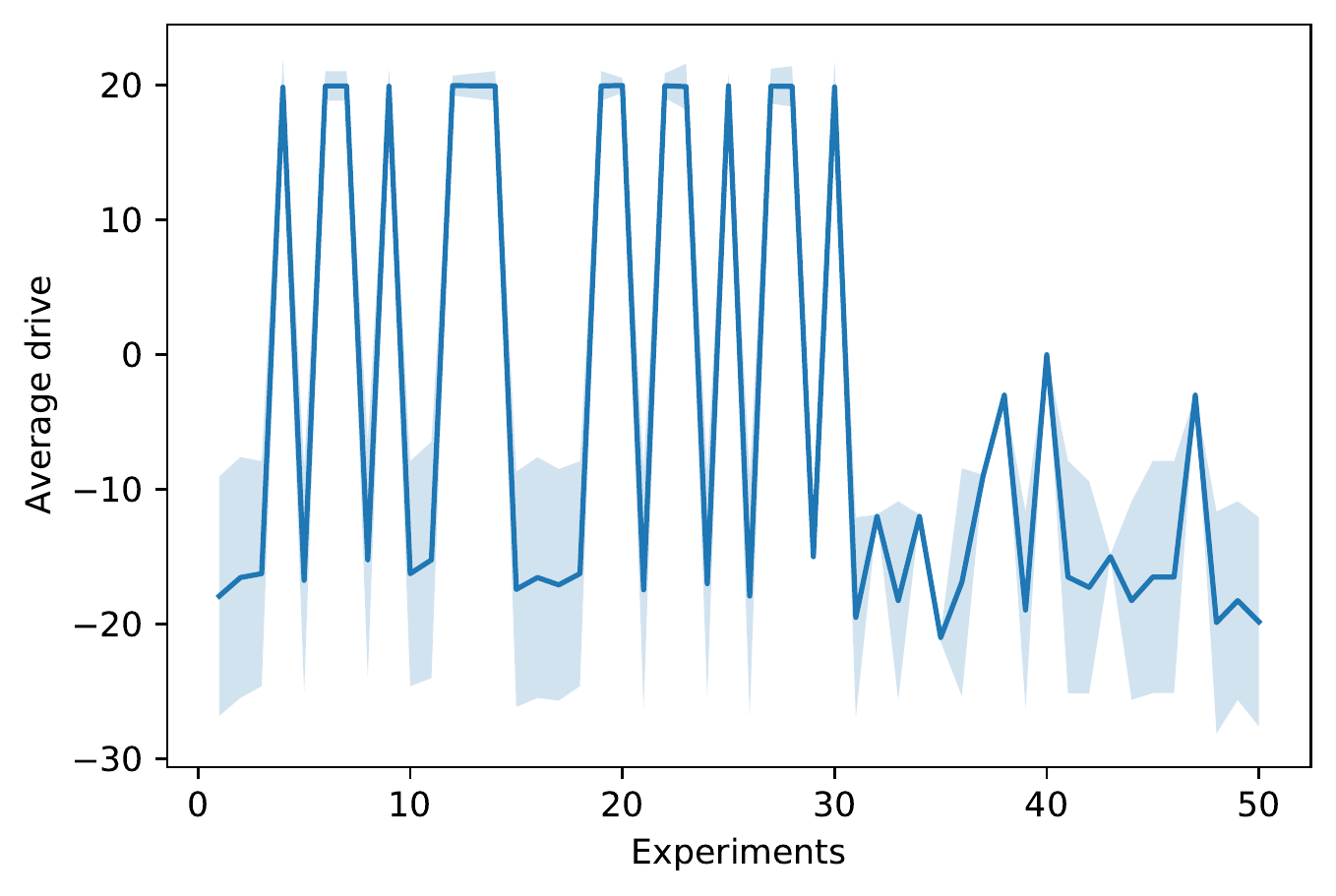}
    \caption{From left to right: EXP10 (slow metabolism), EXP11 (regular metabolism) and EXP12 (fast metabolism). From top to bottom: 1) average reward (top) and number of actions (bottom) per episode; 2) visits per environment position during training; 3) Average Drive Level in 50 tests when testing policies learned.}
    \label{fig:EXP10_EXP11_EXP12}
\end{figure*}

In EXP10, the agent with slow metabolism faces a challenging learning scenario, but, unlike EXP07, it overcomes it, learning a successful policy (Figure \ref{fig:EXP10_EXP11_EXP12}, first row). By analyzing the agent's movement through the environment (Figure \ref{fig:EXP10_EXP11_EXP12} first column, second row), we see that the agent stayed longer at recharge stations $A$ and $C$, which are the ones with the most significant associated pleasure. However, by preferring station $A$, the agent did not only aim to satisfy its pleasure. It also learned to balance between pleasure/\textit{liking} and homeostasis as this is the lowest recharge value station. Hence, the more suitable to the agent's profile.

In the testing phase, the agent of EXP10 (Figure \ref{fig:EXP10_EXP11_EXP12}, first column, last row) learned to stay in homeostasis and was successful in most experiments. This agent found a balance between gain in homeostasis and pleasure. In this case, the insertion of pleasure in an environment with different recharge values for each station was advantageous. It provided better results compared with the equivalent environment without pleasure - EXP04. Also, comparing the pleasure (the exact motivational mechanism) but with a different environment (EXP07), the agent's average drive is better in EXP10, indicating that the agent learns the balance between the energy needed and pleasure wanted. 

In EXP11, the agent with a regular metabolism successfully learned to balance needs and pleasure. As in EXP08, it achieves the best reward per episode (Figure \ref{fig:EXP10_EXP11_EXP12}, second column, first row) compared to the slow and fast metabolism agents. As shown in Figure \ref{fig:EXP10_EXP11_EXP12} - second column, second row -, the agent visits recharge stations $B$ and $C$ more frequently, which provide higher recharge values necessary for metabolism. Regarding pleasure, these stations offer a reasonable ($L_B = 2$) and maximum value ($L_C = 4$), respectively. Therefore, the decision between satisfying the drive and the pleasure was balanced.

In the testing phase, the agent of EXP11  (Figure \ref{fig:EXP10_EXP11_EXP12}, second column, last row) shows that the agent's average drive is above homeostasis in most experiments. So, the agent prefers having positive drives and living more to risk losing energy. Compared with EXP05 (same metabolism and same dynamics of recharge value but without pleasure), in EXP11, the average drive is further from homeostasis, indicating that pleasure is essential to this agent. Even though pleasure is important to this agent, the agent's performance tends to improve in an environment that allows different recharging strategies. We can see this when comparing EXP11 (different recharge + pleasure) with EXP08 (same recharge + pleasure). The average drive of EXP11 is closer to homeostasis than EXP08, so the agent appreciates pleasure but tries to balance it better with the needs. 

Regarding EXP12, given the high energy consumption of this agent, the initialization of the battery sensor and positioning in the environment significantly impact learning. Figure \ref{fig:EXP10_EXP11_EXP12} shows that the agent was able to learn, even though it could not perform the maximum number of actions in training episodes. Looking at Figure \ref{fig:EXP10_EXP11_EXP12} - second column, second row -, we see that the agent focuses on staying at station $B$ as it provides the maximum recharge value but not a high pleasure ($L_B = 2$). With that, we can see it priorities its needs and not pleasure.

In the testing phase, the agent of EXP12 (Figure \ref{fig:EXP10_EXP11_EXP12}, right column, last row) shows that the agent's average drive is the maximum possible in many cases, indicating that the agent is focused on staying alive, independent of homeostasis or pleasure associated. This behavior is similar to EXP06 (different recharge value without pleasure). We can conclude that pleasure is not vital for a fast metabolism agent. Nevertheless, comparing the results of EXP12 with EXP09, we can conclude that environments with different recharge values result in distinct behaviors for the same agent. In the same recharge environment as in EXP09, the agent's policy is unstable with the drive above and below homeostasis. In contrast, in an environment with different recharge possibilities (EXP12), the agent's drive is the maximum in many cases. In conclusion, for agents with a fast metabolism, an environment with varying values in recharge areas impacts the agent's behavior more than pleasure insertion.

Table \ref{tab:summayResults} shows a summary of all the experiments results.

\begin{table*}[htb]
\caption{Summary of the results. Experiments EXP01 - EXP06 uses M1 mechanism (\textit{wanting}) and EXP07 - EXP12 uses M2 mechanism (\textit{wanting + liking}). All the stations provide the same recharge value in EXP01 - EXP03 and EXP07 - EXP09, while for EXP04 - EXP06 and EXP10 - EXP12, each station provides a different recharge value.} \label{tab:summayResults}%
\begin{center}
{
\begin{tabular}{p{1.5cm}  p{14cm}}
\toprule
 Exp. no. & Summary\\
\midrule
EXP01& Able to remain in homeostasis in most of the cases using the strategy of being close to the station (during the training phase, stayed close to stations $A$ and $C$) \\
EXP02& Able to remain close to homeostasis (usually a bit above) using the strategy of being close (or inside) the stations (closer than EXP01)  \\
EXP03& For this agent is harder to stay in homeostasis due to the high energy need (and the station’s recharge value is exactly the metabolism consumption). The strategy learned is to stay inside the closest station \\
EXP04&  Remains in homeostasis (or close enough) or with the maximum positive drive. The policy learned is stay near (nearest than it was in EXP01) the station $A$ (smallest $\delta D_{i_{P_j}}/\delta t$) \\
EXP05&  Remains in homeostasis but increases the number of negative drives compared to EXP02. The strategy learned is to stay close to station $D$ (which offers a good metabolism recharge) or $B$. The only station that this agent does not visit is $A$ (the energy provided is not good for its needs)  \\
EXP06& Stable drives, try to achieve the maximum positive drive or not stay too long with negative drives. The policy learned is to stay in station $B$ (maximum recharge)  \\
\bottomrule
EXP07& Prioritizes the pleasure instead of the drive reduction by staying more time inside station $C$ (maximum pleasure) \\
EXP08 & Prefers to satisfy the pleasure instead of the drive reduction. This agent visits more times to station $C$, $A$ and $B$, respectively \\
EXP09 & For this agent survival is more important than getting pleasure, so it goes to the closest station independently of the pleasure value associated \\
EXP10 & Balances drive reduction and getting pleasure by visiting station $A$ (better for drive reduction) and $C$ (better for pleasure)  \\
EXP11& Balances drive reduction and getting pleasure by visiting stations $B$ and $C$. For this agent, station $A$ is not beneficial for the drive, and station $D$ is not good for pleasure  \\
EXP12& Prefers the drive reduction by going to station $B$ (maximum recharge value – more than required for survival in each time step) \\
\bottomrule
\end{tabular}}
\end{center}
\end{table*}

\section{Discussion and Conclusion}\label{sec:conc}
In this work, we proposed two motivational models based on Hull's Drive Reduction Theory and the Liking mechanisms associated with pleasure. Based on these models, we built autonomous robots that learned how to behave driven by their motivational mechanism. We trained these agents via a reinforcement learning strategy. We simulated three agents with different energy consumption rates to analyze how agents with distinct metabolic curves would act in the same environment. We also simulated distinct environments to assess how they impact learning and the strategies adopted by each agent.

We observed all agents trying to maintain homeostasis when considering only the drive reduction (wanting) model. However, when survival is at risk, the agent plays safe and keeps itself as fed as possible, with little concern on equilibrium. This is the case for the fast metabolism agent. When we introduced the hedonic dimensions, the agents successfully learned to balance needs and pleasure according to their characteristics and the environment. Nevertheless, pleasure is not relevant when the agent needs to survive. The regular metabolism agent achieved the most efficient policies in our most complex scenario with different recharge areas. Because it does not drop energy too fast or too slow, it takes advantage of the environment dynamics to perform as well as possible. 

In general, the results show that both the environment and the agent's inner characteristics play an essential role in the resulting strategy learned, as it happens with humans. Our results show that if we change the environment, the agents adapt to it to maximize their returns. However, if we submit different agents to the same experiences, the resulting behavior defers drastically. This shows that we can derive very different agents by changing minor characteristics, such as the metabolic rate, while maintaining everything else. 

Regarding the hedonic dimension, referenced as pleasure in this work, it has a specific fixed value for each recharge station. It is independent of the metabolism, the driving urgency, and the level of deprivation of that particular resource. We use this setting as a first step to understanding the impact of having preferences incompatible with the agent's needs and that are not updated. Using this setting can make it harder for the robot to adapt to the environment while trying to maintain homeostasis. The next step of our research is to develop a dynamic hedonic value updated during the interaction and according to the robot's context.

In future works, we plan to execute these experiments in a more complex and non-deterministic environment. Also, we want to analyze agents with different levels of homeostasis and simulate addictive behavior to understand the underlying learning process. One crucial part of our research is studying decision-making when we have multiple drives, so we intend to add more drives in future work. Moreover, we want to analyze the behavior in challenging situations, as a specific station is closed at a determined time or another agent is using it. We want to see the agent's behavior in situations like that.

As emotions intensely impact motivation, we have been working on an emotional architecture that can be used as part of motivational and decision-making systems. This model assesses how our emotional state influences the reward function associated with drive reduction and changes the robot's decisions.

\section*{Acknowledgments}
L. M. Berto is funded by CAPES, QuintoAndar and São Paulo Research Foundation (FAPESP) grant 2021/07050-0. E. L. Colombini is partially funded by CNPq PQ-2 grant (315468/2021-1). A. S. Simoes is partially funded by CNPq PQ-2 grant (312703/2019-8). This work was carried out within the scope of PPI-Softex with support from the MCTI, through the Technical Cooperation Agreement [01245.013778/2020-21].

\bibliographystyle{IEEEtran}
\bibliography{main}

\end{document}